\documentclass{article}

\usepackage[table]{xcolor} %
\usepackage[final]{corl_2026} %

\usepackage{algorithm}
\usepackage{algpseudocode}
\usepackage{courier}
\usepackage{amsmath}
\usepackage{comment}
\usepackage{amsfonts}
\usepackage{graphicx}
\usepackage{url}
\usepackage{booktabs}
\usepackage{tabularx}
\usepackage{wrapfig}

\usepackage{booktabs}
\usepackage{multirow}
\usepackage{threeparttable}
\usepackage{adjustbox}

\usepackage{array}
\usepackage{float} %
\usepackage{tikz}
\usetikzlibrary{positioning,arrows.meta,calc}

\let\setdropshade\relax
\let\dropcell\relax

\let\dropshadeglobal\relax

\newcommand{\setdropshade}[1]{%
  \def\shade{red!6}%
  \ifdim #1pt>5pt  \def\shade{red!10}\fi
  \ifdim #1pt>10pt \def\shade{red!14}\fi
  \ifdim #1pt>15pt \def\shade{red!18}\fi
  \ifdim #1pt>20pt \def\shade{red!22}\fi
  \ifdim #1pt>25pt \def\shade{red!26}\fi
  \ifdim #1pt>30pt \def\shade{red!30}\fi
  \ifdim #1pt>35pt \def\shade{red!34}\fi
  \ifdim #1pt>40pt \def\shade{red!38}\fi
  \ifdim #1pt>45pt \def\shade{red!42}\fi
  \ifdim #1pt>50pt \def\shade{red!46}\fi
  \ifdim #1pt>55pt \def\shade{red!50}\fi
  \ifdim #1pt>60pt \def\shade{red!54}\fi
  \ifdim #1pt>65pt \def\shade{red!58}\fi
  \ifdim #1pt>70pt \def\shade{red!62}\fi
  \ifdim #1pt>75pt \def\shade{red!66}\fi
  \ifdim #1pt>80pt \def\shade{red!70}\fi
  \ifdim #1pt>85pt \def\shade{red!74}\fi
  \ifdim #1pt>90pt \def\shade{red!78}\fi
  \ifdim #1pt>95pt \def\shade{red!82}\fi
}

\newcommand{\dropcell}[1]{%
  \begingroup
  \setdropshade{#1}%
  \xdef\dropshadeglobal{\shade}%
  \endgroup
  \cellcolor{\dropshadeglobal}$\downarrow$\,#1%
}

\usepackage{hyperref}

\usepackage{tcolorbox}
\tcbuselibrary{breakable}
\usepackage{enumitem}
\newcommand\blfootnote[1]{%
  \begingroup
  \renewcommand\thefootnote{}\footnote{#1}%
  \addtocounter{footnote}{-1}%
  \endgroup
}
\title{Disentangling Spurious Correlations in Vision-Language-Action Models via Predicting Domain-Invariant Latent Lookahead}

\author{
  Junghyun Kim$^{1,2*}$ \,\,\, Ngseo Kim$^{2*}$ \,\,\, ChungWoo Lee$^2$ \,\,\, Seoyeon Lee$^2$ \,\,\, Woo-Jeong Baek$^5$ \and
  \textbf{Adam Zhou$^1$ \,\,\, Chip Huyen$^1$ \,\,\, Jun-Ki Lee$^{2\dagger}$ \,\,\, Gi-Cheon Kang$^{3\dagger}$ \,\,\, Byoung-Tak Zhang$^{2,4\dagger}$}
  \\ \\
  $^1$OpenMind, San Francisco, CA, USA \,\,\,
  $^2$Seoul National University, Seoul, Korea \and
  $^3$Ajou University, Korea \,\,\,
  $^4$Tommoro Robotics, Korea \,\,\,
  $^5$Hyundai Motors, Korea
}

\hypersetup{
  pdftitle={Disentangling Spurious Correlations in Vision-Language-Action Models via Predicting Domain-Invariant Latent Lookahead},
  pdfauthor={Junghyun Kim, Ngseo Kim, ChungWoo Lee, Seoyeon Lee, Woo-Jeong Baek, Adam Zhou, Chip Huyen, Jun-Ki Lee, Gi-Cheon Kang, Byoung-Tak Zhang}
}
\makeatletter
\renewcommand{\@conferencelocation}{Austin, Texas, USA}
\makeatother

\begin{document}
\maketitle
\blfootnote{$^*$Equal contribution. $^\dagger$ Corresponding authors.}

\begin{abstract}
Vision-Language-Action (VLA) models remain brittle under visual distribution shifts, often relying on spurious correlations tied to domain-specific factors rather than task-relevant structure. We propose \emph{Domain-Invariant Latent Lookahead} (DILL), a representation-learning framework that mitigates shortcut learning in VLA policies. Our key idea is to supervise policies with domain-invariant future latents learned from domain-transformed trajectory data. A Task-Domain Encoder is trained with contrastive objectives and Gaussian disentanglement regularization to separate task-relevant structure from domain-specific visual variation. The learned encoder then provides future latents for VLA policy learning through lookahead prediction and domain disentanglement, encouraging the policy to focus on task-relevant structure rather than incidental visual factors. 
Counterfactual task--view evaluations show that DILL reduces shortcut reliance, while LIBERO-Plus evaluations demonstrate improved visual robustness, with 69.1\% average success---11.4 percentage points above the strongest baseline. Real-world manipulation experiments further support DILL's applicability beyond controlled simulation. Complementary latent-space diagnostics show that these behavioral gains are accompanied by representations that better preserve task-consistent structure while suppressing domain-specific variation. Our project page is available at \url{https://dill-vla.github.io/}.

\end{abstract}


\section{Introduction}

A longstanding goal of robot learning is to build robots that generalize across a wide range of tasks and environments. Recent advances in Vision-Language-Action (VLA) models~\citep{rt2,octo,openvla,black2024pi0} trained on large-scale robot datasets~\citep{oxe,khazatsky2024droid} have established a promising path toward generalist robot policies by grounding actions in language and visual observations. Despite this progress, VLA models often remain brittle under distribution shift, particularly under incidental visual variation such as changes in viewpoint, background, lighting, or camera configuration~\citep{libero_plus,zhang2025vlaarena,wang2026liberox}, which should be irrelevant to task success. This brittleness undermines reliable out-of-distribution generalization and remains a major obstacle to practical deployment.

\begin{figure}[!t]
    \centering
    \includegraphics[width=0.78\linewidth]{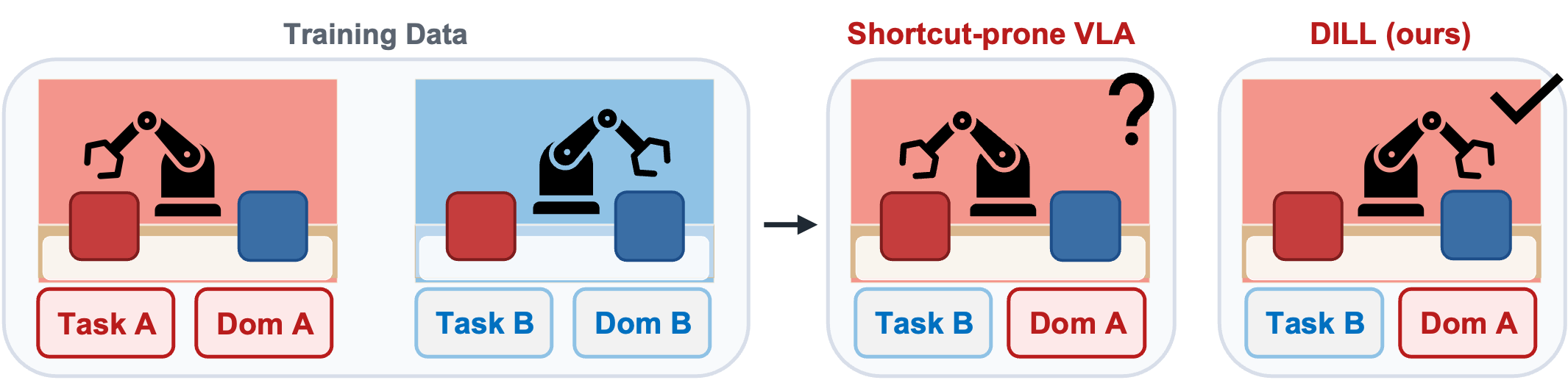}
    \vspace{-0.2em}
    \caption{\textbf{Shortcut learning under task--domain confounding.}
    When tasks and visual domains are spuriously correlated, a VLA may use domain-specific appearance as a shortcut for action selection. DILL reduces this failure by conditioning actions on a domain-invariant latent lookahead.}
    \label{fig:shortcut_teaser}
    \vspace{-1.0em}
\end{figure}

Such failures can be naturally understood through the lens of \emph{shortcut learning}~\citep{geirhos2020shortcut,shortcut_learning_in_GRPs}, where policies rely on task-irrelevant visual cues that correlate with successful actions in the training distribution rather than the task-relevant semantics required for robust control. VLA models are particularly susceptible to shortcut learning for two reasons. First, robot datasets often exhibit limited within-dataset diversity and are fragmented across collection sources, causing the same task or action to repeatedly co-occur with particular visual cues~\citep{shortcut_learning_in_GRPs}. 
Second, VLA models often inherit representations from pretrained vision-language backbones~\citep{radford2021learning,bai2025qwen25vltechnicalreport} that are optimized for visual understanding or image-language alignment rather than control, and may therefore preserve domain-specific visual information irrelevant to action generation. When mapped to actions, these representations can allow spurious visual correlations in training data to be absorbed into the policy's decision rule~\citep{de2019causal_confusion_in_IL}, producing the task-substitution failure illustrated in Fig.~\ref{fig:shortcut_teaser}: under counterfactual task--domain recomposition, the policy may execute the task associated with the observed domain.

To mitigate such shortcuts, a VLA policy should learn representations that preserve task-relevant structure while discarding incidental visual variability. A natural learning signal for this purpose is future state prediction~\citep{zheng2025flare,sun2026vla_jepa}. Predicting how a task will evolve can encourage models to capture factors that determine future task states, such as object configuration, task progress, and action-conditioned scene changes. In this sense, future prediction provides a form of supervision that is more closely tied to control than to static visual appearance. However, future prediction alone is not sufficient. If the target is a raw future observation or an entangled visual latent, the model may still preserve domain-specific factors such as viewpoint or background that help predict future appearance but do not determine the correct action. Thus, the key question is not simply whether a VLA policy should predict the future, but \emph{what representation of the future it should predict}. We argue that the predictive target should be \emph{domain-invariant}: it should retain the future task structure needed for control while suppressing domain-specific factors that can act as shortcuts.

We propose \emph{Domain-Invariant Latent Lookahead} (DILL), a predictive representation learning framework for robust VLA control. DILL first learns a Task-Domain Encoder from domain-transformed trajectory chunks using contrastive objectives over two pair types: \emph{task-positive pairs} that preserve trajectory chunk across domain changes, and \emph{domain-positive pairs} that share the same domain condition across different chunks. 
A Gaussian disentanglement loss further separates task and domain latents: task latents remain stable across changes in viewpoint, environment appearance, camera configuration, and visual degradation, while domain latents capture such domain-specific factors.
During policy learning, the pretrained Task-Domain Encoder provides future latent supervision: the VLA policy predicts a lookahead latent aligned with the future task latent, while its current representation is regularized to be disentangled from the corresponding domain latent. The predicted lookahead latent and current representation jointly condition the action head, encouraging the policy to base its actions on task-relevant future structure rather than incidental visual factors.

The central effect of DILL is to reshape the information that the policy uses to choose actions: instead of allowing domain-specific visual cues to enter the action head as reliable proxies for the task, DILL conditions actions on a predicted future latent that is stable across domain changes and informative about the commanded behavior.
This makes the learned action representation less tied to where or how a scene is observed, and more tied to the information needed to produce the correct action. 
Empirically, we evaluate this effect through complementary behavioral and representational analyses. Under counterfactual task--view compositions, DILL substantially reduces shortcut reliance and improves OOD task success, showing that policies are less likely to substitute the commanded task with the task spuriously associated with the observed view. 
{On LIBERO-Plus visual perturbations, DILL achieves the highest average success rate among matched-input baselines and a substantially smaller average performance drop across camera, lighting, background, and sensor-noise shifts.} 
We further test DILL on a physical robot to assess its effectiveness beyond controlled simulation.
Latent-space diagnostics further show that the final action-conditioning representation is organized more by task-consistent trajectory content than by shared visual appearance. Together, these results show that shortcut learning in VLA policies can be mitigated not only by increasing data diversity, but by explicitly shaping the predictive representation used for control.

\section{Related Work}

\subsection{Spurious Correlations in Robot Learning}
Spurious correlations~\cite{shortcut_learning_in_GRPs} have emerged as an important challenge in robot learning, where policies may rely on incidental visual factors, such as camera viewpoint, background, and lighting, that correlate with successful actions in the training data~\cite{libero_plus,zhang2025vlaarena,zhou2025libero_pro,wang2026liberox}. Prior work has addressed this issue from two perspectives: \emph{data-centric} and \emph{representation-centric} approaches. Data-centric approaches increase nuisance diversity through novel-view synthesis~\cite{VISTA} or synthetic viewpoint augmentation~\cite{shortcut_learning_in_GRPs}. Within representation-centric approaches, one line of work suppresses task-irrelevant factors by building invariance into visual representations, for example, by reducing sensitivity to viewpoint changes~\cite{liu2024robouniview,MVWM,ReViWo}. Another line of work injects task-relevant structure, such as spatial information~\cite{zhang2025inspire}. Our work bridges these directions by using \emph{predictive future latents} as task-relevant supervision while explicitly suppressing task-irrelevant visual factors through domain-invariant latent disentanglement.

\subsection{World Models and Future Prediction in VLA Models}

Recent VLA work has incorporated \emph{future prediction} or \emph{world modeling} signals to complement direct perception-to-action learning by anticipating the consequences of actions~\cite{zheng2025flare,hu2024video_prediction_policy,zhao2025cotvla}. One line performs pixel-level imagination by generating future frames or subgoal observations before acting, as in the \emph{GR} series, Ctrl-World, CoT-VLA, and SuSIE~\cite{wu2023unleashing_gr1,cheang2024gr2,guo2025ctrl,zhao2025cotvla,black2023zero_susie}. These approaches provide interpretable visual foresight, but they can be costly and prone to compounding errors. The second line moves future prediction into latent space by predicting or aligning future latent representations, including Video Prediction Policy, FLARE, VLA-JEPA, and FRAPPE~\cite{hu2024video_prediction_policy,zheng2025flare,sun2026vla_jepa,zhao2026frappe,assran2023ijepa,assran2025vjepa2}. These methods avoid explicit pixel rollouts while encouraging representations that capture future structure useful for control. In contrast, DILL focuses on what the policy is asked to predict: rather than aligning to raw or entangled future latents, it learns future targets whose task-relevant structure is separated from domain-specific visual variation.

\section{Method}

\begin{figure}[htbp]
    \centering
    \includegraphics[width=\linewidth]{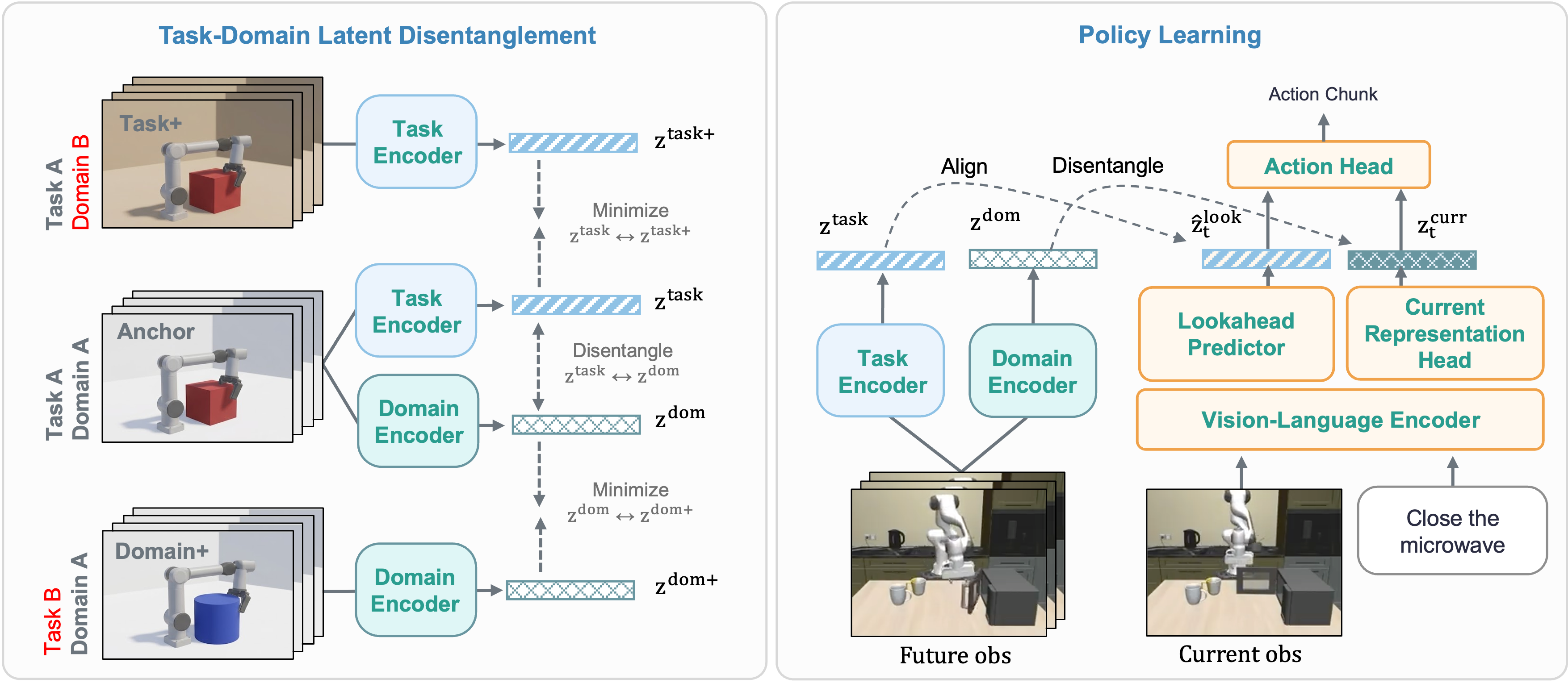}
    \caption{
    \textbf{Overview of Domain-Invariant Latent Lookahead.}
    We first learn a Task-Domain Encoder that maps observation chunks into task and domain latents.
    During policy learning, a Lookahead Predictor predicts a future task latent from the current observation and instruction, while a Current Representation Head produces a domain-disentangled current representation.
    The two policy latents condition the action head for control.
    }
    \label{fig:main}
\end{figure}

\subsection{Overview}

Domain-Invariant Latent Lookahead (DILL) trains VLA policies with domain-invariant predictive supervision through a two-stage procedure.
As shown in Fig.~\ref{fig:main}, we first learn a Task-Domain Encoder using domain-transformed trajectory data.
The task encoder produces task latents that are encouraged to remain invariant across domain changes, while the domain encoder produces domain latents that capture domain-specific visual variation.
Both latents are learned using contrastive objectives defined by task-positive and domain-positive pairs, together with a Gaussian disentanglement loss that encourages the task and domain latents to be separated.

We then use the learned Task-Domain Encoder to supervise VLA policy learning.
For each training trajectory, the encoder provides future task and domain latents as supervisory signals.
From the current observation and language instruction, the policy predicts a lookahead latent and a current representation: the former is aligned with the future task latent, while the latter is regularized to be disentangled from the domain latent.
These two policy latents are fed to the action head for control, encouraging the policy to combine a predicted future task representation with a current representation discouraged from carrying domain-specific visual information.
As a result, the policy is trained to rely on task-relevant future structure rather than shortcut-inducing visual cues.
At test time, the policy requires only the current observation and instruction.

\subsection{Task-Domain Latent Disentanglement}
\label{sec:task_domain_disentanglement}

The Task-Domain Encoder learns task and domain latents from video observations: task latents are encouraged to preserve task-relevant structure, while domain latents capture domain-specific visual factors.
Let \(\mathbf{o}_t^i=(o_t^i,\ldots,o_{t+T_v-1}^i)\) denote an observation chunk from trajectory \(i\), and let \(\mathcal T_{\eta}\) denote a domain transformation with condition \(\eta\), where \(\eta\) specifies the full domain condition.
We construct task-positive and domain-positive pairs:
\begin{equation}
    \underbrace{
    \big(
    \mathcal T_{\eta_1}(\mathbf{o}_t^i),
    \mathcal T_{\eta_2}(\mathbf{o}_t^i)
    \big)
    }_{\text{task-positive}},
    \quad
    \eta_1 \neq \eta_2,
    \qquad
    \underbrace{
    \big(
    \mathcal T_{\eta}(\mathbf{o}_t^i),
    \mathcal T_{\eta}(\mathbf{o}_{t'}^j)
    \big)
    }_{\text{domain-positive}},
    \quad
    (i,t) \neq (j,t') .
\end{equation}
Task-positive pairs preserve the same observation chunk while changing the domain condition, whereas domain-positive pairs share the same domain condition across different observation chunks.
The transformation families include viewpoint changes, environment appearance variation, camera heterogeneity, and visual degradation.
See Appendix~\ref{app:domain_pairs} for details.

Let \(x=\mathcal T_{\eta}(\mathbf{o})\) denote a transformed observation chunk.
The Task-Domain Encoder consists of two separate video encoders: a task encoder and a domain encoder.
Each encoder maps the input observation chunk to a pooled latent vector:
\begin{equation}
    z^{\mathrm{task}}
    =
    E_{\psi}^{\mathrm{task}}(x), 
    \qquad
    z^{\mathrm{dom}}
    =
    E_{\xi}^{\mathrm{dom}}(x).
\end{equation}
Implementation details are provided in Appendix~\ref{app:impl_details}.

Both encoders are trained with the same InfoNCE form but with different positive pair constructions.
For a minibatch of positive pairs \(\{(x_i,x_i^+)\}_{i=1}^{B}\), let \(z_i\) and \(z_i^+\) denote the pooled latent vectors from the corresponding encoder:
\begin{equation}
    \mathcal L_{\mathrm{NCE}}
    =
    -
    \frac{1}{B}
    \sum_{i=1}^{B}
    \log
    \frac{
        \exp(\mathrm{sim}(z_i,z_i^+)/\tau)
    }{
        \sum_{k=1}^{B}
        \exp(\mathrm{sim}(z_i,z_k^+)/\tau)
    }
    \label{eq:nce}
\end{equation}
where \(\mathrm{sim}(\cdot,\cdot)\) denotes cosine similarity and \(\tau\) is a temperature.
Applying Eq.~\ref{eq:nce} to the task encoder with task-positive pairs and to the domain encoder with domain-positive pairs gives \(\mathcal L_{\mathrm{task}}\) and \(\mathcal L_{\mathrm{dom}}\), respectively.

To further disentangle the task and domain latents, we introduce a Gaussian disentanglement loss by applying SIGReg~\cite{balestriero2025lejepaprovablescalableselfsupervised} to their concatenation.
For each minibatch, we form a joint task-domain latent:
\begin{equation}
    u_i^{\mathrm{td}}
    =
    [z_i^{\mathrm{task}};z_i^{\mathrm{dom}}],
    \qquad
    \mathcal L_{\mathrm{dis}}^{\mathrm{td}}
    =
    \mathcal R_{\mathrm{dis}}
    \left(
    \{u_i^{\mathrm{td}}\}_{i=1}^{B}
    \right),
\end{equation}
where \([\cdot;\cdot]\) denotes concatenation.
Unlike applying distributional regularization to each latent separately, this joint regularizer acts on the concatenated task-domain latent.
Under a Gaussian approximation, matching the joint latent to an isotropic Gaussian discourages cross-covariance between the task and domain latents, thereby encouraging approximate disentanglement.
Appendix~\ref{app:gaussian_disentanglement} provides the corresponding argument.

The task-domain disentanglement objective is
\begin{equation}
    \mathcal L_{\mathrm{TDD}}
    =
    \lambda_{\mathrm{task}}\mathcal L_{\mathrm{task}}
    +
    \lambda_{\mathrm{dom}}\mathcal L_{\mathrm{dom}}
    +
    \lambda_{\mathrm{dis}}^{\mathrm{td}}
    \mathcal L_{\mathrm{dis}}^{\mathrm{td}}.
\end{equation}

\paragraph{Task-domain encoder pretraining.}
We pretrain the task and domain encoders on large-scale domain-transformed trajectory chunks from ManiSkill~\citep{tao2024maniskill3}, MimicGen~\citep{mandlekar2023mimicgen}, and a subset of OXE~\citep{oxe}.
We refer to these data as the source trajectory collection.
The pretrained encoders define the task and domain latent spaces used later to provide future supervision for VLA policy learning.

\subsection{VLA Policy Learning with Domain-Invariant Latent Lookahead}
\label{sec:latent_lookahead_policy}

During VLA policy training, the pretrained Task-Domain Encoder provides future latent supervision. Given the current observation \(o_t\) and language instruction \(\ell\), the vision-language backbone~\citep{bai2025qwen25vltechnicalreport} produces image-language tokens:
\begin{equation}
    H_t^{\mathrm{vlm}} = M_{\theta}(o_t,\ell).
\end{equation}
Two policy heads map these tokens to a predicted lookahead latent and a current representation:
\begin{equation}
    \hat z_t^{\mathrm{look}}
    =
    P_{\theta}^{\mathrm{look}}(H_t^{\mathrm{vlm}}),
    \qquad
    z_t^{\mathrm{curr}}
    =
    P_{\theta}^{\mathrm{curr}}(H_t^{\mathrm{vlm}}).
\end{equation}

Let \(\mathbf{o}^{+}_t=(o_{t+1},\ldots,o_{t+T_v})\) denote the future observation chunk.
Using the pretrained Task-Domain Encoder, we extract a future task target and its domain latent:
\begin{equation}
    z_t^{\mathrm{task},\star}
    =
    E_{\bar\psi}^{\mathrm{task}}
    \left(
    \mathbf{o}^{+}_t
    \right),
    \qquad
    z_t^{\mathrm{dom},\star}
    =
    E_{\bar\xi}^{\mathrm{dom}}
    \left(
    \mathbf{o}^{+}_t
    \right)
    .
\end{equation}

The lookahead latent is aligned with the future task target using the InfoNCE objective in Eq.~\ref{eq:nce}.
For each minibatch, we instantiate Eq.~\ref{eq:nce} by using the predicted lookahead latents as anchors,
\(z_i=\hat z_i^{\mathrm{look}}\), and the stop-gradient future task latents as positives,
\(z_i^+=\mathrm{sg}(z_i^{\mathrm{task},\star})\).
We denote this loss by \(\mathcal L_{\mathrm{look}}\).

We also disentangle the current representation from the domain latent extracted by the domain encoder:
\begin{equation}
    u_i^{\mathrm{curr}}
    =
    \left[
    z_i^{\mathrm{curr}};
    \mathrm{sg}
    \left(
    z_i^{\mathrm{dom},\star}
    \right)
    \right],
    \qquad
    \mathcal L_{\mathrm{dis}}^{\mathrm{curr}}
    =
    \mathcal R_{\mathrm{dis}}
    \left(
    \{u_i^{\mathrm{curr}}\}_{i=1}^{B}
    \right).
\end{equation}
Since the current and future chunks come from the same trajectory and domain condition, \(z_t^{\mathrm{dom},\star}\) provides the corresponding domain latent.
The Gaussian disentanglement loss encourages \(z_t^{\mathrm{curr}}\) to be separated from the domain latent while preserving information useful for action prediction.

The lookahead latent provides predictive task information about the future observation chunk, while the current representation provides action-relevant information from the present input after domain disentanglement:
\begin{equation}
    \hat a_{t:t+K_a-1}
    =
    A_{\phi}
    \left(
    [\hat z_t^{\mathrm{look}};z_t^{\mathrm{curr}}]
    \right).
\end{equation}
We supervise the action chunk with behavior cloning:
\begin{equation}
    \mathcal L_{\mathrm{act}}
    =
    \frac{1}{K_a}
    \sum_{k=0}^{K_a-1}
    \left\|
    \hat a_{t+k}
    -
    a_{t+k}
    \right\|_1.
\end{equation}
The final policy objective is
\begin{equation}
    \mathcal L_{\mathrm{policy}}
    =
    \lambda_{\mathrm{act}}\mathcal L_{\mathrm{act}}
    +
    \lambda_{\mathrm{look}}\mathcal L_{\mathrm{look}}
    +
    \lambda_{\mathrm{dis}}^{\mathrm{curr}}
    \mathcal L_{\mathrm{dis}}^{\mathrm{curr}}.
\end{equation}

\paragraph{Policy pretraining and adaptation.}
We pretrain the policy-side modules on action-labeled trajectories from the source trajectory collection.
For downstream settings such as LIBERO~\citep{liu2023libero} or real-world robot data, we fine-tune the policy with the same lookahead, disentanglement, and action losses.
At deployment, the policy receives only the current observation and instruction.
The Task-Domain Encoder is not used at inference.

\section{Experiments}
\label{sec:experiments}

We evaluate the central claim of this paper: domain-invariant latent lookahead mitigates shortcut learning in vision-language-action policies. We combine controlled simulation, where task--domain correlations and visual shifts can be systematically manipulated, with real-world manipulation experiments. Specifically, we ask:

\noindent\textbf{1.} Does our approach reduce shortcut reliance under counterfactual task--view compositions?

\noindent\textbf{2.} Does our approach improve robustness under visual distribution shifts?

\noindent\textbf{3.} Do the learned latents successfully separate task structure from task-irrelevant domain factors?

\subsection{LIBERO Shortcut Diagnostic Under Counterfactual Task--View Compositions}
\label{sec:libero_shortcut}

We first test shortcut mitigation in a controlled LIBERO diagnostic~\cite{shortcut_learning_in_GRPs}. This diagnostic intentionally creates a spurious correlation between task identity and camera viewpoint during training, then breaks this correlation at test time to measure whether a policy follows the commanded task or the task spuriously associated with the observed view.

{\textbf{Protocol.}
During training, two task groups, Task-L and Task-R, are observed only from left- and right-view ranges, respectively. At test time, we swap these associations: Task-R is evaluated at the left-view boundary and Task-L at the right-view boundary. Details are in Appendix~\ref{app:libero_shortcut}.\par}

\textbf{Metrics.}
We report OOD success and shortcut degree. OOD success measures whether the commanded task is completed under the counterfactual view. Shortcut degree measures whether the policy instead executes the task group spuriously associated with the observed view during training. Lower shortcut degree indicates less shortcut reliance.

\begin{wrapfigure}{r}{0.6\linewidth}
\vspace{-0.65\baselineskip}
\centering
\resizebox{\linewidth}{!}{%
\begin{tikzpicture}[x=1pt,y=1pt,font=\scriptsize]
\definecolor{crbaseline}{rgb}{0.52,0.55,0.59}
\definecolor{crentangled}{rgb}{0.80,0.47,0.20}
\definecolor{crnohead}{rgb}{0.44,0.64,0.79}
\definecolor{crdill}{rgb}{0.13,0.35,0.53}
\path[use as bounding box] (0,-13) rectangle (244,73);
\node[anchor=east,font=\scriptsize\bfseries] at (61,68) {Model};
\node[font=\scriptsize\bfseries] at (101,68) {Shortcut degree $\downarrow$};
\node[font=\scriptsize\bfseries] at (199,68) {OOD success $\uparrow$};
\foreach \start in {70,169} {
  \foreach \tick/\label in {0/0.0,30/0.5,60/1.0} {
    \draw[black!40,line width=0.3pt] (\start+\tick,-1) -- (\start+\tick,-3);
    \node[anchor=north,inner sep=1pt,text=black!65] at (\start+\tick,-4) {\label};
  }
  \draw[black!40,line width=0.3pt] (\start,-1) -- (\start+60,-1);
}
\newcommand{\crdiagrow}[6]{%
  \node[anchor=east,inner sep=1pt,text=#6] at (61,#2) {#1};
  \fill[#5] (70,#2-2.6) rectangle ({70+60*#3},#2+2.6);
  \fill[#5] (169,#2-2.6) rectangle ({169+60*#4},#2+2.6);
  \node[anchor=west,inner sep=1pt,text=#6] at ({72+60*#3},#2) {#3};
  \node[anchor=west,inner sep=1pt,text=#6] at ({171+60*#4},#2) {#4};
}
\crdiagrow{MiniVLA}{55}{1.00}{0.00}{crbaseline}{black}
\crdiagrow{$\pi_0$}{45}{0.60}{0.00}{crbaseline}{black}
\crdiagrow{Base VLA}{35}{0.73}{0.00}{crbaseline}{black}
\crdiagrow{Entangled LA}{25}{0.65}{0.00}{crentangled}{black}
\crdiagrow{DILL w/o CH}{15}{0.06}{0.44}{crnohead}{black}
\crdiagrow{DILL}{5}{0.05}{0.58}{crdill}{black}
\end{tikzpicture}%
}
\vspace{-0.65em}
\caption{\textbf{LIBERO shortcut diagnostic.}}
\label{fig:libero_shortcut_bar}
\vspace{-0.65em}
\end{wrapfigure}
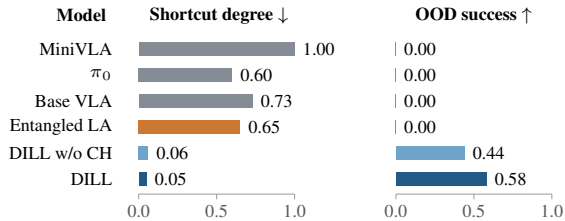

\textbf{Compared methods.}
\emph{Base VLA} is a behavior-cloning baseline built from DILL's underlying VLA. It maps the current observation and instruction to actions and is trained only with action supervision, without source augmentation, latent lookahead, or task--domain supervision. We then compare two lookahead variants. \emph{Entangled latent lookahead (Entangled LA)} predicts future representations from the original video encoder, which does not separate task and domain factors. \emph{DILL w/o CH} instead predicts disentangled future task latents, but omits the disentangled current head (CH). These two variants share policy architecture and augmented source data, isolating the choice of predictive target. Full DILL additionally disentangles the current representation used for action prediction.\par

\textbf{Results.}
Figure~\ref{fig:libero_shortcut_bar} shows that Base VLA fails to complete the commanded task under swapped views (zero OOD success), while frequently executing the task associated with the observed view (shortcut degree $0.73$). Thus, its failures reflect task substitution, not just difficulty acting from an unfamiliar view. MiniVLA and $\pi_0$ exhibit the same pattern in their respective evaluations. DILL raises OOD success to $0.58$ and reduces shortcut degree to $0.05$, recovering commanded behavior while largely avoiding view-induced task substitution.\par

The target ablation shows why future prediction alone is insufficient in this setting. Entangled LA still has zero OOD success and a shortcut degree of $0.65$. Replacing its predictive targets with disentangled future task latents (DILL w/o CH) raises success to $0.44$ and reduces shortcut degree to $0.06$. With architecture and augmentation exposure matched, this contrast supports the importance of domain-invariant targets beyond future prediction alone. Disentangling the current representation in full DILL further raises success from $0.44$ to $0.58$, with little change in shortcut degree ($0.06$ to $0.05$). Its additional benefit is therefore better execution of the commanded task, beyond the shortcut reduction already achieved by invariant lookahead targets. Additional component ablations appear in Appendix~\ref{app:libero_shortcut_ablations}.\par

\providecommand{\snum}[1]{{\footnotesize #1}}

\begin{table*}[!t]
\centering
\small
\setlength{\tabcolsep}{3.0pt}
\renewcommand{\arraystretch}{0.90}
\begin{threeparttable}
\begin{tabularx}{\linewidth}{
@{}
>{\raggedright\arraybackslash}p{0.24\linewidth}
*{6}{>{\centering\arraybackslash}X}
@{}}
\toprule
\multirow{2}{*}{Model}
& \multirow{2}{*}{Original}
& \multicolumn{4}{c}{Visual perturbations}
& \multirow{2}{*}{Average} \\
\cmidrule(lr){3-6}
& & Camera & Light & BG & Noise & \\
\midrule

\multirow{2}{*}{OpenVLA~\cite{openvla}}
  & \multirow{2}{*}{\snum{76.5}}
  & \snum{0.8} & \snum{8.1} & \snum{34.8} & \snum{15.2} & \snum{14.7} \\
  &
  & \snum{\dropcell{75.7}} & \snum{\dropcell{68.4}} & \snum{\dropcell{41.7}} & \snum{\dropcell{61.3}} & \snum{\dropcell{61.8}} \\

\multirow{2}{*}{WorldVLA~\cite{worldvla}}
  & \multirow{2}{*}{\snum{79.1}}
  & \snum{0.1} & \snum{43.7} & \snum{17.1} & \snum{10.9} & \snum{18.0} \\
  &
  & \snum{\dropcell{79.0}} & \snum{\dropcell{35.4}} & \snum{\dropcell{62.0}} & \snum{\dropcell{68.2}} & \snum{\dropcell{61.2}} \\

\multirow{2}{*}{UniVLA~\cite{univla}}
  & \multirow{2}{*}{\snum{95.5}}
  & \snum{1.8} & \snum{69.0} & \snum{81.0} & \snum{21.2} & \snum{43.3} \\
  &
  & \snum{\dropcell{93.7}} & \snum{\dropcell{26.5}} & \snum{\dropcell{14.5}} & \snum{\dropcell{74.3}} & \snum{\dropcell{52.3}} \\

\multirow{2}{*}{NORA~\cite{nora}}
  & \multirow{2}{*}{\snum{87.9}}
  & \snum{2.2} & \snum{45.7} & \snum{58.6} & \snum{12.8} & \snum{29.8} \\
  &
  & \snum{\dropcell{85.7}} & \snum{\dropcell{42.2}} & \snum{\dropcell{29.3}} & \snum{\dropcell{75.1}} & \snum{\dropcell{58.1}} \\

\multirow{2}{*}{\mbox{OpenVLA-OFT~\cite{oft}}}
  & \multirow{2}{*}{\snum{95.3}}
  & \snum{10.4} & \snum{76.8} & \snum{93.6} & \snum{49.9} & \snum{57.7} \\
  &
  & \snum{\dropcell{84.9}} & \snum{\dropcell{18.5}} & \snum{\dropcell{1.7}} & \snum{\dropcell{45.4}} & \snum{\dropcell{37.6}} \\

\midrule

\multirow{2}{*}{{Base VLA + SA}}
  & \multirow{2}{*}{{\snum{82.0}}}
  & {\snum{39.3}} & {\snum{56.8}} & {\snum{50.5}} & {\snum{6.3}} & {\snum{38.2}} \\
  &
  & {\snum{\dropcell{42.7}}} & {\snum{\dropcell{25.2}}} & {\snum{\dropcell{31.5}}} & {\snum{\dropcell{75.7}}} & {\snum{\dropcell{43.8}}} \\

\multirow{2}{*}{\textbf{DILL (Ours)}}
  & \multirow{2}{*}{\snum{81.6}}
  & \snum{68.4} & \snum{69.4} & \snum{70.0} & \snum{68.7} & \snum{69.1} \\
  &
  & \snum{\dropcell{13.2}} & \snum{\dropcell{12.2}} & \snum{\dropcell{11.6}} & \snum{\dropcell{12.9}} & \snum{\dropcell{12.5}} \\

\bottomrule
\end{tabularx}

\end{threeparttable}

\vspace{-0.35em}
\caption{\textbf{Zero-shot robustness evaluation on visual perturbations in LIBERO-Plus~\cite{libero_plus}.} {For each model, the first row reports success (\%) and the second its drop from Original in percentage points. Average is the unweighted mean across the four visual categories.} {The bottom block matches policy architecture and augmented source data (SA: source augmentation).} %
}
\label{tab:liberoplus_visual_only}
\vspace{-0.5em}
\end{table*}

\subsection{LIBERO-Plus Zero-Shot Robustness Evaluation Under Visual Distribution Shifts}
\label{sec:liberoplus}

We next evaluate whether shortcut mitigation translates into stronger robustness under visual distribution shifts. 
We use the visual perturbation categories in LIBERO-Plus~\cite{libero_plus} as a controlled zero-shot evaluation: no LIBERO-Plus perturbed images are used for training. All models are adapted on the original LIBERO training split.

\textbf{Compared methods.}
To ensure a controlled comparison, the main table includes only methods evaluated with third-person RGB observations, excluding wrist-camera images. Under this matched-input protocol, we compare against OpenVLA~\cite{openvla}, OpenVLA-OFT~\cite{oft}, WorldVLA~\cite{worldvla}, UniVLA~\cite{univla}, and NORA~\cite{nora}.

\textbf{Results.}
DILL achieves the highest average success across the four visual perturbation categories ($69.1\%$; Table~\ref{tab:liberoplus_visual_only}), exceeding the strongest external baseline, OpenVLA-OFT, by $11.4$ percentage points. This advantage does not come from higher original LIBERO performance: OpenVLA-OFT and UniVLA score higher without perturbations, but their average drops under visual shifts are $37.6$ and $52.3$ points, respectively, compared with $12.5$ for DILL.
DILL's largest advantages are under camera changes ($68.4\%$ success) and sensor noise ($68.7\%$), where all external baselines remain below $11\%$ and $50\%$, respectively. OpenVLA-OFT is stronger on background and lighting changes, but DILL maintains $68.4$--$70.0\%$ success across all four categories, indicating more consistent robustness across visual shifts.
To test whether augmentation exposure alone accounts for this robustness, we train Base VLA with DILL's augmented source data while retaining the behavior-cloning objective (Base VLA + SA). This control nearly matches DILL on original LIBERO ($82.0\%$ versus $81.6\%$), yet its average perturbed success is much lower ($38.2\%$ versus $69.1\%$). DILL outperforms this control in all four categories, indicating that augmentation exposure alone does not explain its robustness gains. The full LIBERO-Plus breakdown is in Appendix~\ref{app:libero_plus_full}.\par

\begin{figure*}[!t]
\centering
\setlength{\tabcolsep}{0pt}

\begin{tabular}{@{}ccc@{}}
\makebox[0.326\textwidth][c]{\footnotesize \textbf{Task Encoder}} &
\makebox[0.326\textwidth][c]{\footnotesize \textbf{Domain Encoder}} &
\makebox[0.326\textwidth][c]{\footnotesize \textbf{Final VLA Latent}}
\end{tabular}

\vspace{0.15em}
\includegraphics[width=0.98\textwidth]{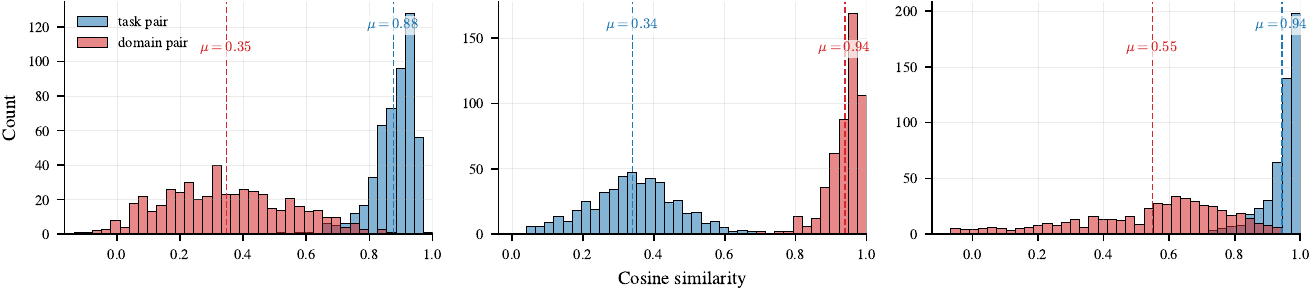}
\vspace{-0.4em}

\caption{\textbf{Pairwise similarity diagnostics for learned latents.}
We compare cosine-similarity distributions for task pairs and domain pairs constructed from unseen task--domain combinations. From left to right, the panels show the task encoder, domain encoder, and final VLA latent. The task encoder should group task pairs, the domain encoder should group domain pairs, and the final VLA latent should preserve task-consistent structure while suppressing domain-specific variation.}
\label{fig:vla_latent_similarity_main}
\vspace{-0.6em}
\end{figure*}

\subsection{Pairwise Diagnostics of Learned Latents}
\label{sec:repr}

{We also analyze} whether the learned latent spaces separate task-relevant structure from domain-specific visual variation. We evaluate three representations: the task encoder output, the domain encoder output, and the final VLA latent provided to the action head. For the final VLA latent, we use the concatenated action-conditioning representation, \textit{i.e.}, the current policy representation together with the predicted lookahead latent.

\textbf{Protocol.}
We compute cosine similarity between normalized latents for two types of unseen task and domain pairs. \emph{Task pairs} share the same underlying trajectory content but differ in visual domain or augmentation. \emph{Domain pairs} share the same visual domain or augmentation pattern but differ in trajectory content. These pair combinations are not observed during training, so the diagnostic tests whether the learned representations generalize beyond memorized pairings. A task-centric representation should assign higher similarity to task pairs than to domain pairs, while a domain-centric representation should show the opposite behavior. Additional details are provided in Appendix~\ref{app:representation_diagnostics}.

\textbf{Results.}
Figure~\ref{fig:vla_latent_similarity_main} shows that the learned factorization behaves as intended. The task encoder assigns high similarity to task pairs and substantially lower similarity to domain pairs, with mean similarities of $0.88$ and $0.35$, respectively. Conversely, the domain encoder assigns high similarity to domain pairs and lower similarity to task pairs, with mean similarities of $0.94$ and $0.34$, indicating that the task and domain encoders capture complementary factors rather than collapsing to the same representation. The final VLA latent also remains strongly task-centric: it assigns high similarity to task pairs ($\mu=0.94$) while keeping domain pairs noticeably lower ($\mu=0.55$). Since this is the representation directly provided to the action head, the result suggests that the policy is conditioned on latent features that are stable across domain changes but still discriminative across different trajectory content. This supports the mechanism behind the robustness gains in Table~\ref{tab:liberoplus_visual_only}: domain-invariant latent lookahead encourages the policy to act on task-consistent structure rather than domain-specific visual cues.

\begingroup

\subsection{Real-World Experiments}
\label{sec:real_world}

{To test whether DILL's benefits extend to the physical world, we use two tasks for shortcut diagnosis and three for visual robustness. {The shortcut diagnostic swaps target-color/viewpoint associations learned during training to test instruction following when visual cues become misleading.} We compare against Base VLA without task--domain supervision and DILL-Current, which replaces the future task target with the current task latent. Policies are fine-tuned on real-world demonstrations, while the source-pretrained Task-Domain Encoder remains frozen, without adaptation or paired-view training on these tasks. Appendix~\ref{app:real_world_protocol} details both protocols, illustrated in Figures~\ref{fig:real_world_counterfactual_examples} and~\ref{fig:real_world_robustness_examples}.\par}

\begin{table}[!t]
\centering

\small
\setlength{\tabcolsep}{1pt}
\renewcommand{\arraystretch}{1.05}
\begin{tabularx}{\linewidth}{@{}l *{5}{>{\centering\arraybackslash}X}@{}}
\toprule
& \multicolumn{2}{c}{Shortcut diagnosis} & \multicolumn{3}{c}{Robustness} \\
\cmidrule(lr){2-3} \cmidrule(lr){4-6}
Model & CTF $\uparrow$ & SD $\downarrow$ & No pert. $\uparrow$ & Predefined $\uparrow$ & New $\uparrow$ \\
\midrule
Base VLA & $29.7{\scriptstyle\,\pm5.4}$ & $47.2{\scriptstyle\,\pm4.8}$ & $84.1{\scriptstyle\,\pm5.5}$ & $48.3{\scriptstyle\,\pm5.9}$ & $64.0{\scriptstyle\,\pm6.4}$ \\
DILL-Current & $\mathbf{83.3}{\scriptstyle\,\pm4.8}$ & $\mathbf{0.0}{\scriptstyle\,\pm0.0}$ & $\mathbf{88.9}{\scriptstyle\,\pm2.7}$ & $78.2{\scriptstyle\,\pm7.6}$ & $77.2{\scriptstyle\,\pm5.1}$ \\
\textbf{DILL} & $80.2{\scriptstyle\,\pm5.5}$ & $\mathbf{0.0}{\scriptstyle\,\pm0.0}$ & $\mathbf{88.9}{\scriptstyle\,\pm2.7}$ & $\mathbf{83.8}{\scriptstyle\,\pm4.2}$ & $\mathbf{82.0}{\scriptstyle\,\pm4.0}$ \\
\bottomrule
\end{tabularx}
\caption{\textbf{Real-world shortcut diagnosis and robustness (\%).} CTF measures command-following under swapped target-color/viewpoint pairings; SD is shortcut degree (Section~\ref{sec:libero_shortcut}). Predefined perturbations are held-out instances of encoder-training transformation families; new perturbations (cast shadows, dynamic backgrounds, and foreground clutter) are absent from pair construction.}
\label{tab:real_world_main}
\vspace{-0.5em}
\end{table}

{DILL raises counterfactual command-following success from Base VLA's $29.7\%$ to $80.2\%$ and reduces observed shortcut degree from $47.2\%$ to zero (Table~\ref{tab:real_world_main}). DILL-Current also attains $83.3\%$ success with no observed shortcut execution, indicating that task-invariant supervision can suppress this failure mode without future prediction.\par}

{Future targets provide an additional benefit under visual shifts. At identical $88.9\%$ unperturbed success, DILL exceeds DILL-Current by $5.6$ and $4.8$ percentage points in mean success under predefined and new perturbations, respectively. The latter are absent from encoder pair construction, indicating that lookahead's robustness gains extend beyond the transformation families used to learn invariance.\par}
\endgroup

\section{Limitations}

DILL is designed to reduce shortcut reliance induced by visual-domain factors. 
In our experiments, these factors include changes in viewpoint, scene appearance, camera configuration, and visual degradation. 
While this captures a common source of brittleness in VLA policies, it does not cover all possible spurious correlations. 
For example, biases in initial states, object layouts, language templates, task frequencies, or demonstrator styles may also influence action prediction without being part of the intended task semantics. 
Addressing such non-visual shortcuts would require defining additional nuisance factors, obtaining corresponding paired interventions, or developing objectives that can discover them more automatically.

Another important consideration is the trade-off between robustness and in-distribution performance. 
In the training distribution, domain-dependent cues can be highly predictive of demonstrated actions, even when they are not causally tied to the intended task semantics. 
Moreover, the same visual factor may play different roles across contexts: a background pattern may be a spurious cue in one setting, but part of the task-relevant scene configuration in another. 
Overly strong invariance may therefore suppress information that is useful for control in some situations, reducing peak in-distribution performance even as it improves robustness when those cues become misleading. 
This suggests that robustness should not be pursued by uniformly removing domain-specific information in all cases. 
Instead, future work should explore adaptive objectives that preserve action-relevant factors and suppress action-irrelevant shortcuts in a context-dependent manner.

\section{Conclusion}

A central obstacle to robust VLA control is that policies can treat domain-specific appearance cues as action-relevant when they are only reliable within the training distribution.
Domain-Invariant Latent Lookahead addresses this shortcut learning problem by supervising VLA representations with future latents in a space where domain-specific visual variation has been separated from task-relevant structure.
This encourages the policy to rely less on incidental appearance factors such as background, viewpoint, or lighting, and more on action-relevant scene information that remains stable across domains.
{Controlled simulation studies show reduced shortcut reliance and improved robustness to visual shifts, while experiments on a physical robot provide evidence of these benefits beyond simulation.}
Our results suggest that improving robustness in VLA models is not only a matter of scaling data or architectures, but also of shaping what the policy learns.

\clearpage

\acknowledgments{
This work was partly supported by grants funded by the Korean government through IITP
(RS-2022-II220951-LBA/5\%,
RS-2022-II220953-PICA/5\%,
RS-2026-25553157-MIACC/10\%,
IITP-2026-RS-2023-00255968/10\%,
RS-2026-25617480/10\%,
and RS-2026-25552043/10\%),
NRF
(RS-2024-00353991-SPARC/10\%,
RS-2023-00274280-HEI/10\%,
and RS-2026-25518808/10\%),
KEIT (RS-2025-25453780/10\%),
and KIAT (RS-2025-25460896/10\%).
}

\bibliography{example}  %

\clearpage

\appendix

\clearpage
\section*{Appendix Overview}
\label{app:overview}

This appendix provides additional details, analyses, and supplementary experiments for DILL. 
Appendix~\ref{app:method_details} describes the method implementation and training objectives.
Appendix~\ref{app:simulation_benchmark_experiments} provides simulation benchmark details, including the LIBERO shortcut diagnostic, ablations, and the full LIBERO-Plus breakdown.
Appendix~\ref{app:representation_diagnostics} provides additional representation diagnostics.
{Appendix~\ref{app:real_world_protocol} describes the real-world evaluation protocol.}

\vspace{0.5em}
\noindent\textbf{Contents.}
\begin{itemize}
    \item Appendix~\ref{app:method_details}: Method details and training objectives.
    \item Appendix~\ref{app:simulation_benchmark_experiments}: Simulation benchmark experiments.
    \item Appendix~\ref{app:representation_diagnostics}: Representation diagnostics.
    \item {Appendix~\ref{app:real_world_protocol}: Real-world evaluation protocol.}
\end{itemize}

\appendix

\section*{APPENDIX}
\addcontentsline{toc}{section}{Appendix}

\section{Additional Method Details}
\label{app:method_details}

\subsection{Domain Transformations and Positive Pair Mining}
\label{app:domain_pairs}

This subsection supports Sec.~\ref{sec:task_domain_disentanglement} by detailing how we construct the domain-transformed positive pairs to train the Task-Domain Encoder.
A domain condition \(\eta\) specifies the full transformation condition, including the transformation family and its sampled parameters or random seed.
Task-positive pairs vary the domain condition while preserving the same underlying observation chunk, whereas domain-positive pairs preserve the domain condition across different observation chunks.

\paragraph{Data sources.}
Task-domain encoder pretraining uses a source trajectory collection composed of MimicGen~\citep{mandlekar2023mimicgen}, ManiSkill~\citep{tao2024maniskill3}, and subsets of OXE~\citep{oxe}.
The OXE subset includes Bridge and Fractal/RT-1 trajectories.
Table~\ref{tab:source_collection_summary} summarizes the pretraining data.
In our main experiments, the Task-Domain Encoder is pretrained once on this source trajectory collection and then kept fixed during VLA policy pretraining and downstream policy learning.
This design decouples Task-Domain Encoder pretraining from downstream policy learning: downstream users can train only the policy-side modules unless they choose to further adapt the Task-Domain Encoder with additional domain transformations.

\begin{table}[h]
    \centering
    \footnotesize
    \caption{
    Source trajectory collection used for Task-Domain Encoder pretraining.
    The MimicGen row aggregates 16 task datasets, while OXE aggregates Bridge and Fractal/RT-1 subsets.
    }
    \label{tab:source_collection_summary}
    \begin{tabularx}{\linewidth}{@{}>{\raggedright\arraybackslash}X
                                  >{\raggedleft\arraybackslash}p{0.20\linewidth}
                                  >{\raggedleft\arraybackslash}p{0.20\linewidth}
                                  >{\raggedleft\arraybackslash}p{0.16\linewidth}@{}}
        \toprule
        Source & Frames & Trajectories & Tasks \\
        \midrule
        MimicGen, 16 datasets
        & 1,010,618 & 3,200 & 16 \\
        ManiSkill, 2 datasets
        & 1,632,291 & 14,057 & 17 \\
        OXE, Bridge + Fractal/RT-1
        & 5,785,810 & 140,404 & 20,553 \\
        \midrule
        Total
        & 8,428,719 & 157,661 & 20,586 \\
        \bottomrule
    \end{tabularx}
\end{table}

For simulation data, controllable rendering enables viewpoint changes and, when segmentation masks are available, mask-based background replacement.
We additionally apply image-level transformations directly to frames; these transformations are used for both simulation data and recorded trajectories such as the OXE subsets.

\begin{figure}[h]
    \centering
    \setlength{\tabcolsep}{3pt}
    \renewcommand{\arraystretch}{1.0}
    \begin{tabularx}{\linewidth}{@{}*{5}{>{\centering\arraybackslash}X}@{}}
        \includegraphics[width=\linewidth]{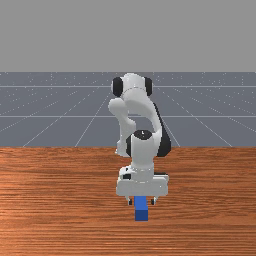}
        &
        \includegraphics[width=\linewidth]{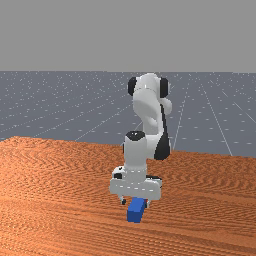}
        &
        \includegraphics[width=\linewidth]{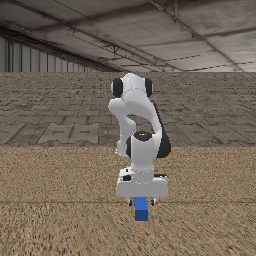}
        &
        \includegraphics[width=\linewidth]{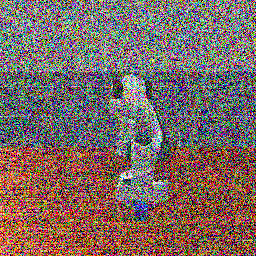}
        &
        \includegraphics[width=\linewidth]{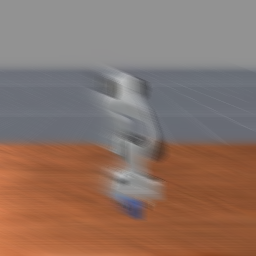}
        \\[3pt]
        {\scriptsize \shortstack[c]{Anchor}}
        &
        {\scriptsize \shortstack[c]{Viewpoint\\Changes}}
        &
        {\scriptsize \shortstack[c]{Environment\\Appearance}}
        &
        {\scriptsize \shortstack[c]{Camera\\Heterogeneity}}
        &
        {\scriptsize \shortstack[c]{Visual\\Degradation}}
    \end{tabularx}
    \caption{
    Representative single-family task-positive transformations.
    Each column after the anchor shows the same trajectory chunk under one family of domain-specific visual factors.
    }
    \label{fig:domain_pair_examples}
\end{figure}

\paragraph{Transformation families.}
We group domain transformations into four families that reflect common visual domain shifts: viewpoint changes, environment appearance variation, camera heterogeneity, and visual degradation.
A sampled domain condition can include a single transformation or a combination of transformations from these families.
Table~\ref{tab:domain_transform_families} summarizes the transformation types and the shared domain condition used for domain-positive pairs.
Figure~\ref{fig:domain_pair_examples} illustrates simple single-family task-positive examples; the training procedure samples a broader range of transformation types, parameters, and combinations.

\begin{table}[h]
    \centering
    \footnotesize
    \caption{
    Domain transformation families used for positive pair mining.
    A domain condition \(\eta\) consists of a transformation family and its sampled parameters or random seed. The last column lists the condition shared by domain-positive pairs.
    }
    \label{tab:domain_transform_families}
    \begin{tabularx}{\linewidth}{@{}>{\raggedright\arraybackslash}p{0.22\linewidth}
                                      >{\raggedright\arraybackslash}p{0.36\linewidth}
                                      >{\raggedright\arraybackslash}X@{}}
        \toprule
        Family & Transformations & Shared condition for domain positives \\
        \midrule
        Viewpoint changes
        & Rendered camera view, random crop, intrinsics change, radial distortion, warping
        & Camera ID or camera parameters; crop seed; focal scale; distortion and warp parameters \\
        \midrule
        Environment appearance
        & Background replacement, lighting changes, color changes
        & Background texture IDs; mask regions; replacement mode; lighting or color parameters \\
        \midrule
        Camera heterogeneity
        & Camera unprocessing, sensor noise, camera color response changes
        & Camera-pipeline seed; color matrices; gamma; ISO level; shot/read noise; channel gains \\
        \midrule
        Visual degradation
        & Blur, weather effects, compression, pixelation, image corruptions
        & Corruption type, severity, and random seed \\
        \bottomrule
    \end{tabularx}
\end{table}

\begin{figure}[h]
    \centering
    \setlength{\tabcolsep}{1.2pt}
    \renewcommand{\arraystretch}{0.9}

    \newcommand{\bgteximg}[1]{%
        \includegraphics[width=\linewidth,height=0.50in,keepaspectratio]{#1}%
    }

    \begin{tabularx}{\linewidth}{@{}*{10}{>{\centering\arraybackslash}X}@{}}

        \multicolumn{10}{@{}l@{}}{\footnotesize\textbf{Table textures}} \\
        \bgteximg{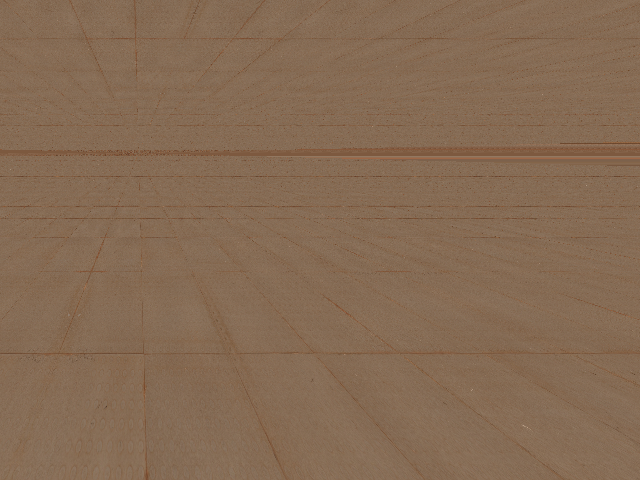} &
        \bgteximg{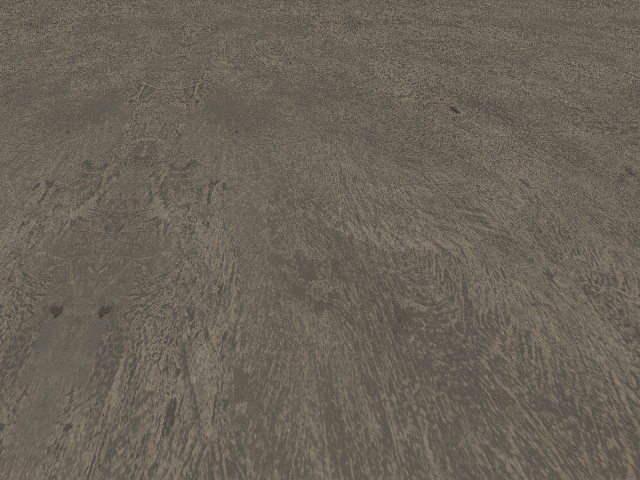} &
        \bgteximg{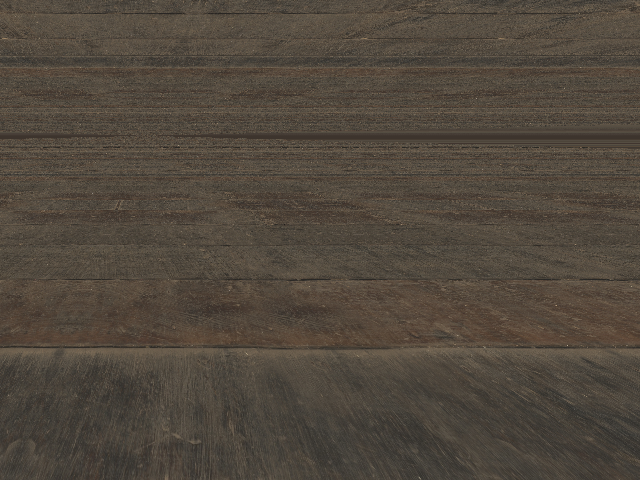} &
        \bgteximg{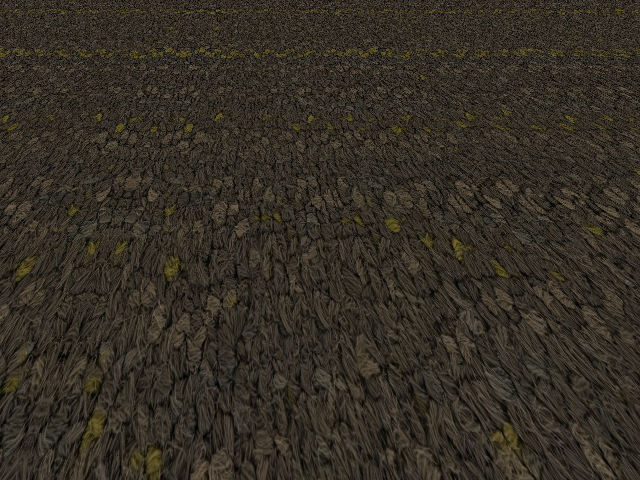} &
        \bgteximg{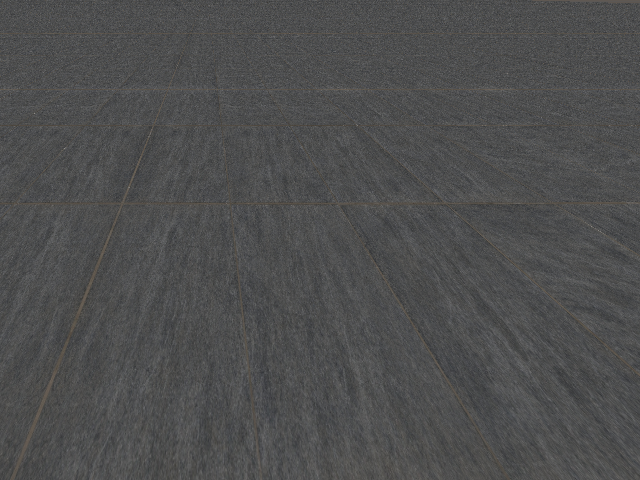} &
        \bgteximg{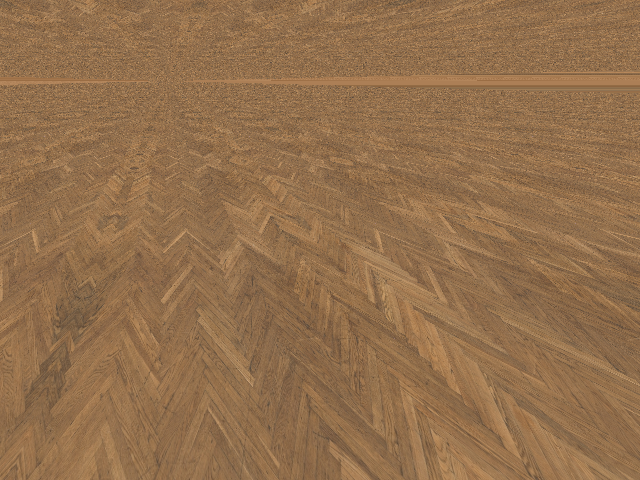} &
        \bgteximg{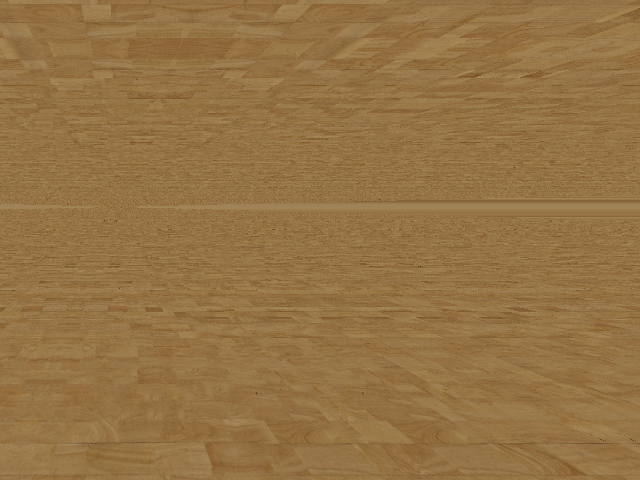} &
        \bgteximg{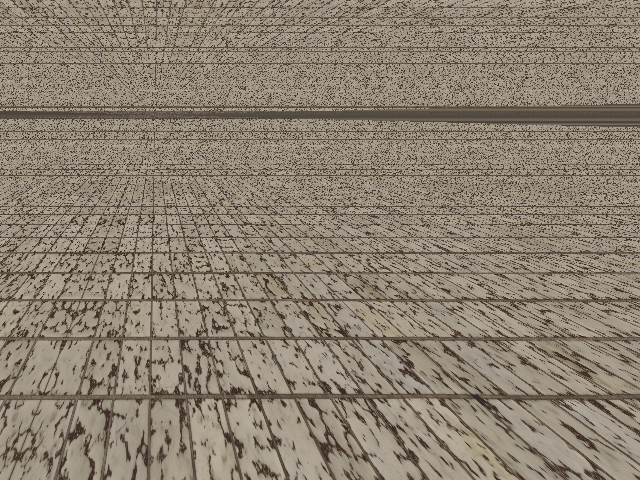} &
        \bgteximg{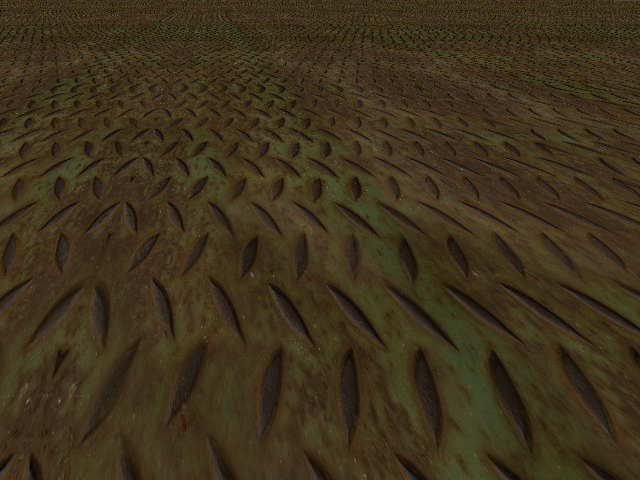} &
        \bgteximg{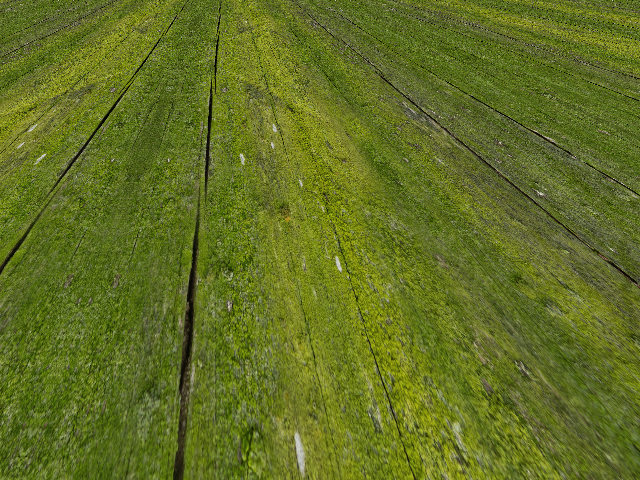}
        \\[3pt]

        \multicolumn{10}{@{}l@{}}{\footnotesize\textbf{Floor textures}} \\
        \bgteximg{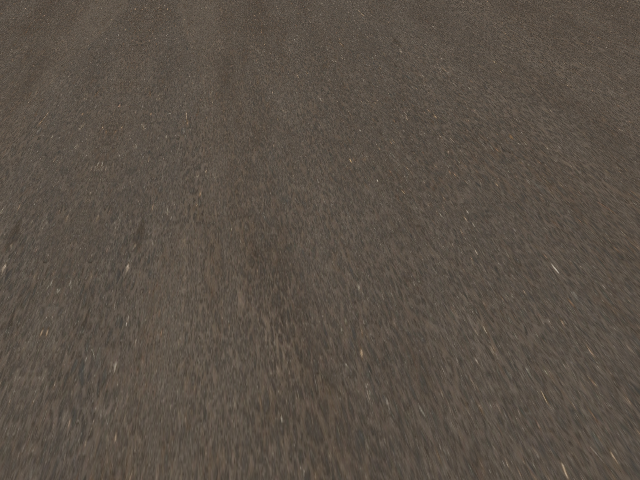} &
        \bgteximg{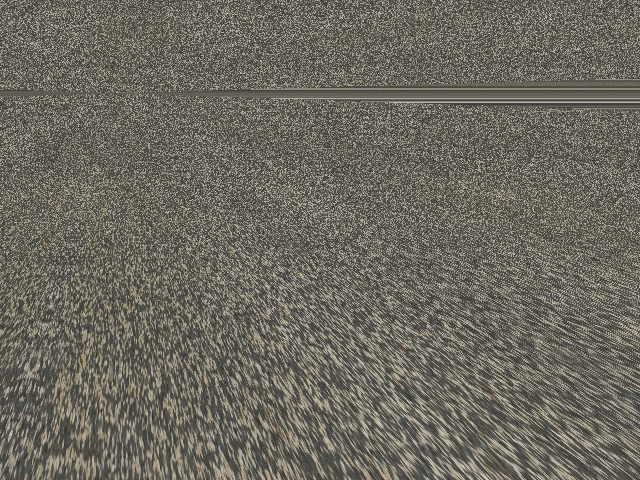} &
        \bgteximg{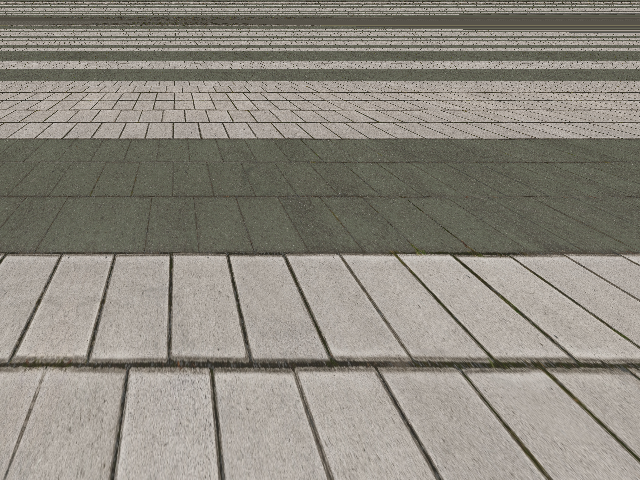} &
        \bgteximg{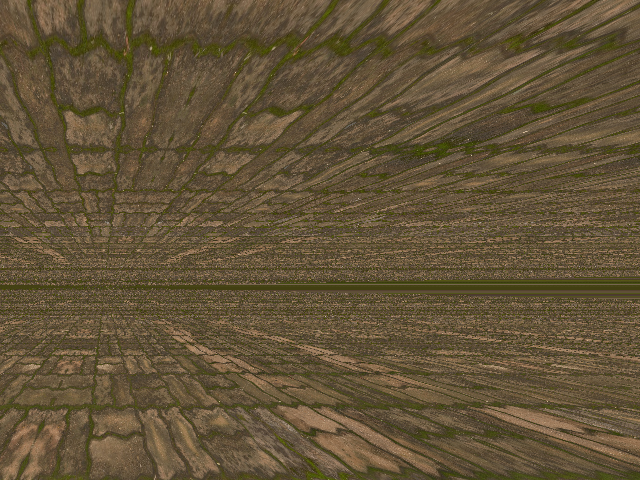} &
        \bgteximg{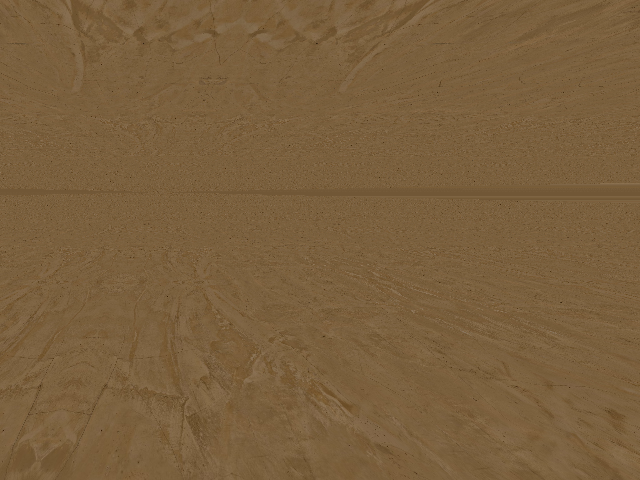} &
        \bgteximg{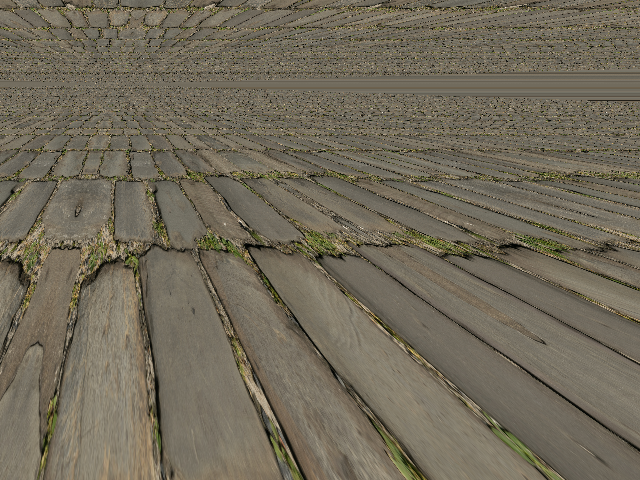} &
        \bgteximg{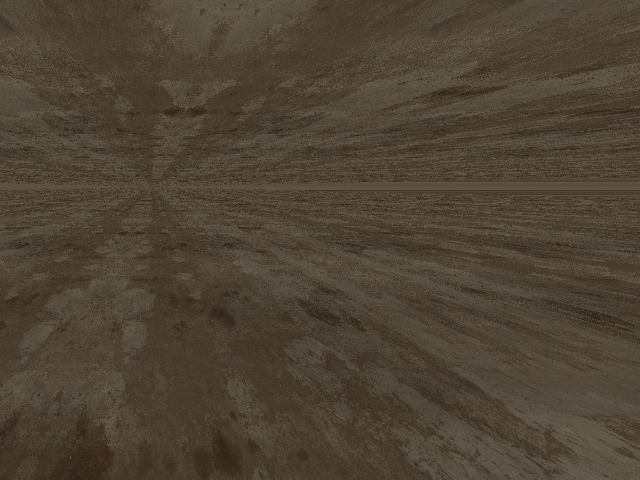} &
        \bgteximg{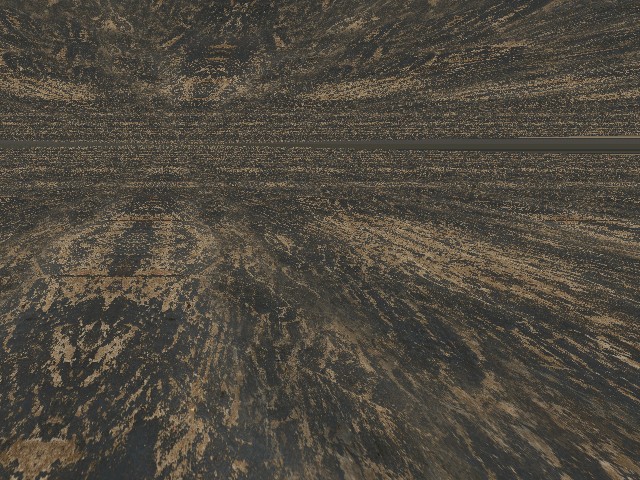} &
        \bgteximg{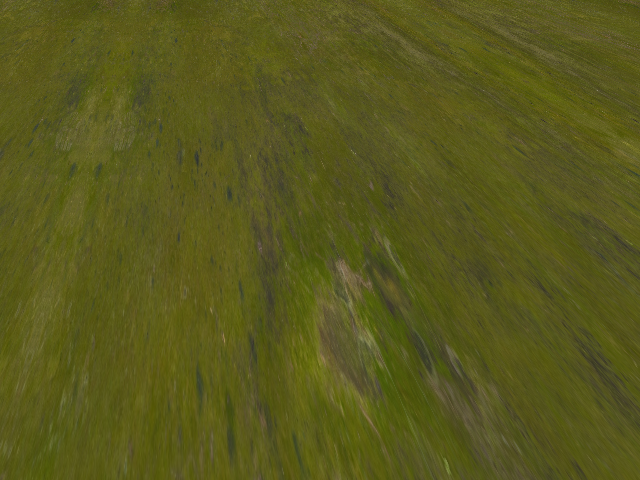} &
        \bgteximg{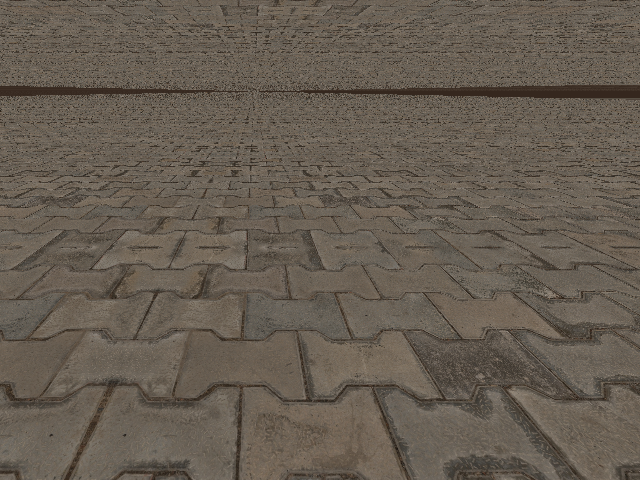}
        \\[3pt]

        \multicolumn{10}{@{}l@{}}{\footnotesize\textbf{Wall/background assets}} \\
        \bgteximg{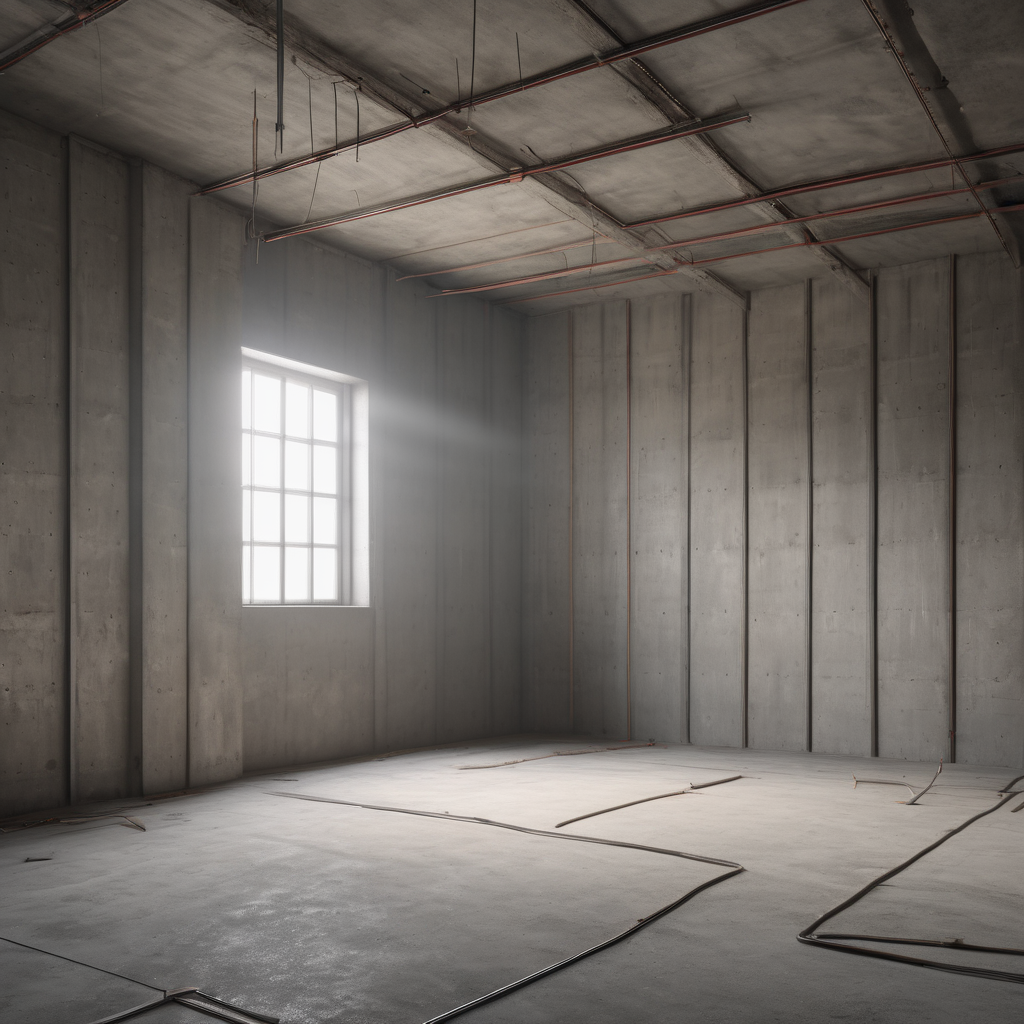} &
        \bgteximg{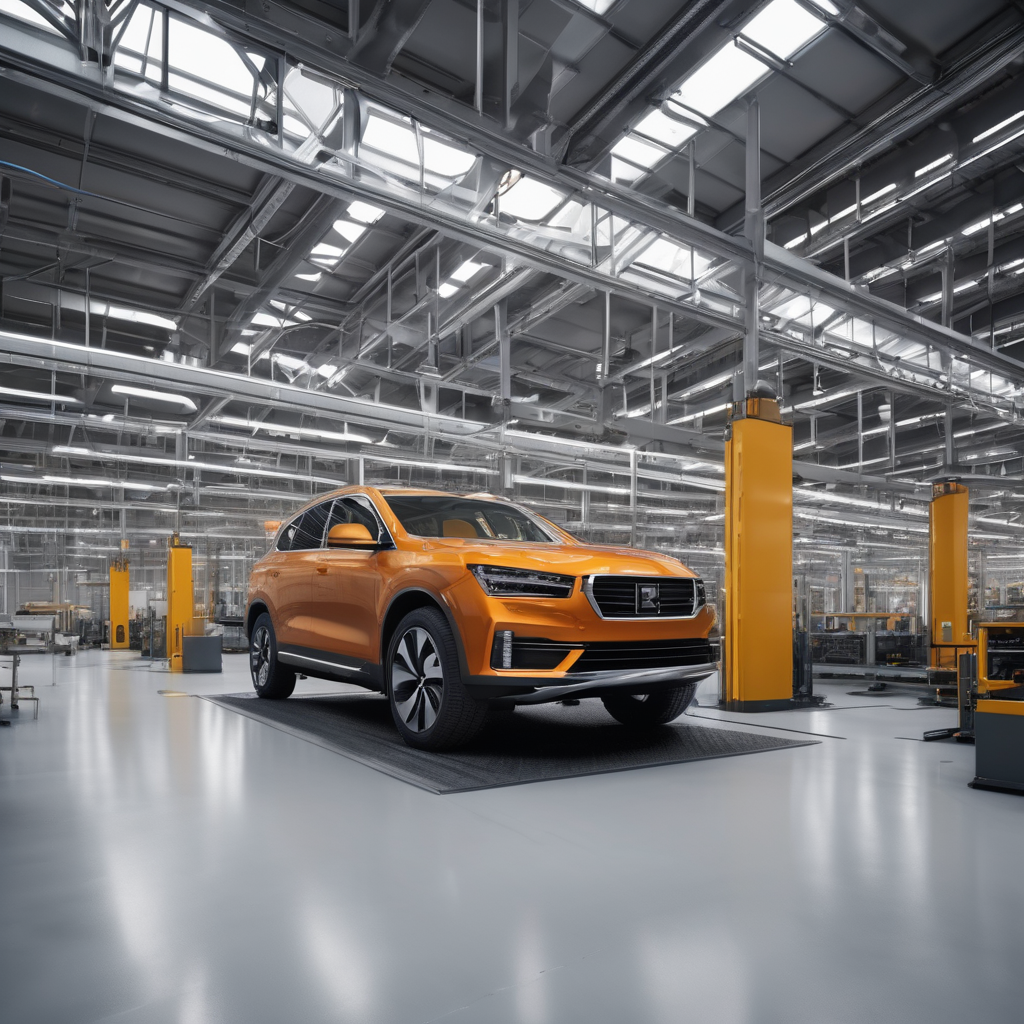} &
        \bgteximg{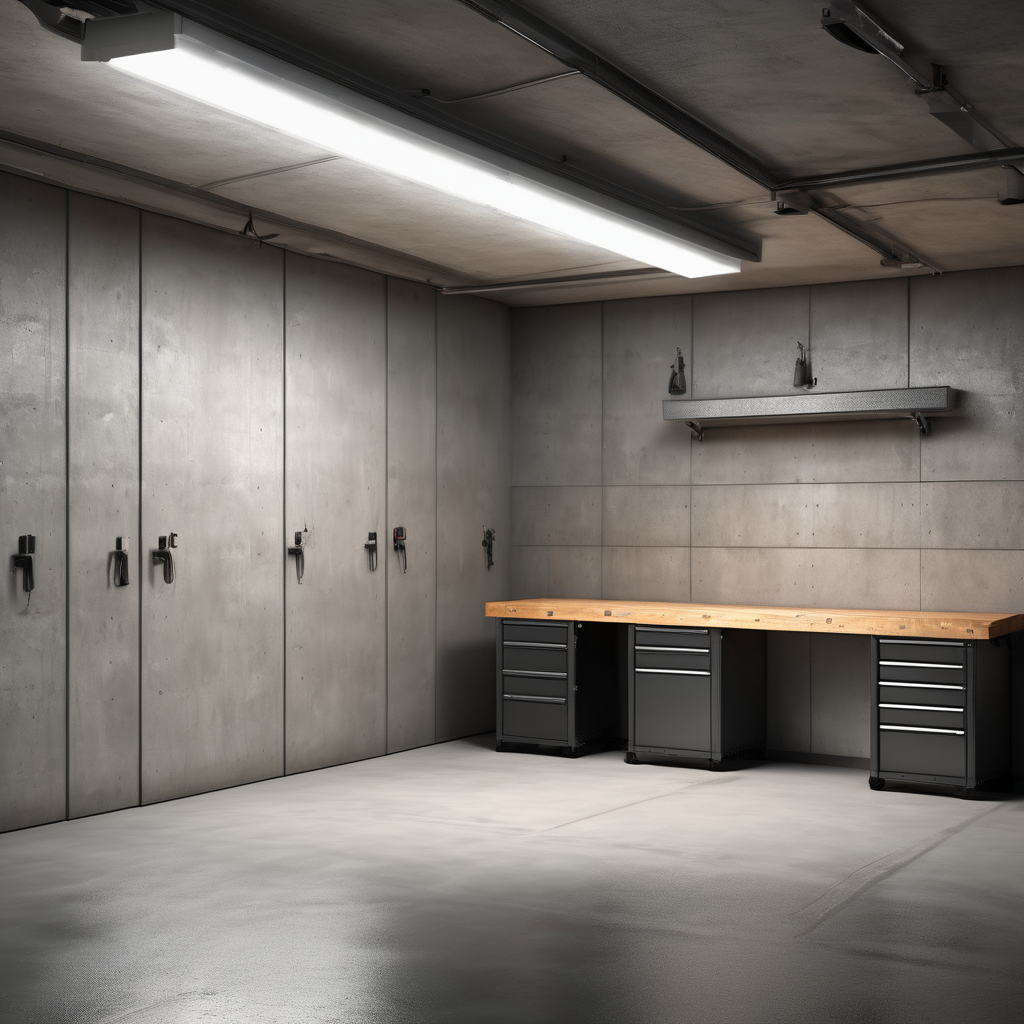} &
        \bgteximg{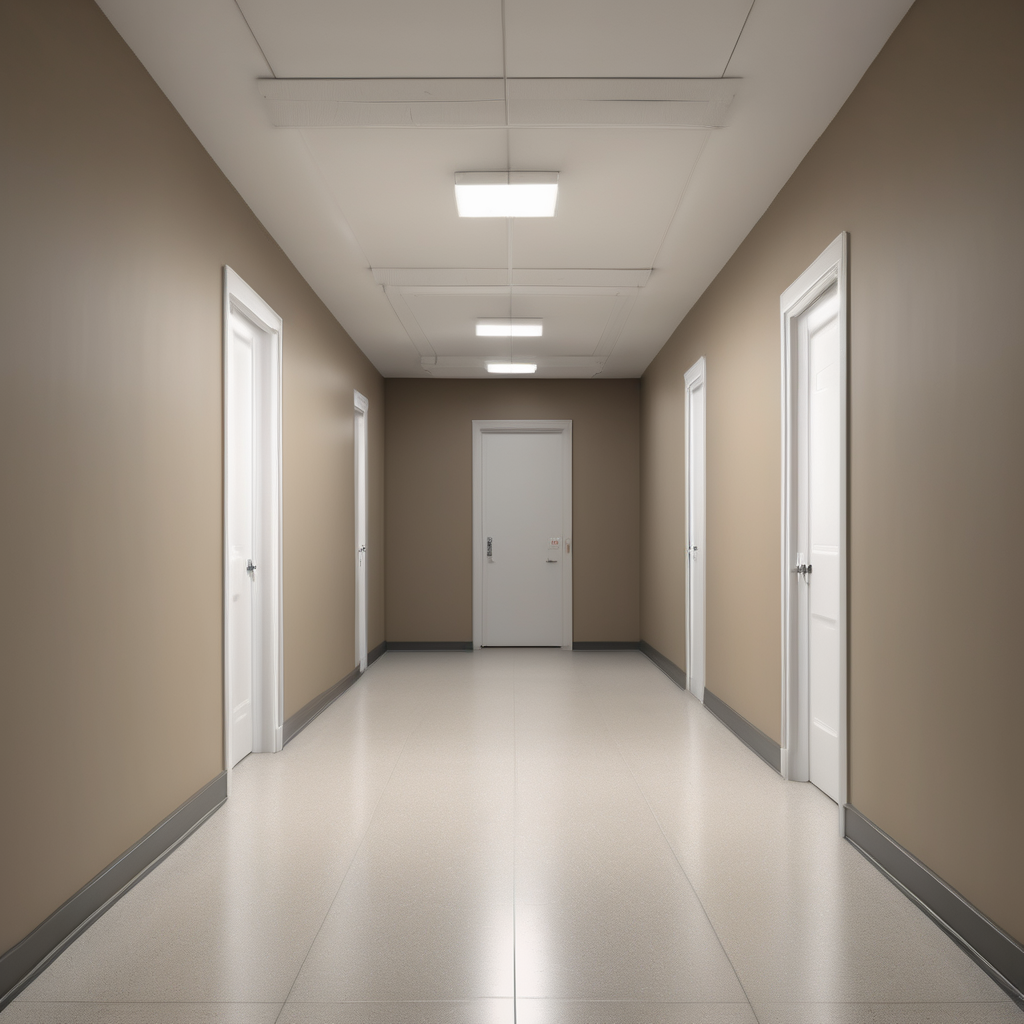} &
        \bgteximg{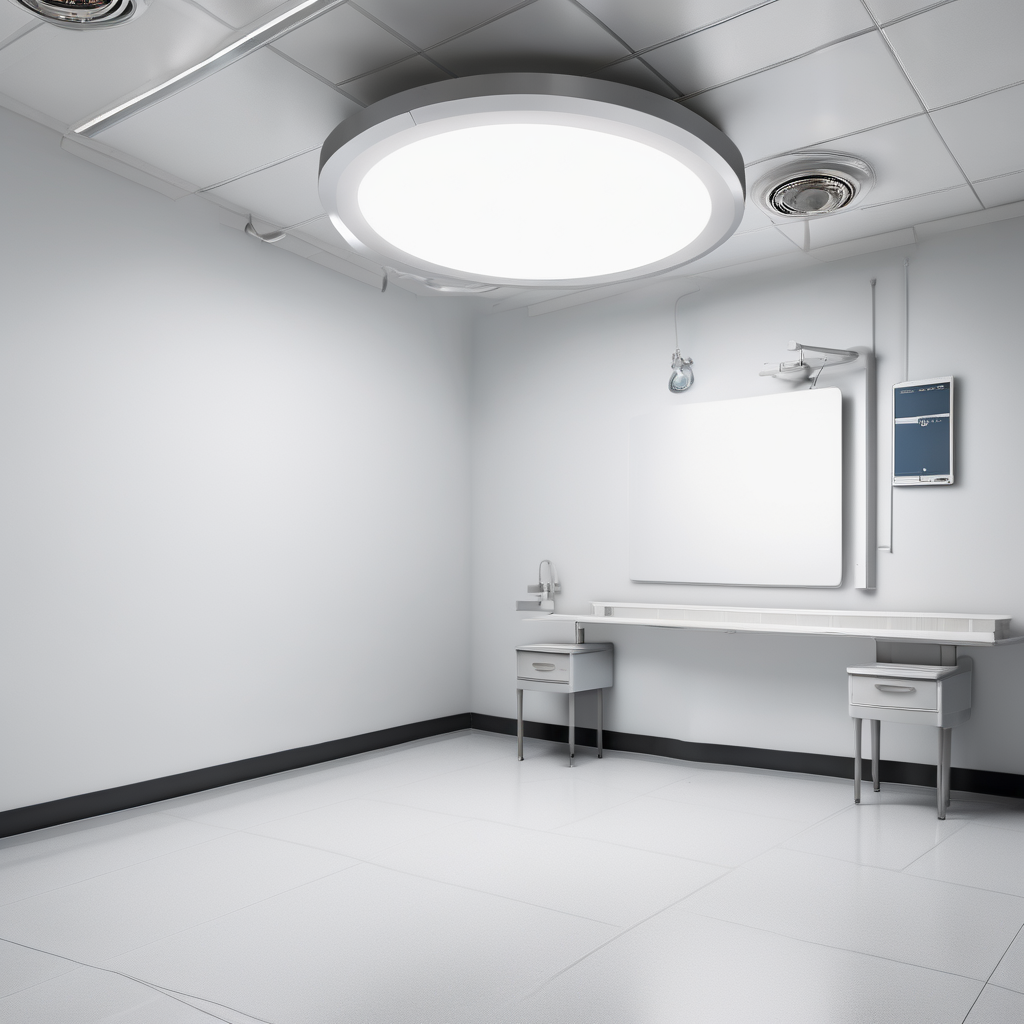} &
        \bgteximg{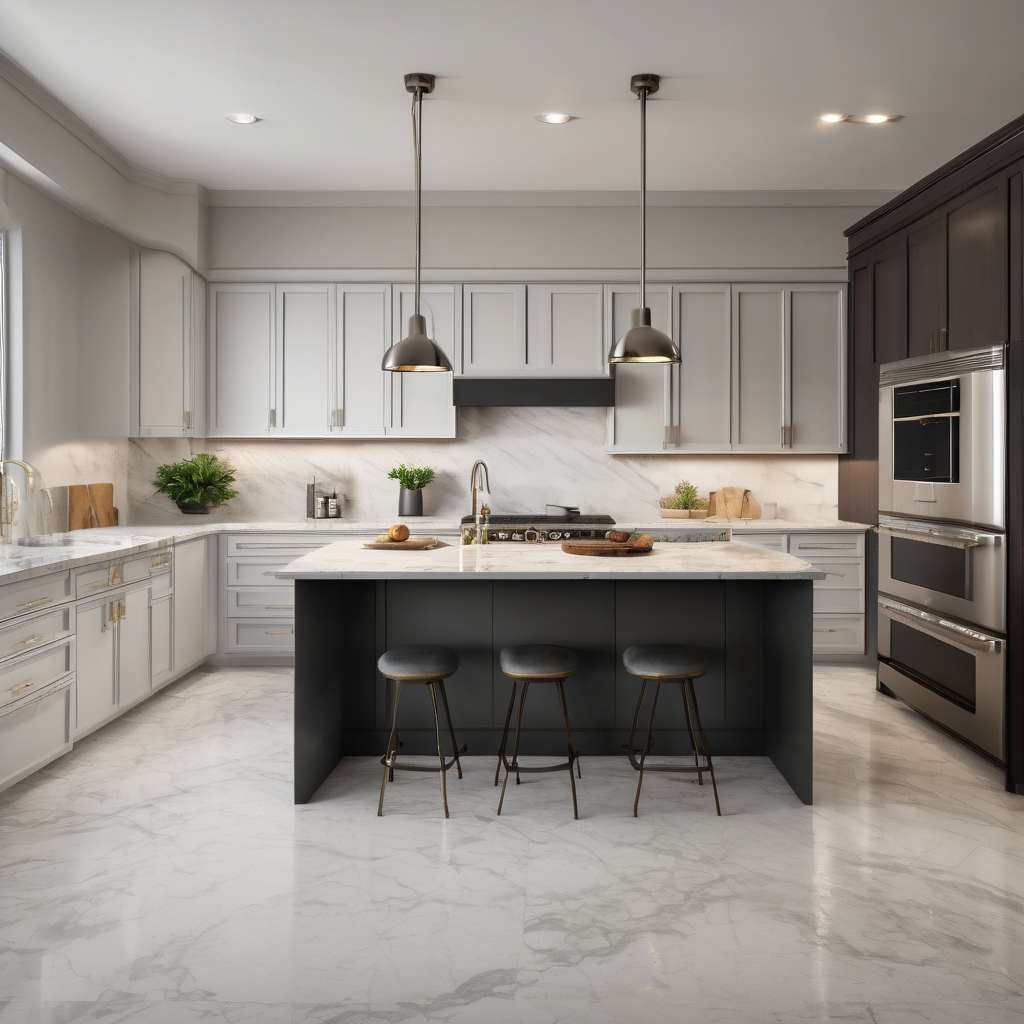} &
        \bgteximg{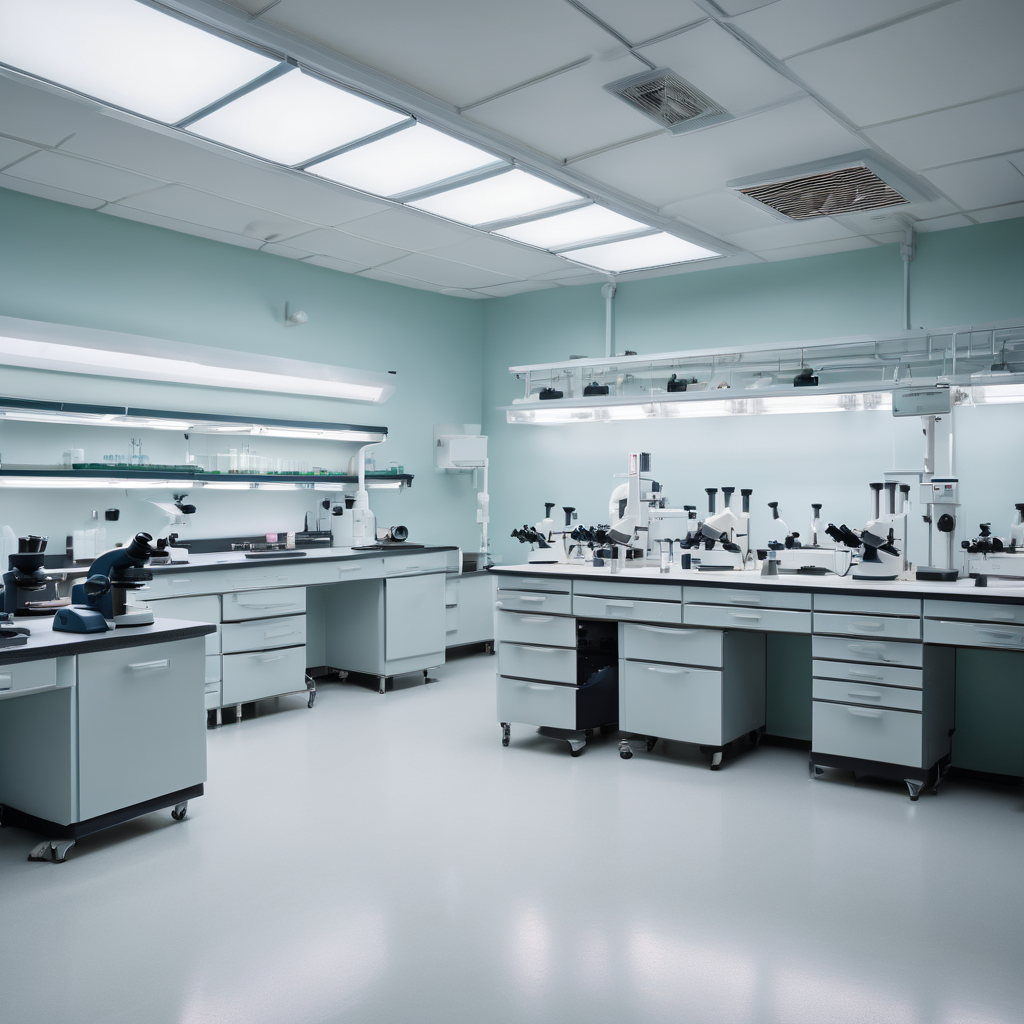} &
        \bgteximg{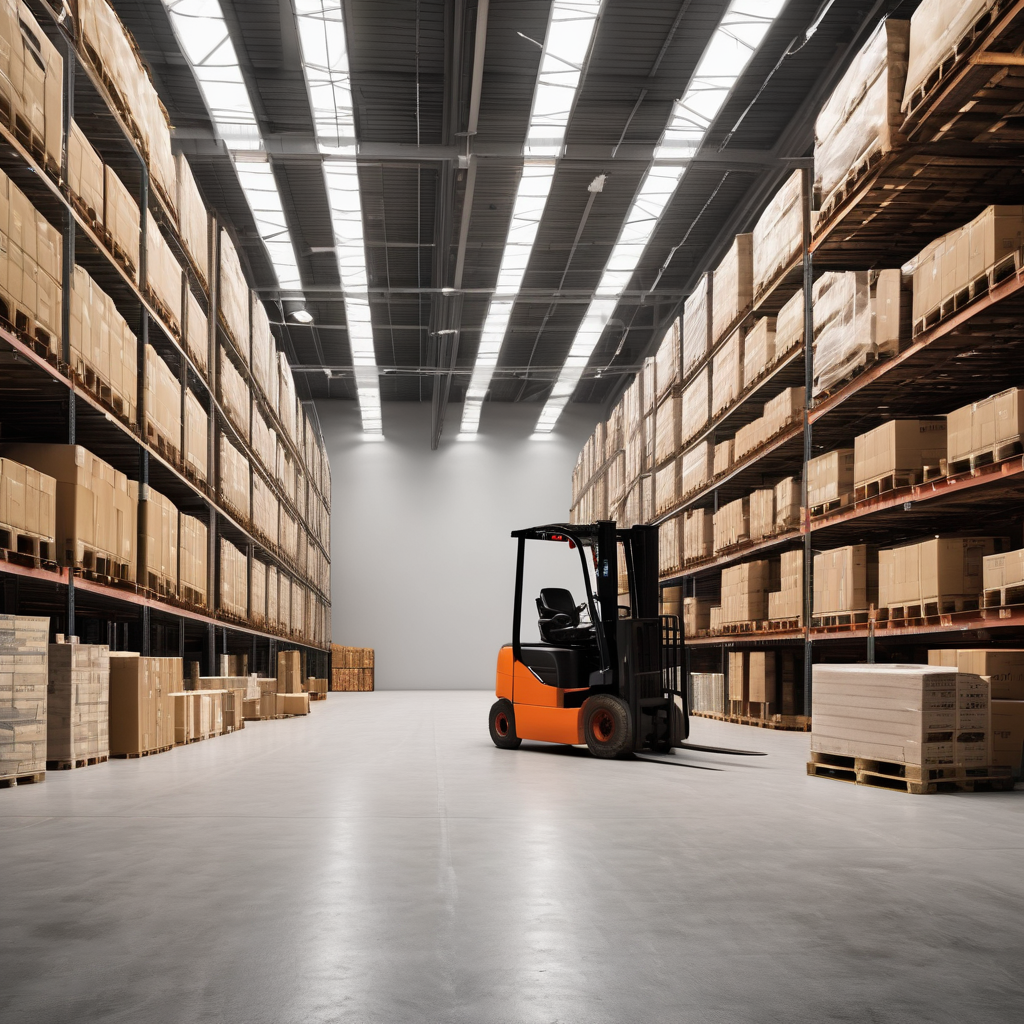} &
        \bgteximg{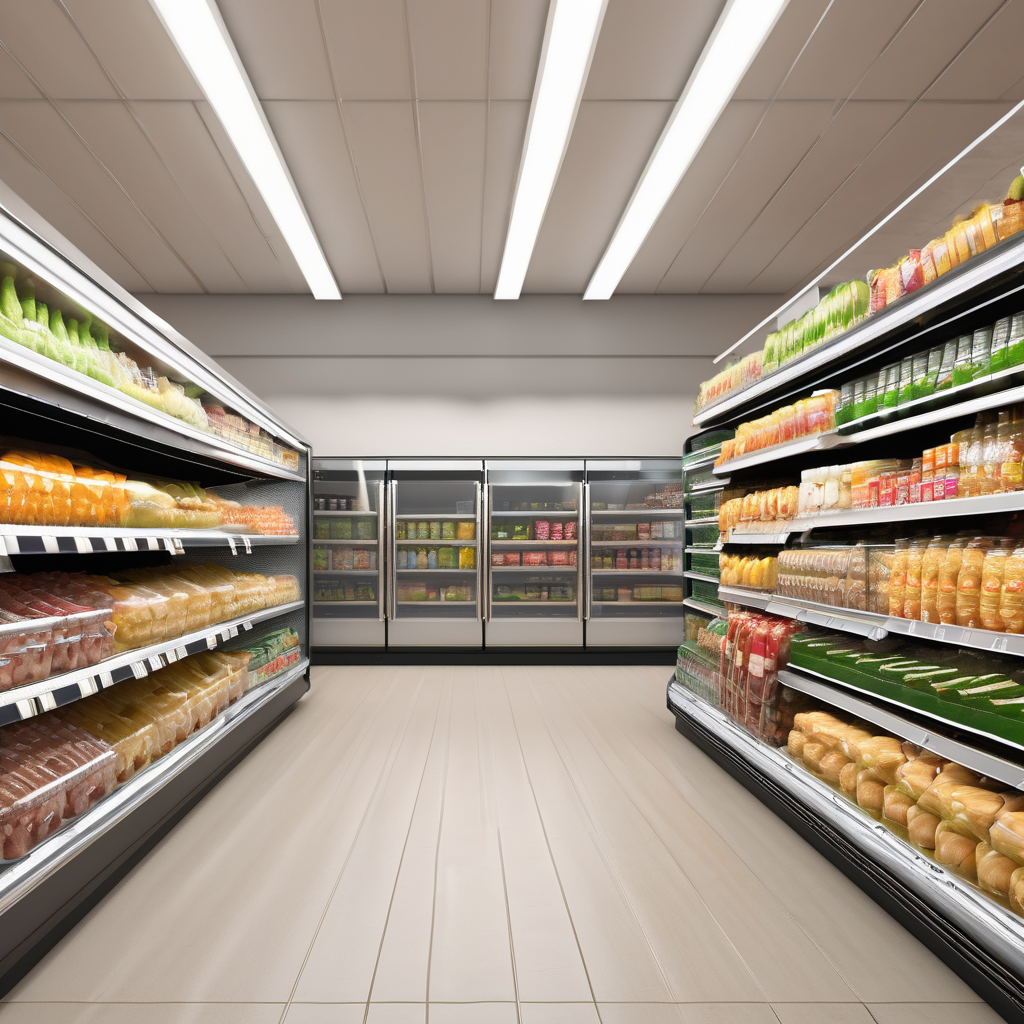} &
        \bgteximg{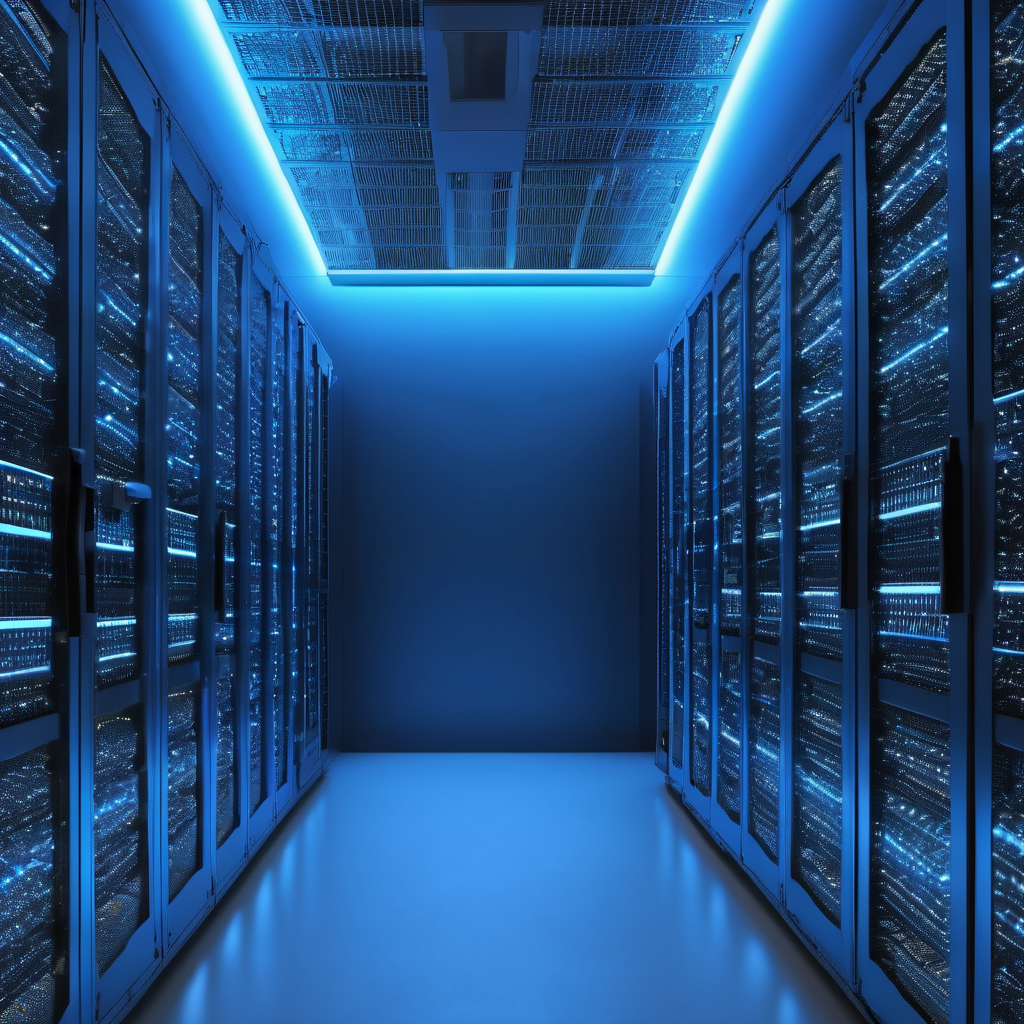}
    \end{tabularx}

    \caption{
    Representative assets for mask-based background replacement.
    We sample table and floor textures from Poly Haven and generate wall and background assets with Stable Diffusion XL.
    }
    \label{fig:background_examples}
\end{figure}

\paragraph{Background replacement.}
For ManiSkill trajectories~\citep{tao2024maniskill3} with available segmentation masks, we perform mask-based background replacement for table, floor, and wall regions.
Table and floor textures are sampled from Poly Haven~\citep{polyhaven}, while wall and background assets are generated with Stable Diffusion XL using text-to-image prompts for robot-relevant environments such as factories, kitchens, laboratories, workspaces, loading docks, and server rooms~\citep{podell2023sdxl,stabilityai2023sdxlbase}.
The resulting texture pool contains 92 table textures, 163 floor textures, and 604 wall/background textures, for a total of 859 region-specific assets.
At training time, textures are sampled by region and composited into the corresponding segmentation masks.
For domain-positive pairs, the same texture IDs, mask regions, and replacement mode define the shared domain condition.
Representative samples from the texture pool are shown in Fig.~\ref{fig:background_examples}.

\paragraph{Pair mining details.}
For the task encoder, each task-positive pair is constructed from the same episode and the same temporal window.
Each training item samples two non-overlapping windows from the same episode, denoted by
\(\mathbf{o}_{t_a}^{i}\) and \(\mathbf{o}_{t_b}^{i}\), and samples two domain conditions \(\eta_1\) and \(\eta_2\).
We then form two task-positive pairs:
\begin{equation}
    \big(
    \mathcal T_{\eta_1}(\mathbf{o}_{t_a}^{i}),
    \mathcal T_{\eta_2}(\mathbf{o}_{t_a}^{i})
    \big),
    \qquad
    \big(
    \mathcal T_{\eta_1}(\mathbf{o}_{t_b}^{i}),
    \mathcal T_{\eta_2}(\mathbf{o}_{t_b}^{i})
    \big).
\end{equation}
Within each pair, the two clips share the same underlying temporal window but differ in domain condition.
Across the two pairs, the temporal windows are distinct while the task identity and domain conditions are shared.
After collation, the two task-positive pairs per item are flattened from \([B,2,\cdots]\) to \([2B,\cdots]\) before applying InfoNCE.
Thus, examples from the same episode but different temporal windows are included as ordinary in-batch negatives.
This discourages the task encoder from solving the contrastive objective using task identity alone, while avoiding false negatives from the same underlying temporal segment.

For the domain encoder, a domain-positive pair is constructed by applying the same domain condition to two different clips.
Each training item first samples a source pair \((\mathbf{o}_{a},\mathbf{o}_{b})\), either from two non-overlapping windows of the same episode or from cross-task clips within the same camera bucket.
It then samples four distinct domain-transformation types without replacement, with corresponding domain conditions \(\eta_1,\ldots,\eta_4\), and forms four domain-positive pairs:
\begin{equation}
    \big(
    \mathcal T_{\eta_m}(\mathbf{o}_{a}),
    \mathcal T_{\eta_m}(\mathbf{o}_{b})
    \big),
    \qquad
    m=1,\ldots,4.
\end{equation}
Within each pair, the clips differ in trajectory content but share the same domain condition.
Across the four pairs in the same item, the source clips are fixed while the domain-transformation type changes.
After collation, the four domain-positive pairs per item are flattened from \([B,4,\cdots]\) to \([4B,\cdots]\) before applying InfoNCE.
Therefore, the domain encoder receives controlled in-batch comparisons: for an anchor under domain condition \(\eta_m\), other clips from the same item but with \(\eta_{m'}\neq\eta_m\) appear among the standard in-batch negatives.
No separate hard-negative loss is used; the dataloader simply ensures that informative non-matching domain conditions are present in the minibatch.

All transformations are applied consistently across the frames of a chunk so that each video chunk corresponds to a coherent visual domain.

\subsection{Task-Domain Latent Disentanglement Objectives}
\label{app:disentanglement_objectives}

\paragraph{InfoNCE objective.}
For a minibatch of positive pairs \(\{(x_i,x_i^+)\}_{i=1}^{B}\), let \(z_i\) and \(z_i^+\) denote the pooled latent vectors from the corresponding encoder.
We use the InfoNCE loss
\begin{equation}
    \mathcal L_{\mathrm{NCE}}
    =
    -
    \frac{1}{B}
    \sum_{i=1}^{B}
    \log
    \frac{
        \exp(\mathrm{sim}(z_i,z_i^+)/\tau)
    }{
        \sum_{k=1}^{B}
        \exp(\mathrm{sim}(z_i,z_k^+)/\tau)
    },
\end{equation}
where \(\mathrm{sim}(\cdot,\cdot)\) denotes cosine similarity and \(\tau\) is a temperature.
For the task encoder, the positive pairs are task-positive pairs; for the domain encoder, the positive pairs are domain-positive pairs.
The resulting losses are \(\mathcal L_{\mathrm{task}}\) and \(\mathcal L_{\mathrm{dom}}\).

\paragraph{Gaussian disentanglement loss.}
The task and domain encoders are trained with different positive pairs, but without an additional constraint, the resulting task and domain latents can still encode overlapping information.
We therefore define a Gaussian disentanglement loss that regularizes concatenated latents toward an isotropic Gaussian distribution.

For a minibatch, we construct
\begin{equation}
    u_i^{\mathrm{td}}
    =
    [z_i^{\mathrm{task}};z_i^{\mathrm{dom}}],
    \qquad
    i=1,\ldots,B,
\end{equation}
where \([\cdot;\cdot]\) denotes concatenation.
Here \(z_i^{\mathrm{task}},z_i^{\mathrm{dom}}\in\mathbb R^{D_z}\), and therefore
\(u_i^{\mathrm{td}}\in\mathbb R^{D}\) with \(D=2D_z\).
We implement the disentanglement regularizer \(\mathcal R_{\mathrm{dis}}\) using SIGReg~\citep{balestriero2025lejepaprovablescalableselfsupervised}.
Given a batch of vectors \(U=\{u_i\}_{i=1}^{B}\subset\mathbb R^{D}\), we sample \(M\) random unit directions \(r_m\sim\mathrm{Unif}(\mathbb S^{D-1})\), project \(s_{i,m}=r_m^\top u_i\), and match each projected distribution to a standard Gaussian.
Using a Gaussian kernel with bandwidth \(\sigma\), the Epps--Pulley statistic for the projected samples \(\{s_{i,m}\}_{i=1}^{B}\) along direction \(r_m\) is
\begin{align}
    \mathrm{EP}_{\sigma}(\{s_{i,m}\}_{i=1}^{B})
    &=
    \frac{1}{B^2}
    \sum_{i=1}^{B}
    \sum_{j=1}^{B}
    \exp
    \left(
        -\frac{(s_{i,m}-s_{j,m})^2}{2\sigma^2}
    \right) \nonumber \\
    &\quad
    -
    \frac{2\sigma}{B\sqrt{\sigma^2+1}}
    \sum_{i=1}^{B}
    \exp
    \left(
        -\frac{s_{i,m}^2}{2(\sigma^2+1)}
    \right)
    +
    \frac{\sigma}{\sqrt{\sigma^2+2}}.
\end{align}
The regularizer averages this statistic over random projection directions:
\begin{equation}
    \mathcal R_{\mathrm{dis}}(U)
    =
    \frac{1}{M}
    \sum_{m=1}^{M}
    \mathrm{EP}_{\sigma}
    \left(
    \{r_m^\top u_i\}_{i=1}^{B}
    \right).
\end{equation}
The task-domain disentanglement loss is
\begin{equation}
    \mathcal L_{\mathrm{dis}}^{\mathrm{td}}
    =
    \mathcal R_{\mathrm{dis}}
    \left(
    \{u_i^{\mathrm{td}}\}_{i=1}^{B}
    \right).
\end{equation}

\paragraph{Why Gaussian matching encourages disentanglement.}
\label{app:gaussian_disentanglement}
The Gaussian disentanglement loss does not guarantee exact independence for arbitrary distributions.
Its motivation is clearest under a joint Gaussian approximation.
Let
\[
    u
    =
    \begin{bmatrix}
    z^{a} \\
    z^{b}
    \end{bmatrix}
\]
be a jointly Gaussian random vector with covariance
\begin{equation}
    \Sigma
    =
    \begin{bmatrix}
    \Sigma_{a} & C \\
    C^\top & \Sigma_{b}
    \end{bmatrix}.
\end{equation}
If the concatenated latent is isotropic Gaussian, then \(\Sigma=I\), which implies
\begin{equation}
    \Sigma_{a}=I,
    \qquad
    \Sigma_{b}=I,
    \qquad
    C=0.
\end{equation}
For jointly Gaussian variables, zero cross-covariance is sufficient for independence, so
\begin{equation}
    p(z^{a},z^{b})
    =
    p(z^{a})p(z^{b}).
\end{equation}
Thus, driving the concatenated latent \([z^{a},z^{b}]\) toward an isotropic Gaussian encourages the two components to be disentangled.

We use this argument twice.
For task-domain latent disentanglement, \(z^{a}=z^{\mathrm{task}}\) and \(z^{b}=z^{\mathrm{dom}}\).
For policy learning, \(z^{a}=z^{\mathrm{curr}}\) and \(z^{b}=z^{\mathrm{dom},\star}\).
In practice, this should be interpreted as an approximate disentanglement regularizer rather than a guarantee of exact independence.

\paragraph{Total disentanglement objective.}
The full task-domain disentanglement objective is
\begin{equation}
    \mathcal L_{\mathrm{TDD}}
    =
    \lambda_{\mathrm{task}}\mathcal L_{\mathrm{task}}
    +
    \lambda_{\mathrm{dom}}\mathcal L_{\mathrm{dom}}
    +
    \lambda_{\mathrm{dis}}^{\mathrm{td}}
    \mathcal L_{\mathrm{dis}}^{\mathrm{td}}.
\end{equation}

\subsection{Policy Objectives}
\label{app:policy_objectives}

\paragraph{Lookahead contrastive alignment.}
For a minibatch, the Lookahead Predictor outputs \(\{\hat z_i^{\mathrm{look}}\}_{i=1}^{B}\), and the pretrained task encoder provides future task targets \(\{z_i^{\mathrm{task},\star}\}_{i=1}^{B}\).
We train the Lookahead Predictor with the in-batch InfoNCE loss
\begin{equation}
    \mathcal L_{\mathrm{look}}
    =
    -
    \frac{1}{B}
    \sum_{i=1}^{B}
    \log
    \frac{
        \exp
        \left(
        \mathrm{sim}
        \left(
        \hat z_i^{\mathrm{look}},
        \mathrm{sg}(z_i^{\mathrm{task},\star})
        \right)
        /\tau_{\mathrm{look}}
        \right)
    }{
        \sum_{k=1}^{B}
        \exp
        \left(
        \mathrm{sim}
        \left(
        \hat z_i^{\mathrm{look}},
        \mathrm{sg}(z_k^{\mathrm{task},\star})
        \right)
        /\tau_{\mathrm{look}}
        \right)
    }.
\end{equation}
The positive pair is the predicted lookahead and its corresponding future task target; for each anchor, all other future task targets in the batch serve as negatives. The stop gradient $\mathrm{sg}(\cdot)$ prevents this loss from updating the pretrained task encoder.

\paragraph{Current Representation Head domain disentanglement.}
For a minibatch, let \(\{z_i^{\mathrm{curr}}\}_{i=1}^{B}\) denote current representations produced by the Current Representation Head, and let \(\{z_i^{\mathrm{dom},\star}\}_{i=1}^{B}\) denote the corresponding domain latents extracted by the pretrained domain encoder.
We construct
\begin{equation}
    u_i^{\mathrm{curr}}
    =
    [z_i^{\mathrm{curr}};\mathrm{sg}(z_i^{\mathrm{dom},\star})],
    \qquad
    i=1,\ldots,B.
\end{equation}
The current-domain disentanglement loss is
\begin{equation}
    \mathcal L_{\mathrm{dis}}^{\mathrm{curr}}
    =
    \mathcal R_{\mathrm{dis}}
    \left(
    \{u_i^{\mathrm{curr}}\}_{i=1}^{B}
    \right).
\end{equation}
Under the Gaussian approximation described above, this encourages the current representation to be disentangled from the domain component while retaining information needed for action prediction.

\paragraph{Action loss and policy objective.}
The action head receives both policy latents:
\begin{equation}
    \hat a_{t:t+K_a-1}
    =
    A_\phi
    \left(
    [\hat z_t^{\mathrm{look}};z_t^{\mathrm{curr}}]
    \right).
\end{equation}
The behavior cloning loss is
\begin{equation}
    \mathcal L_{\mathrm{act}}
    =
    \frac{1}{K_a}
    \sum_{k=0}^{K_a-1}
    \left\|
    \hat a_{t+k}
    -
    a_{t+k}
    \right\|_1.
\end{equation}
The full policy objective is
\begin{equation}
    \mathcal L_{\mathrm{policy}}
    =
    \lambda_{\mathrm{act}}\mathcal L_{\mathrm{act}}
    +
    \lambda_{\mathrm{look}}\mathcal L_{\mathrm{look}}
    +
    \lambda_{\mathrm{dis}}^{\mathrm{curr}}
    \mathcal L_{\mathrm{dis}}^{\mathrm{curr}}.
\end{equation}

\subsection{Architecture and Implementation Details}
\label{app:impl_details}

\paragraph{Task and domain encoders.}
In the main text, \(E_{\psi}^{\mathrm{task}}\) and \(E_{\xi}^{\mathrm{dom}}\) denote the full task and domain encoder branches that output pooled latent vectors.
Each branch consists of a VJEPA2 video backbone~\citep{assran2025vjepa2}, branch-specific LoRA adapters~\citep{hu2022lora}, and an attention-pooling projection head.

\paragraph{Multi-query attention pooling.}
Let \(H\in\mathbb R^{N\times D_v}\) be a sequence of input tokens and let \(Q\in\mathbb R^{R\times D_q}\) be \(R\) learned query vectors. We project queries, keys, and values using
\(W_q\in\mathbb R^{D_q\times d}\),
\(W_k\in\mathbb R^{D_v\times d}\), and
\(W_v\in\mathbb R^{D_v\times D_p}\).
A multi-query attention pooler computes
\begin{align}
    A
    &=
    \mathrm{softmax}
    \left(
    \frac{(QW_q)(HW_k)^\top}{\sqrt{d}}
    \right), \\
    U
    &=
    AHW_v
    \in
    \mathbb R^{R\times D_p}.
\end{align}
The pooled tokens \(U\) are flattened and further projected to the output latent dimension \(D_z\):
\begin{equation}
    \mathrm{Head}(H)
    =
    W_o\,\mathrm{vec}(U)+b_o,
    \qquad
    W_o\in\mathbb R^{D_z\times (R D_p)},\quad
    b_o\in\mathbb R^{D_z}.
\end{equation}
The projection heads for the task encoder, domain encoder, Lookahead Predictor, and Current Representation Head all use this attention-pooling-plus-linear architecture, with separate parameters.

\paragraph{Policy-side heads.}
The policy uses a Qwen2.5-VL~\cite{bai2025qwen25vltechnicalreport} backbone with LoRA adaptation.
On top of the VLM tokens, we train two AttentiveLatentHead modules and one ResNetActionHead.
The Lookahead Predictor maps VLM tokens to \(\hat z_t^{\mathrm{look}}\), and the Current Representation Head maps the same tokens to \(z_t^{\mathrm{curr}}\).
Each latent head uses 8 learned query tokens, a two-layer attentive pooler with 16 attention heads, and a linear projection to a 4096-dimensional latent.

\paragraph{Action head.}
The action head is an MLP-ResNet that maps the concatenated policy latents to an action chunk.
In the dual-head setting, the action head input dimension is \(4096+4096=8192\).
We use hidden dimension 2048, two residual MLP blocks, and output a chunk size of \(K_a=50\).

\begin{table}[h]
    \centering
    \footnotesize
    \caption{
    Policy-side head budget for the VLA implementation.
    }
    \label{tab:policy_head_budget}
    \begin{tabularx}{\linewidth}{@{}>{\raggedright\arraybackslash}X
                                  >{\raggedleft\arraybackslash}p{0.22\linewidth}@{}}
        \toprule
        Module & Parameters \\
        \midrule
        Lookahead Predictor, AttentiveLatentHead
        & 167.85M \\
        Current Representation Head, AttentiveLatentHead
        & 167.85M \\
        ResNetActionHead, dual-head input
        & 28.48M \\
        \midrule
        Total policy-side heads
        & 364.18M \\
        \bottomrule
    \end{tabularx}
\end{table}

\subsection{Training Stage Summary}
\label{app:training_schedule}

Table~\ref{tab:training_schedule} summarizes which modules are updated in each training stage.
In the main experiments, the Task-Domain Encoder is pretrained once on the source trajectory collection and then kept fixed.
This decouples reusable Task-Domain Encoder training from downstream VLA policy training.

\begin{table}[h]
    \centering
    \footnotesize
    \caption{
    Training Stage Summary for Domain-Invariant Latent Lookahead.
    }
    \label{tab:training_schedule}
    \begin{tabularx}{\linewidth}{@{}>{\raggedright\arraybackslash}p{0.20\linewidth}
                                      >{\raggedright\arraybackslash}p{0.27\linewidth}
                                      >{\raggedright\arraybackslash}p{0.31\linewidth}
                                      >{\raggedright\arraybackslash}X@{}}
        \toprule
        Stage & Data & Updated modules & Objective \\
        \midrule
        Task-domain encoder pretraining
        & Source trajectory collection
        & Task/domain encoder LoRA adapters and projection heads
        & \(\mathcal L_{\mathrm{TDD}}\) \\
        \midrule
        VLA policy pretraining
        & Source trajectory collection
        & VLM LoRA, Lookahead Predictor, Current Representation Head, action head
        & \(\mathcal L_{\mathrm{policy}}\)\\
        \midrule
        Downstream policy tuning
        & Downstream action-labeled demonstrations (e.g., LIBERO)
        & VLM LoRA, Lookahead Predictor, Current Representation Head, action head
        & \(\mathcal L_{\mathrm{policy}}\)\\
        \bottomrule
    \end{tabularx}
\end{table}

\paragraph{Optional Task-Domain Encoder adaptation.}
Although the main experiments keep the pretrained Task-Domain Encoder fixed for downstream policy tuning, the same task-domain disentanglement objective could be used to adapt the task and domain encoders on additional downstream demonstration datasets. When controllable rendering or segmentation masks are unavailable, such optional adaptation would rely on image-level transformations such as cropping, warping, lighting and color changes, camera-pipeline perturbations, and visual corruptions.

\begingroup

\subsection{Computational cost}
\label{app:computational_cost}

Task-Domain Encoder pretraining requires 273 A100 GPU-hours once; the resulting encoder is reused and remains frozen during downstream policy training. It is not used at deployment. Table~\ref{tab:app_computational_cost} separates downstream tuning cost, peak training memory, and inference latency. DILL increases downstream tuning cost by $42\%$ and peak memory by $1.5$ GB relative to Base VLA, while the measured latency increases by $0.3$ ms per 30-action chunk. Thus, the additional cost is concentrated in training rather than deployment.

\begin{table}[ht]
\centering

\small
\setlength{\tabcolsep}{6pt}
\renewcommand{\arraystretch}{1.08}
\begin{tabularx}{\linewidth}{@{}>{\raggedright\arraybackslash}Xcc@{}}
\toprule
Metric & Base VLA & DILL \\
\midrule
Downstream tuning cost ($\times$ Base VLA) & 1.00 & 1.42 \\
Peak training memory (GB) & 20.9 & 22.4 \\
Inference latency (ms per 30-action chunk) & 69.9 & 70.2 \\
\bottomrule
\end{tabularx}
\caption{\textbf{Training and inference costs.} The one-time Task-Domain Encoder pretraining cost is reported separately in the text. Latency is measured in a separate benchmark with 30-action chunks.}
\label{tab:app_computational_cost}
\end{table}
\par
\endgroup

\section{Simulation Benchmark Experiments}
\label{app:simulation_benchmark_experiments}

\subsection{LIBERO Shortcut Diagnostic}
\label{app:libero_shortcut}

We use the LIBERO shortcut diagnostic of~\cite{shortcut_learning_in_GRPs} to test whether a policy follows the commanded task or instead executes the task spuriously associated with the observed view. Unlike standard robustness evaluation, this diagnostic explicitly separates shortcut-driven task substitution from general execution failure.

\paragraph{Task--view confounding.}
The benchmark constructs two confounded task--view islands. We refer to the left-view task group as Task-L and the right-view task group as Task-R. Task-L contains LIBERO task IDs $\{0,1,3,5,8\}$ and is observed only in the left-view range ($10^\circ$--$25^\circ$) during training. Task-R contains task IDs $\{2,4,6,7,9\}$ and is observed only in the right-view range ($55^\circ$--$70^\circ$). At test time, we evaluate counterfactual task--view compositions by swapping these associations: Task-R is evaluated at the left viewpoint ($10^\circ$), and Task-L is evaluated at the right viewpoint ($70^\circ$). A task-faithful policy should follow the language instruction under these swapped views, whereas a shortcut-prone policy will execute the task spuriously associated with the observed view.

\paragraph{Metrics.}
We report two complementary metrics. \emph{OOD success rate} measures whether the commanded task is completed under the counterfactual view. \emph{Shortcut degree} measures whether the policy instead executes the task spuriously associated with the observed view during training, thereby isolating view-induced task substitution.

\subsection{Ablation Study}
\label{app:libero_shortcut_ablations}

We ablate DILL to identify which design choices are responsible for shortcut mitigation. The central question is not whether a model can fit the confounded training distribution, but whether it can avoid using viewpoint as a proxy for task identity when the task--view association is broken. DILL is designed to address this by reshaping the information routed to the action head: a predicted lookahead latent aligned with the future task latent, and a current representation disentangled from the corresponding domain latent. The ablation study therefore asks whether each of these ingredients is necessary, and whether simpler alternatives---more augmented data or generic future prediction---are sufficient.

All variants are evaluated on the LIBERO shortcut diagnostic in Appendix~\ref{app:libero_shortcut}. This diagnostic separates three levels of generalization. \emph{In-dist. SR} measures success on the original task--view training compositions. \emph{Center OOD SR} evaluates each task group at an unseen midpoint viewpoint without swapping task--view association, measuring interpolation to an unseen visual domain. \emph{Counter OOD SR} evaluates the counterfactual task--view swaps, where the model must follow the commanded task rather than the task associated with the observed view. Finally, \emph{shortcut degree} measures how often the model follows the task spuriously associated with the observed view; lower is better.

\textbf{Compared methods.} We compare DILL with several ablative models: 

\noindent\textbf{(1) Base VLA.}
Base VLA is the plain behavior-cloning baseline. It uses the same VLA architecture as DILL, but directly maps the current observation-conditioned representation to actions. It does not use source augmentation, task--domain latent targets, the disentangled current head, or latent lookahead alignment. This is the minimal baseline for testing whether standard VLA behavior cloning is sufficient under task--view confounding.

\smallskip
\noindent\textbf{(2) Base VLA + SA.}
This variant keeps the same behavior-cloning objective and VLA architecture as Base VLA, but trains with the same augmented source data used by DILL for task--domain encoder pretraining. This controls for the data condition: if this variant were sufficient, the gain of DILL could be attributed mainly to additional visual diversity. If not, then the improvement must come from how DILL uses the augmented data to shape the latent space, rather than from the data alone.

\smallskip
\noindent\textbf{(3) Entangled LA.}
This variant adds latent lookahead prediction, but uses the representation from the original video encoder~\cite{assran2025vjepa2} as the future target instead of the task--domain latent targets. Thus, the predicted future latent is not explicitly encouraged to discard domain-specific information, and may still contain viewpoint, lighting, or background cues. This variant tests whether generic future prediction is sufficient, or whether the lookahead target must be domain-invariant.

\smallskip
\noindent\textbf{(4) DILL w/o CH.}
This variant removes the disentangled current head. 
The model still uses task-domain latent targets and latent lookahead alignment, but the current action-conditioning representation is no longer explicitly regularized to be disentangled from the corresponding domain latent.
This tests whether domain-invariant lookahead alone is sufficient, or whether the current representation used by the policy also needs to be explicitly shaped to suppress domain-specific visual cues.

\smallskip
\noindent\textbf{(5) DILL w/o LA.}
This variant keeps the task--domain latent targets and the disentangled current head, but removes latent lookahead alignment. In other words, the current representation is still regularized through the task--domain latent structure, but the future lookahead latent is not trained to align with the domain-invariant target. This tests whether a disentangled current representation alone can mitigate shortcut learning, or whether robust control requires explicitly aligning the predicted future latent as well.

\smallskip
\noindent\textbf{(6) DILL.}
DILL is the full model, combining source augmentation, task--domain latent targets, the disentangled current head, and domain-invariant latent lookahead alignment.

\providecommand{\cmark}{\ensuremath{\surd}}

\begin{table*}[t]
\centering
\small
\setlength{\tabcolsep}{3.0pt}
\renewcommand{\arraystretch}{1.16}
\begin{tabularx}{\linewidth}{
@{}
>{\raggedright\arraybackslash}p{0.190\linewidth}
>{\centering\arraybackslash}p{0.045\linewidth}
>{\centering\arraybackslash}p{0.045\linewidth}
>{\centering\arraybackslash}p{0.045\linewidth}
>{\centering\arraybackslash}p{0.045\linewidth}
@{\hskip 0.50em}
>{\centering\arraybackslash}X
>{\centering\arraybackslash}X
>{\centering\arraybackslash}X
>{\centering\arraybackslash}X
>{\centering\arraybackslash}X
@{}}
\toprule
\multirow{2}{*}{Variant}
& \multicolumn{4}{c}{Components}
& \multicolumn{5}{c}{Metrics} \\
\cmidrule(lr){2-5} \cmidrule(lr){6-10}
& SA
& TD
& CH
& LA
& \shortstack{In-dist.\\SR $\uparrow$}
& \shortstack{Center\\OOD SR $\uparrow$}
& \shortstack{Counter\\OOD SR $\uparrow$}
& \shortstack{Avg.\\SR $\uparrow$}
& \shortstack{Shortcut\\degree $\downarrow$} \\
\midrule

Base VLA
&  &  &  &
& 68\% & 22\% & 0\% & 30\% & 73\% \\
\addlinespace[0.45em]

Base VLA + SA
& \cmark &  &  &
& \textbf{82\%} & 10\% & 2\% & 31\% & 69\% \\
\addlinespace[0.45em]

Entangled LA
& \cmark &  &  & \cmark
& \underline{76\%} & 24\% & 0\% & 33\% & 65\% \\
\addlinespace[0.45em]

DILL w/o CH
& \cmark & \cmark &  & \cmark
& 60\% & \underline{44\%} & 44\% & 49\% & 6\% \\
\addlinespace[0.45em]

DILL w/o LA
& \cmark & \cmark & \cmark &
& 50\% & \textbf{62\%} & \textbf{62\%} & \underline{58\%} & \textbf{3\%} \\

\midrule

\textbf{DILL}
& \cmark & \cmark & \cmark & \cmark
& 66\% & \textbf{62\%} & \underline{58\%} & \textbf{62\%} & \underline{5\%} \\

\bottomrule
\end{tabularx}
\caption{\textbf{Ablation study.}
All variants are evaluated on the LIBERO shortcut diagnostic under the same task--view island protocol. Component columns indicate whether each variant uses source augmentation (SA), task--domain latent targets from the task/domain encoders (TD), the disentangled current head (CH), and latent lookahead alignment (LA). In-dist. SR averages the original training compositions, Center OOD SR evaluates unseen midpoint viewpoints without swapping task identity, and Counter OOD SR evaluates counterfactual task--view swaps. Shortcut degree measures how often the policy follows the task associated with the observed view rather than the commanded task. Blank component entries indicate that the component is absent.}
\label{tab:app_libero_shortcut_ablations}
\end{table*}

\textbf{Results.}
Table~\ref{tab:app_libero_shortcut_ablations} shows that fitting the confounded training distribution is not sufficient for shortcut-free generalization. Base VLA reaches $68\%$ in-distribution success, but obtains $0\%$ Counter OOD success and a high shortcut degree of $73\%$. Adding source augmentation improves in-distribution success to $82\%$ and slightly reduces shortcut degree to $69\%$, indicating that additional visual diversity and action pretraining are helpful for fitting the nominal tasks and can mildly reduce shortcut reliance. However, Counter OOD success remains only $2\%$, showing that data augmentation alone does not teach the policy which visual factors should be ignored when task--view correlations are counterfactually broken.

Entangled LA provides a second partial improvement. Compared to Base VLA + SA, it improves Center OOD success from $10\%$ to $24\%$ and further reduces shortcut degree from $69\%$ to $65\%$, suggesting that future prediction can encourage representations that are somewhat more robust to unseen viewpoints. However, it still obtains $0\%$ Counter OOD success. This failure is informative: predicting a future latent is not sufficient if the target representation remains entangled with viewpoint, background, or other domain-specific cues. The lookahead target must be tied to task-relevant structure rather than to the same visual correlations present in the training distribution.

The lower half of the table isolates the DILL-specific components. DILL w/o CH, which keeps task--domain latent targets and latent lookahead alignment but removes the disentangled current head, reaches $44\%$ Center OOD and $44\%$ Counter OOD success while reducing shortcut degree to $6\%$. This is a large jump over Entangled LA, indicating that aligning lookahead prediction with task--domain latent targets removes much of the view-induced task substitution. DILL w/o LA, which keeps the disentangled current head but removes latent lookahead alignment, achieves the strongest Counter OOD success ($62\%$) and the lowest shortcut degree ($3\%$), but its in-distribution success drops to $50\%$. This suggests that the disentangled current head is highly effective at suppressing shortcut behavior, while latent lookahead alignment helps recover a better balance between nominal task execution and counterfactual generalization.

The full DILL model provides the best overall tradeoff. It matches the best Center OOD success ($62\%$), remains close to the best Counter OOD success ($58\%$), keeps shortcut degree very low ($5\%$), and achieves the highest average success across the three success metrics ($62\%$). The key pattern is that source augmentation and generic lookahead improve some aspects of learning, but do not solve counterfactual task--view generalization. In contrast, the DILL-family variants that use task--domain latent targets sharply reduce shortcut degree, and the full model best preserves both task execution and shortcut resistance. These results support the design principle of DILL: shortcut mitigation requires not only additional visual diversity or future prediction, but a task-structured latent pathway that routes domain-invariant predictive information into action conditioning.

\subsection{LIBERO-Plus evaluation details}
\label{app:libero_plus_full}

We evaluate methods on LIBERO-Plus~\cite{libero_plus}, which tests robustness under controlled perturbations of the original LIBERO benchmark. In the main paper, we report the visual perturbation categories because they are most directly aligned with our claim about visual shortcut mitigation. Here, we provide the full seven-category breakdown. All models are evaluated zero-shot on LIBERO-Plus after training on the original LIBERO setting.

\paragraph{Full perturbation breakdown.}
Table~\ref{tab:app_liberoplus_full} reports success rates on all LIBERO-Plus perturbation categories. \emph{Total} denotes the average over Camera, Robot, Language, Light, BG, Noise, and Layout. The second row for each model reports the absolute drop relative to the original LIBERO score.

\providecommand{\snum}[1]{{\footnotesize #1}}

\begin{table*}[h]
\centering
\small
\setlength{\tabcolsep}{2.0pt}
\renewcommand{\arraystretch}{0.98}
\begin{threeparttable}
\begin{tabularx}{\linewidth}{
@{}
>{\raggedright\arraybackslash}p{0.18\linewidth}
*{9}{>{\centering\arraybackslash}X}
@{}}
\toprule
\multirow{2}{*}{Model}
& \multirow{2}{*}{Original}
& \multicolumn{7}{c}{LIBERO-Plus perturbations}
& \multirow{2}{*}{Total} \\
\cmidrule(lr){3-9}
& & Camera & Robot & Language & Light & BG & Noise & Layout & \\
\midrule

\multirow{2}{*}{OpenVLA~\cite{openvla}}
  & \multirow{2}{*}{\snum{76.5}}
  & \snum{0.8} & \snum{3.5} & \snum{23.0} & \snum{8.1} & \snum{34.8} & \snum{15.2} & \snum{28.5} & \snum{15.6} \\
  &
  & \snum{\dropcell{75.7}} & \snum{\dropcell{73.0}} & \snum{\dropcell{53.5}} & \snum{\dropcell{68.4}} & \snum{\dropcell{41.7}} & \snum{\dropcell{61.3}} & \snum{\dropcell{48.0}} & \snum{\dropcell{60.9}} \\

\multirow{2}{*}{WorldVLA~\cite{worldvla}}
  & \multirow{2}{*}{\snum{79.1}}
  & \snum{0.1} & \snum{27.9} & \snum{41.6} & \snum{43.7} & \snum{17.1} & \snum{10.9} & \snum{38.0} & \snum{25.0} \\
  &
  & \snum{\dropcell{79.0}} & \snum{\dropcell{51.2}} & \snum{\dropcell{37.5}} & \snum{\dropcell{35.4}} & \snum{\dropcell{62.0}} & \snum{\dropcell{68.2}} & \snum{\dropcell{41.1}} & \snum{\dropcell{54.1}} \\

\multirow{2}{*}{UniVLA~\cite{univla}}
  & \multirow{2}{*}{\snum{95.5}}
  & \snum{1.8} & \snum{46.2} & \snum{69.6} & \snum{69.0} & \snum{81.0} & \snum{21.2} & \snum{31.9} & \snum{43.9} \\
  &
  & \snum{\dropcell{93.7}} & \snum{\dropcell{49.3}} & \snum{\dropcell{25.9}} & \snum{\dropcell{26.5}} & \snum{\dropcell{14.5}} & \snum{\dropcell{74.3}} & \snum{\dropcell{63.6}} & \snum{\dropcell{51.6}} \\

\multirow{2}{*}{NORA~\cite{nora}}
  & \multirow{2}{*}{\snum{87.9}}
  & \snum{2.2} & \snum{37.0} & \snum{65.1} & \snum{45.7} & \snum{58.6} & \snum{12.8} & \snum{62.1} & \snum{39.0} \\
  &
  & \snum{\dropcell{85.7}} & \snum{\dropcell{50.9}} & \snum{\dropcell{22.8}} & \snum{\dropcell{42.2}} & \snum{\dropcell{29.3}} & \snum{\dropcell{75.1}} & \snum{\dropcell{25.8}} & \snum{\dropcell{48.9}} \\

\multirow{2}{*}{\mbox{OpenVLA-OFT~\cite{oft}}}
  & \multirow{2}{*}{\snum{95.3}}
  & \snum{10.4} & \snum{38.7} & \snum{70.5} & \snum{76.8} & \snum{93.6} & \snum{49.9} & \snum{69.9} & \snum{55.8} \\
  &
  & \snum{\dropcell{84.9}} & \snum{\dropcell{56.6}} & \snum{\dropcell{24.8}} & \snum{\dropcell{18.5}} & \snum{\dropcell{1.7}} & \snum{\dropcell{45.4}} & \snum{\dropcell{25.4}} & \snum{\dropcell{39.5}} \\

\midrule

\multirow{2}{*}{\textbf{DILL (Ours)}}
  & \multirow{2}{*}{\snum{81.6}}
  & \snum{68.4} & \snum{21.4} & \snum{2.9} & \snum{69.4} & \snum{70.0} & \snum{68.7} & \snum{55.1} & \snum{50.8} \\
  &
  & \snum{\dropcell{13.2}} & \snum{\dropcell{60.2}} & \snum{\dropcell{78.7}} & \snum{\dropcell{12.2}} & \snum{\dropcell{11.6}} & \snum{\dropcell{12.9}} & \snum{\dropcell{26.5}} & \snum{\dropcell{30.8}} \\

\bottomrule
\end{tabularx}
\end{threeparttable}

\vspace{-0.35em}
\caption{\textbf{Full LIBERO-Plus robustness breakdown.}
For each model, the first row shows success rate (\%) on the original LIBERO benchmark and all seven LIBERO-Plus perturbation categories. The second row shows the absolute drop relative to the original score. \emph{Total} denotes the official LIBERO-Plus leaderboard score.}
\label{tab:app_liberoplus_full}
\vspace{-0.5em}
\end{table*}

\paragraph{Results.}
The full breakdown clarifies the scope of DILL's robustness. DILL is strongest on the perturbations that directly change visual appearance or viewpoint, achieving the best performance on Camera ($68.4\%$) and Noise ($68.7\%$), and competitive performance on Light ($69.4\%$) and BG ($70.0\%$). This pattern matches the design of the method: domain-invariant latent lookahead is intended to suppress shortcuts tied to visual-domain cues, such as viewpoint, background texture, and image degradation. In contrast, DILL is not designed to directly solve non-visual shifts. The low Language score reflects a failure mode related to instruction generalization and language grounding, while the Robot score reflects sensitivity to robot initial-state variation. These require different invariances than the visual-domain invariance targeted by our objective.

This distinction is important for interpreting the Total score. DILL obtains a Total score of $50.8\%$, improving over OpenVLA, WorldVLA, UniVLA, and NORA, but remaining below OpenVLA-OFT due primarily to the Language and Robot categories. Rather than indicating a uniform robustness improvement across all axes, the result shows a more specific effect: DILL substantially reduces brittleness under visual distribution shifts, while leaving language and robot-state robustness as separate failure modes.

\section{Representation Diagnostics}
\label{app:representation_diagnostics}

\subsection{Pairwise latent similarity protocol}
\label{app:repr_pairwise_protocol}

We use pairwise latent similarity diagnostics to test whether the action-conditioning representation is organized around task content rather than domain-specific visual factors. For each representation, we compute cosine similarity between normalized latent vectors for two types of unseen task--domain pairs.

\textbf{Task pairs.}
A task pair consists of two observations that share the same underlying trajectory content but differ in visual domain condition. A task-centric latent should assign high similarity to these pairs because the task-relevant trajectory content is preserved across the domain change.

\textbf{Domain pairs.}
A domain pair consists of two observations that share the same visual domain condition but differ in trajectory content. A task-centric latent should assign lower similarity to these pairs because the underlying behavior is different. Conversely, a domain-centric latent would assign high similarity to domain pairs, even when the trajectory content changes.

\textbf{Metric.}
For a task-centric {action-conditioning representation}, we summarize the diagnostic using the task--domain separation gap,
\[
\Delta_{\text{task}} =
\mathbb{E}\left[\mathrm{sim}(z_i, z_j) \mid (i,j) \in \mathcal{P}_{\text{task}}\right]
-
\mathbb{E}\left[\mathrm{sim}(z_i, z_j) \mid (i,j) \in \mathcal{P}_{\text{domain}}\right],
\]
where $\mathcal{P}_{\text{task}}$ denotes task pairs and $\mathcal{P}_{\text{domain}}$ denotes domain pairs. A larger positive gap indicates that the representation is more aligned with task content and less dominated by shared visual domain.

\begin{figure*}[h]
\centering
\begin{minipage}[t]{0.45\textwidth}
    \centering
    \includegraphics[width=0.48\linewidth]{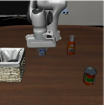}
    \hfill
    \includegraphics[width=0.48\linewidth]{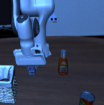}\\[-0.1em]
    \footnotesize \textbf{Task pair:} same trajectory, different domain
\end{minipage}
\hfill
\begin{minipage}[t]{0.45\textwidth}
    \centering
    \includegraphics[width=0.48\linewidth]{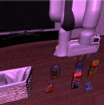}
    \hfill
    \includegraphics[width=0.48\linewidth]{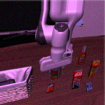}\\[-0.1em]
    \footnotesize \textbf{Domain pair:} same domain, different trajectory
\end{minipage}
\vspace{-0.35em}
\caption{\textbf{Pair construction for latent similarity diagnostics.}
Task pairs test whether a latent remains stable across visual-domain changes when the underlying trajectory content is preserved. Domain pairs test whether a latent collapses observations that share visual-domain cues despite different trajectory content.}
\label{fig:app_pair_examples}
\vspace{-0.6em}
\end{figure*}

\subsection{Final action-conditioning latent}
\label{app:repr_final_vla_latent}

The main paper analyzes the task encoder, domain encoder, and DILL's final action-conditioning latent (see Section~\ref{sec:repr}). Here, we further compare the final action-conditioning latent of DILL against that of Base VLA. This comparison directly tests whether the proposed objective changes the representation used by the policy in a useful way. If DILL works as intended, the action-conditioning latent should become less organized around shared visual domain and more organized around task-relevant trajectory content than the latent learned by an architecture-matched VLA trained with behavior cloning.

For Base VLA, we evaluate the final action-conditioning latent produced by the standard VLA policy. For DILL, we evaluate the final action-conditioning latent formed by concatenating the current representation and the predicted lookahead latent. We expect a task-centric action-conditioning latent to assign higher similarity to task pairs than to domain pairs: observations with the same trajectory content should remain close even when the visual domain changes, while observations that merely share the visual domain should remain separated when their trajectory content differs.

\begin{figure*}[h]
\centering
\setlength{\tabcolsep}{4pt}
\begin{tabular}{cc}
\footnotesize \textbf{Base VLA} &
\footnotesize \textbf{DILL (Ours)} \\[0.25em]
\includegraphics[width=0.47\textwidth]{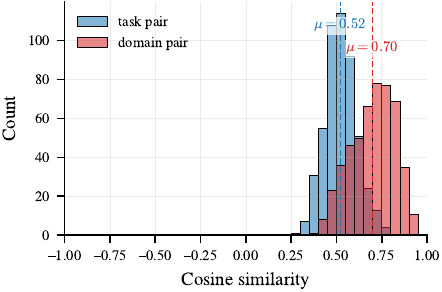} &
\includegraphics[width=0.47\textwidth]{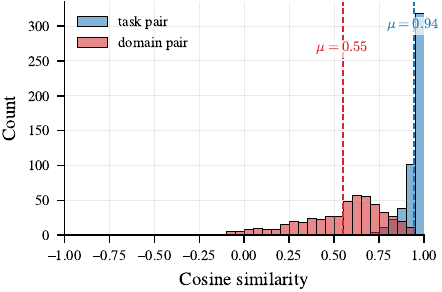}
\end{tabular}
\vspace{-0.4em}
\caption{\textbf{Final action-conditioning latent similarity diagnostics.}
We compare Base VLA and DILL using the final latent representation provided to the action head. A task-centric action-conditioning latent should assign higher similarity to task pairs than to domain pairs.}
\label{fig:app_final_vla_latent_similarity}
\vspace{-0.6em}
\end{figure*}

\begin{table}[h]
\centering
\small
\setlength{\tabcolsep}{7pt}
\renewcommand{\arraystretch}{1.22}
\begin{tabular}{lccc}
\toprule
Representation 
& \shortstack{Task-pair\\similarity $\uparrow$}
& \shortstack{Domain-pair\\similarity $\downarrow$}
& \shortstack{Gap\\$\Delta_{\text{task}} \uparrow$} \\
\midrule
Base VLA final latent
& $0.52$
& $0.70$
& $-0.18$ \\
\addlinespace[0.35em]
DILL final latent
& $0.94$
& $0.55$
& $0.39$ \\
\bottomrule
\end{tabular}

\vspace{0.85em}
\caption{\textbf{Summary of final action-conditioning latent similarity.}
The gap \(\Delta_{\mathrm{task}}\) is computed as mean task-pair similarity minus mean domain-pair similarity. Larger gap indicates a more task-centric action-conditioning representation.}
\label{tab:app_final_latent_similarity_summary}
\end{table}

\paragraph{Interpretation.}
The final action-conditioning latent is the representation most directly tied to control, since it is the input used by the action head. The results in Figure~\ref{fig:app_final_vla_latent_similarity} and Table~\ref{tab:app_final_latent_similarity_summary} show that Base VLA exhibits an undesirable ordering for shortcut-robust behavior: its domain-pair similarity ($0.70$) is higher than its task-pair similarity ($0.52$), yielding a negative gap of $-0.18$. This suggests that the standard VLA latent is more strongly organized by shared visual domain than by shared trajectory content. Such a representation may make the policy more susceptible to relying on domain-specific visual cues when task identity and viewpoint are confounded.

DILL reverses this ordering. Its final action-conditioning latent assigns much higher similarity to task pairs ($0.94$) than to domain pairs ($0.55$), increasing the separation gap from $-0.18$ to $0.39$. This indicates a substantial change in the structure of the representation used for action prediction: the latent becomes stable across domain changes when the underlying trajectory is preserved, while remaining discriminative when the trajectory content changes even within the same visual domain. The domain-pair similarity is not forced to vanish, which is expected because observations can still share scene layout and low-level visual statistics; the important point is that these shared visual domain factors no longer dominate the action-conditioning representation. {These representation-level observations complement the behavioral results: reduced shortcut reliance is accompanied by a more task-consistent geometry in the representation supplied to the action head.}

\begingroup

\subsection{Domain predictability with linear probes}
\label{app:domain_linear_probes}

Pairwise similarity does not directly reveal whether domain attributes remain predictable from a representation. We test this using cross-task linear probes: linear classifiers trained to predict viewpoint, lighting, or sensor-noise labels from the representation supplied to the action head. Accuracy above chance indicates that these domain cues remain linearly accessible.

Mean accuracy is $64.4\%$ for Base VLA and $25.4\%$ for DILL, compared with a chance reference of $21.7\%$. DILL's accuracy is closer to chance, suggesting reduced linear access to visual-domain information in the representation used for control. This complements the similarity analysis without establishing that all domain information has been removed.
\par
\endgroup

\clearpage
\begingroup

\section{Real-World Evaluation Protocol}
\label{app:real_world_protocol}

\paragraph{Robot setup and training.}
We use a 6-DoF Universal Robots UR5e equipped with a Robotiq 2-Finger Gripper. RGB observations are collected from Intel RealSense D435i cameras at a resolution of $640\times480$. Each model is trained with 50 demonstrations per task. For DILL and DILL-Current, the source-pretrained Task-Domain Encoder remains frozen: no real-world encoder fine-tuning or paired real-view data are used for encoder training.

\paragraph{Counterfactual shortcut diagnostic.}
Cup pointing and die placement test whether policies follow the commanded target when target color conflicts with viewpoint cues. We deliberately introduce a perfect correlation between target color and camera viewpoint in the training data: red-target instructions are paired only with the left view, whereas blue-target instructions are paired only with the right view. At test time, we reverse this assignment, evaluating red-target instructions from the right view and blue-target instructions from the left view. The instruction and desired manipulation remain unchanged; only the target-color--viewpoint pairing changes. Thus, both colors and both viewpoints are observed during training, but their test combinations are held out. Figure~\ref{fig:real_world_counterfactual_examples} illustrates the four instructions and their training and counterfactual test views.\par This protocol makes viewpoint an unreliable cue for target selection: a policy that relies on the training association may act on the wrong-colored object instead of following the instruction. Counterfactual (CTF) success measures execution of the commanded task. Shortcut degree (SD) measures execution of the target associated with the observed viewpoint during training. It therefore measures a specific task-substitution failure rather than all unsuccessful trials.\par \begin{figure}[p] \centering \begingroup \small \definecolor{ctfred}{RGB}{163,47,51} \definecolor{ctfblue}{RGB}{37,88,136} \setlength{\fboxsep}{0pt} \newcommand{\ctfimg}[1]{\includegraphics[width=0.194\linewidth]{figures/real_world_exp/frames/#1}} \newcommand{\ctftrain}[1]{\begin{minipage}[b]{0.804\linewidth}#1\end{minipage}} \newcommand{\ctftest}[1]{\begin{minipage}[b]{0.156\linewidth}\includegraphics[width=\linewidth]{figures/real_world_exp/frames/#1}\end{minipage}} \newcommand{\ctfgroup}[4]{\colorbox{#1!9}{\parbox[c][19pt][c]{\linewidth}{\makebox[0.804\linewidth]{\hspace{5pt}\textcolor{#1}{\textbf{#2}}\hfill\textbf{#3}\hfill\phantom{\textbf{#2}}\hspace{5pt}}\makebox[0.04\linewidth]{$\rightarrow$}\makebox[0.156\linewidth]{\textbf{#4}}}}\par\vspace{5pt}} \makebox[0.804\linewidth]{\shortstack{\textbf{Train}\\[-1pt]\footnotesize Demonstrations}}\hfill\makebox[0.156\linewidth]{\shortstack{\textbf{Test}\\[-1pt]\footnotesize Counterfactual}}\par\vspace{6pt} \ctfgroup{ctfred}{Red targets}{Left view}{Right view} \makebox[\linewidth][l]{\textit{``Point at the \textcolor{ctfred}{\textbf{red}} cup.''}}\\[3pt] \ctftrain{\ctfimg{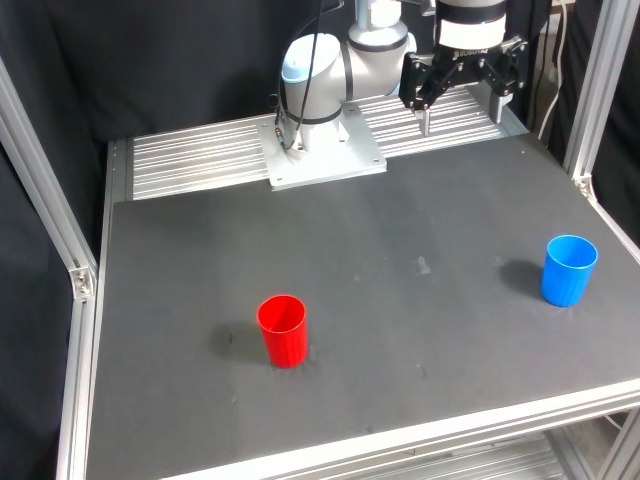}\hfill\ctfimg{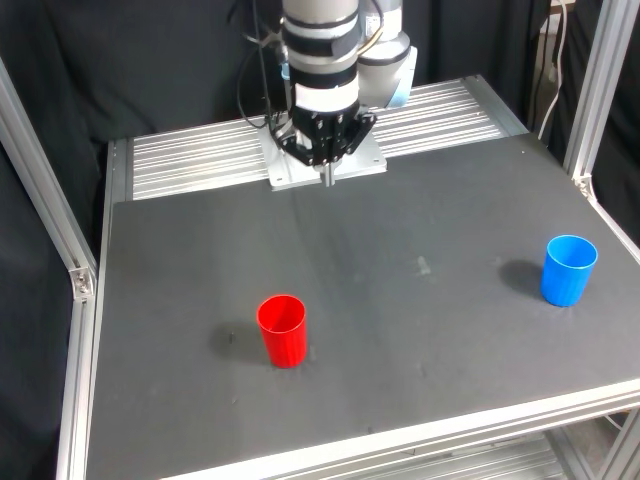}\hfill\ctfimg{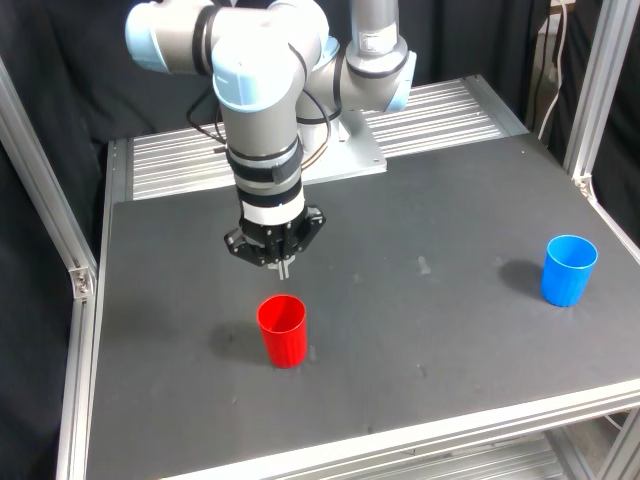}\hfill\ctfimg{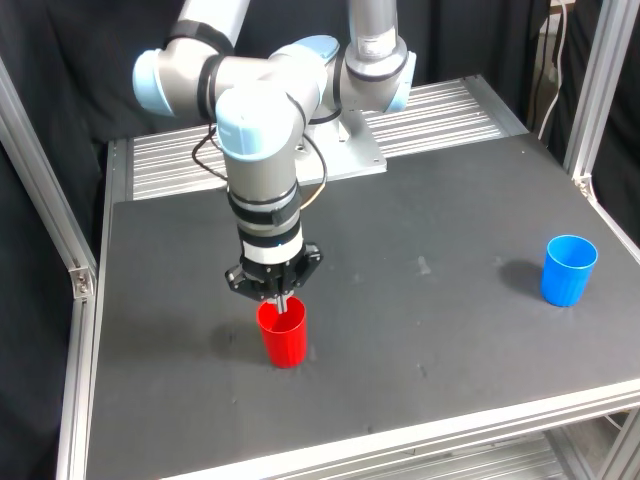}\hfill\ctfimg{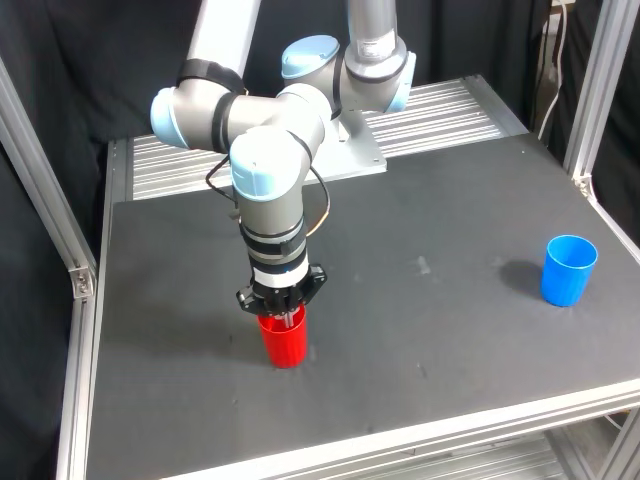}}\hfill\ctftest{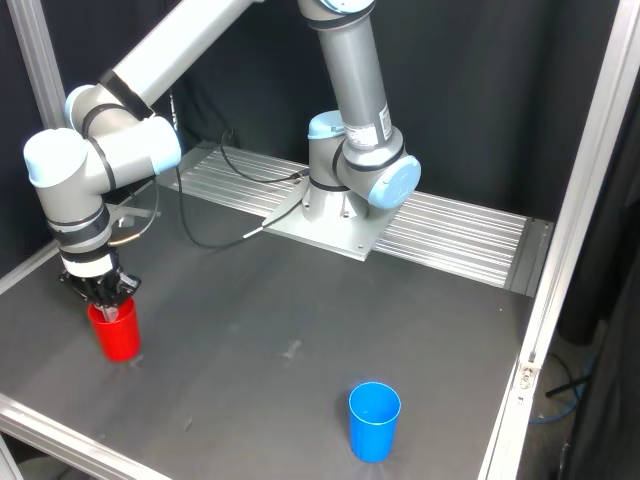}\par \vspace{7pt} \makebox[\linewidth][l]{\textit{``Pick up the \textcolor{ctfred}{\textbf{red}} die and place it in the basket.''}}\\[3pt] \ctftrain{\ctfimg{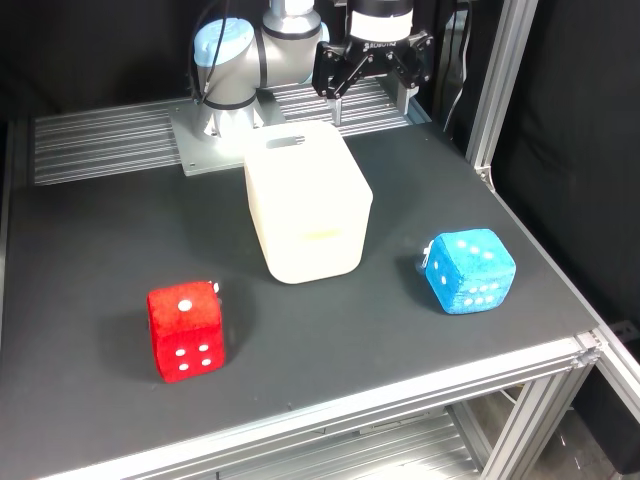}\hfill\ctfimg{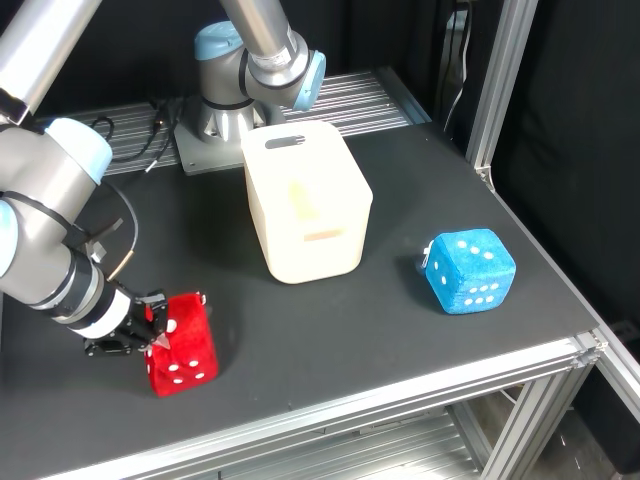}\hfill\ctfimg{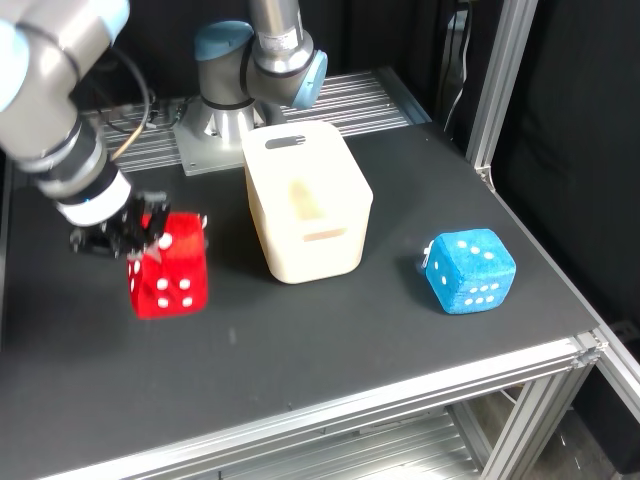}\hfill\ctfimg{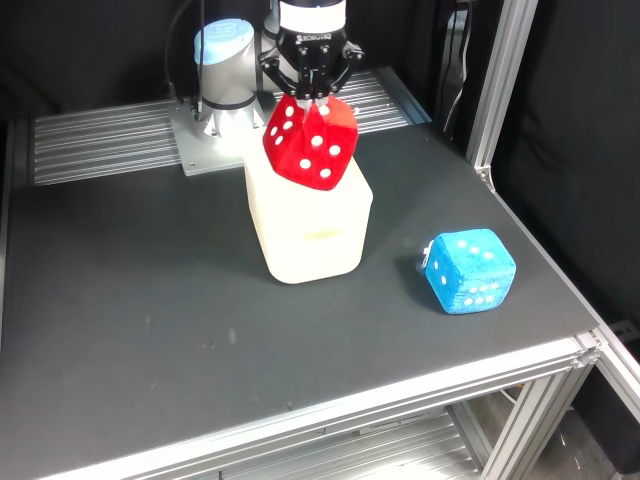}\hfill\ctfimg{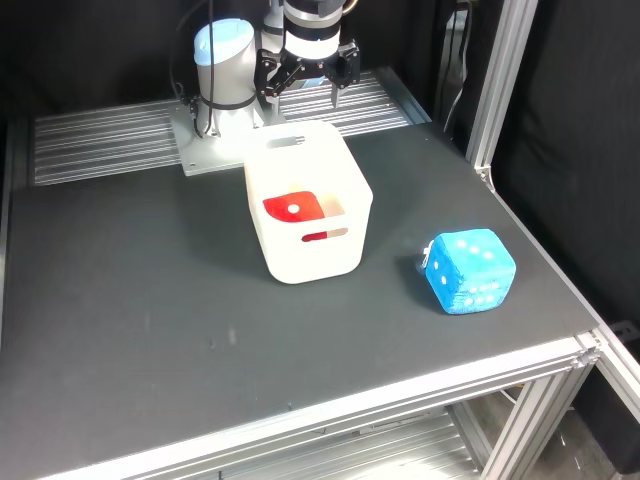}}\hfill\ctftest{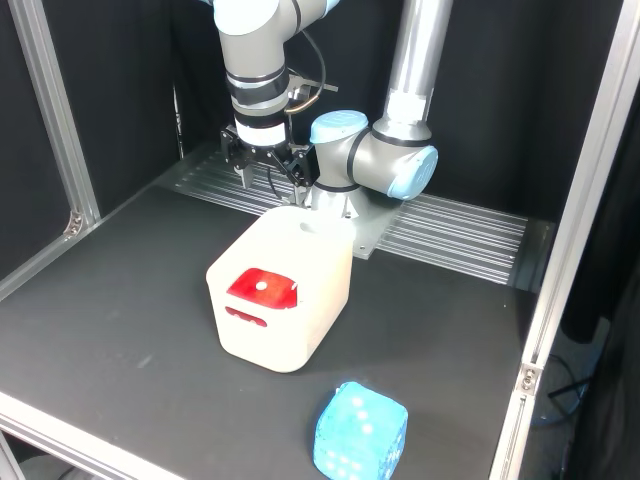}\par \vspace{13pt} \ctfgroup{ctfblue}{Blue targets}{Right view}{Left view} \makebox[\linewidth][l]{\textit{``Point at the \textcolor{ctfblue}{\textbf{blue}} cup.''}}\\[3pt] \ctftrain{\ctfimg{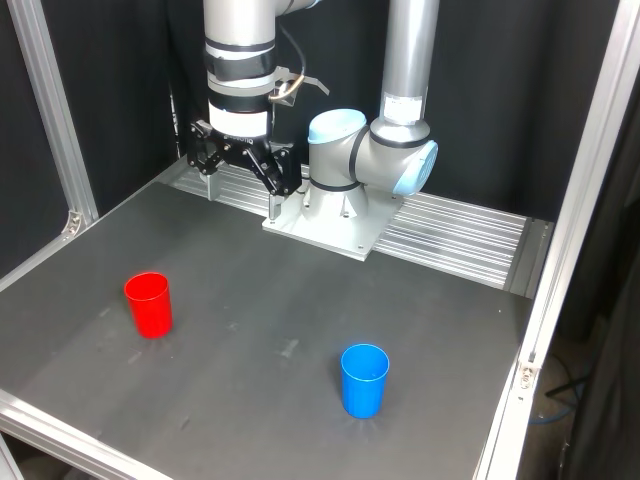}\hfill\ctfimg{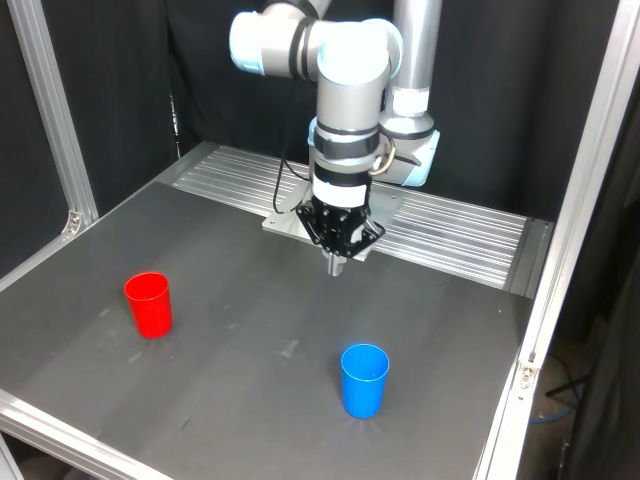}\hfill\ctfimg{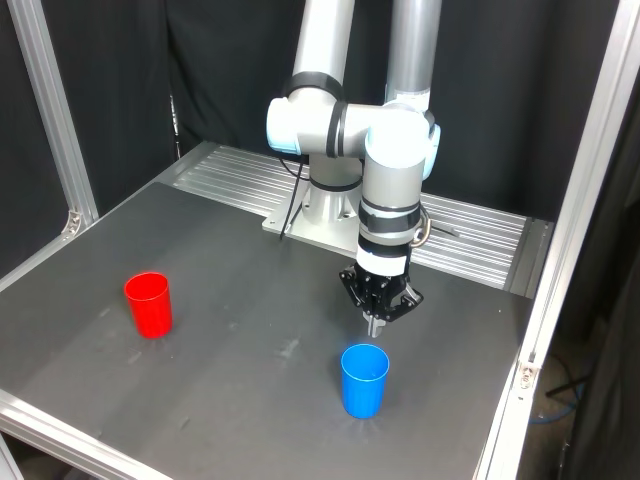}\hfill\ctfimg{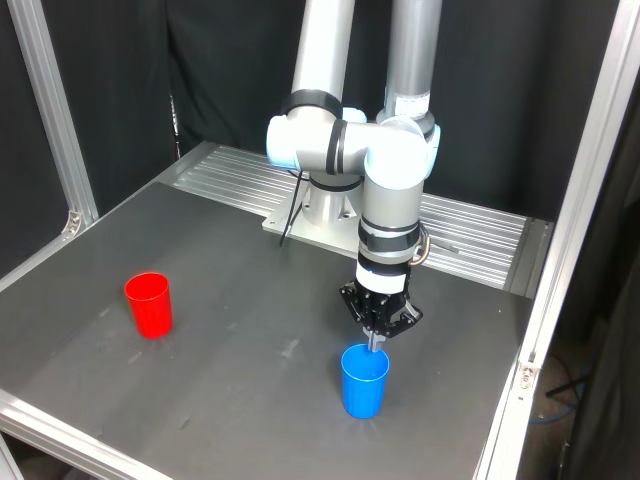}\hfill\ctfimg{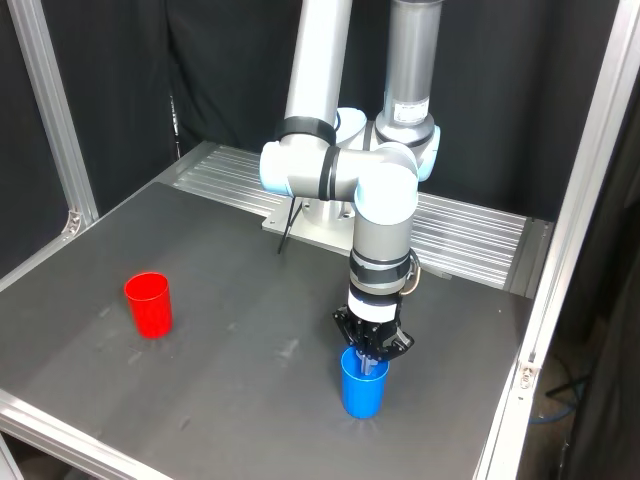}}\hfill\ctftest{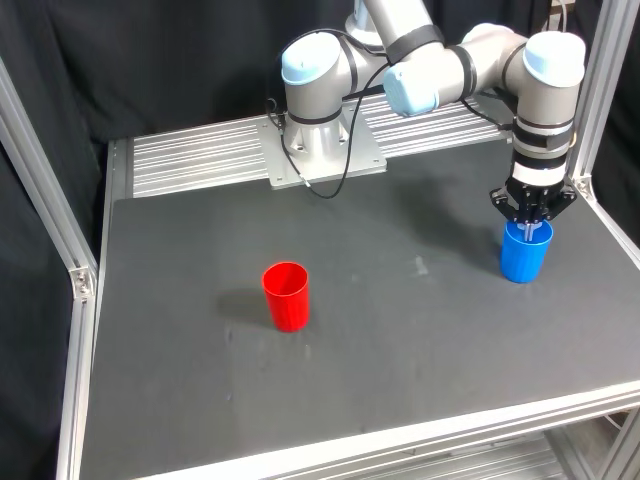}\par \vspace{7pt} \makebox[\linewidth][l]{\textit{``Pick up the \textcolor{ctfblue}{\textbf{blue}} die and place it in the basket.''}}\\[3pt] \ctftrain{\ctfimg{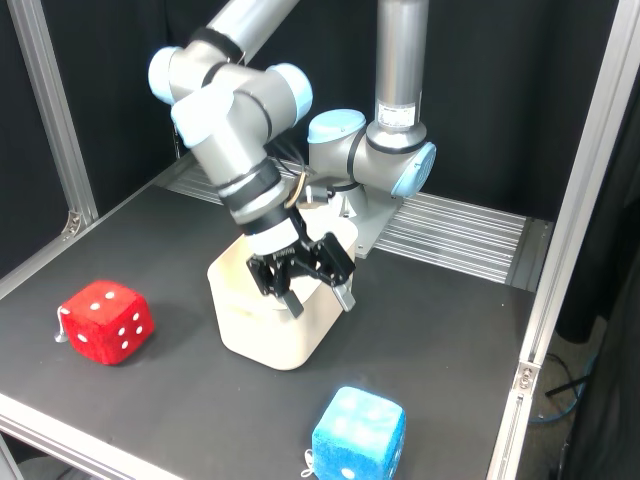}\hfill\ctfimg{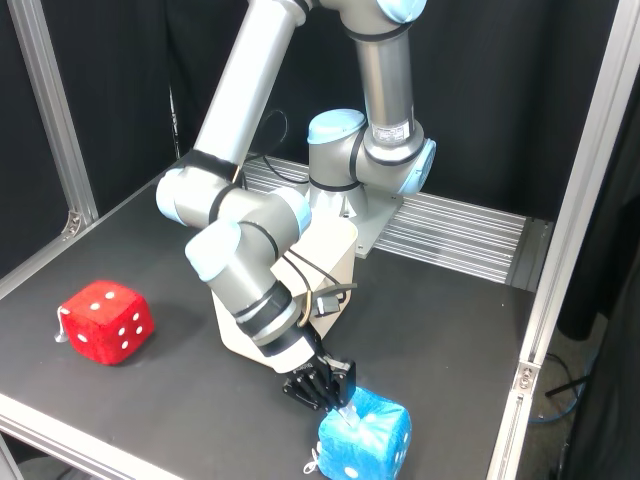}\hfill\ctfimg{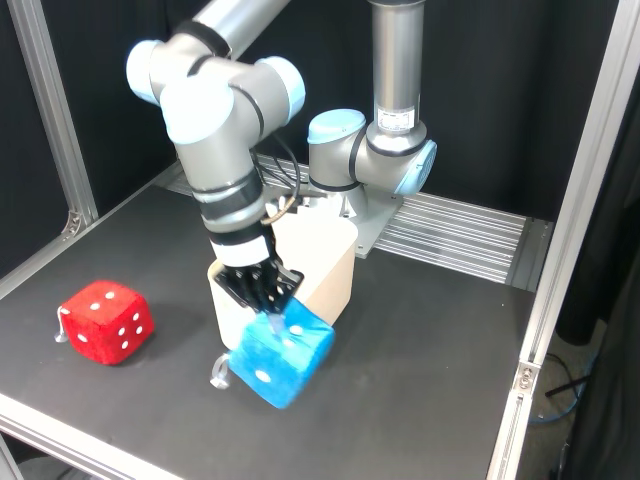}\hfill\ctfimg{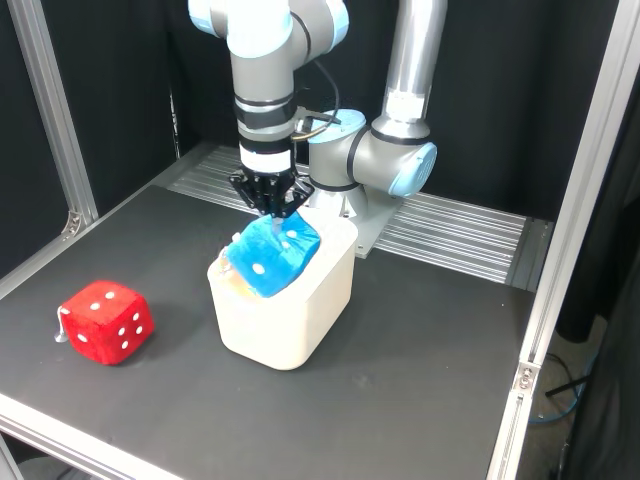}\hfill\ctfimg{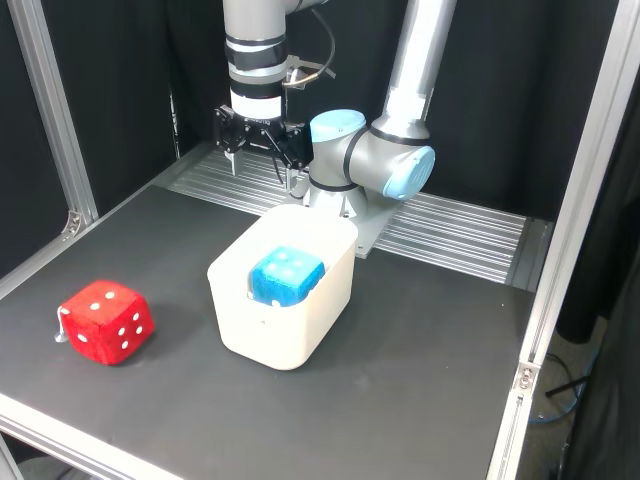}}\hfill\ctftest{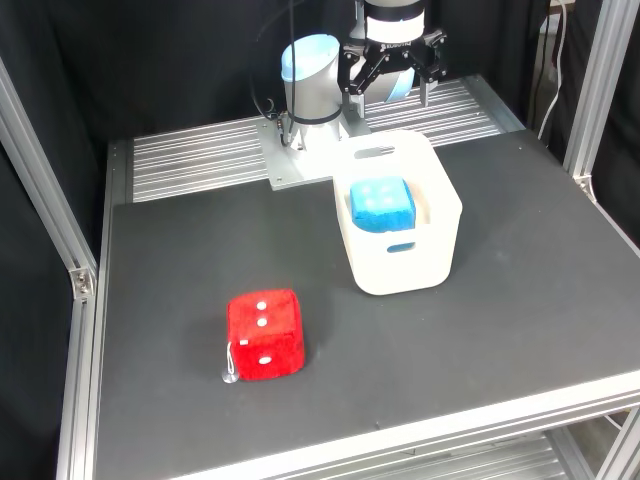}\par \endgroup \caption{\textbf{Real-world counterfactual color--viewpoint compositions.} Training pairs red-target instructions only with the left view and blue-target instructions only with the right view, creating a spurious correlation between target color and viewpoint. Counterfactual testing swaps these pairings while preserving the instruction and desired manipulation. Each row shows five frames from a training demonstration and one example of the held-out test view.} \label{fig:real_world_counterfactual_examples} \end{figure}

\paragraph{Visual robustness.}
Shoe upright placement, tissue pulling, and laptop closing test whether policies retain instruction-following performance under visual changes. Figure~\ref{fig:real_world_robustness_examples}(a) shows training demonstrations of the three tasks, and panel~(b) organizes their shared evaluation conditions. Each task is evaluated under all three conditions, with its instruction and manipulation goal unchanged.\par \emph{No perturbation} uses the standard scene without added visual changes. \emph{Predefined perturbation} uses held-out real instances of transformation families used during Task-Domain Encoder pretraining: viewpoint, background, lighting, and sensor noise. \emph{New perturbation} uses cast shadows, dynamic backgrounds, and foreground clutter, which are absent from the encoder's positive-pair construction. Thus, ``predefined'' refers to the transformation family, not to prior exposure to the real evaluation images. Comparing these conditions tests transfer both within and beyond the transformation families used to learn invariance. For DILL and DILL-Current, the source-pretrained encoder remains frozen throughout real-world policy fine-tuning and evaluation.\par \begin{figure}[p]
\centering
\begingroup
\small
\definecolor{rbink}{RGB}{48,62,73}
\definecolor{rbknown}{RGB}{47,104,135}
\definecolor{rbnew}{RGB}{112,77,130}
\setlength{\fboxsep}{0pt}
\newcommand{\rbimg}[1]{\includegraphics[width=0.194\linewidth]{figures/figure10_final/#1}}
\newcommand{\rbheading}[1]{\colorbox{rbink!8}{\parbox[c][18pt][c]{\linewidth}{\hspace{5pt}\textbf{#1}}}\par\vspace{5pt}}
\rbheading{(a) Training demonstrations (front view)}
\makebox[\linewidth][l]{\textit{``Pick up the shoe and stand it upright.''}}\\[3pt]
\rbimg{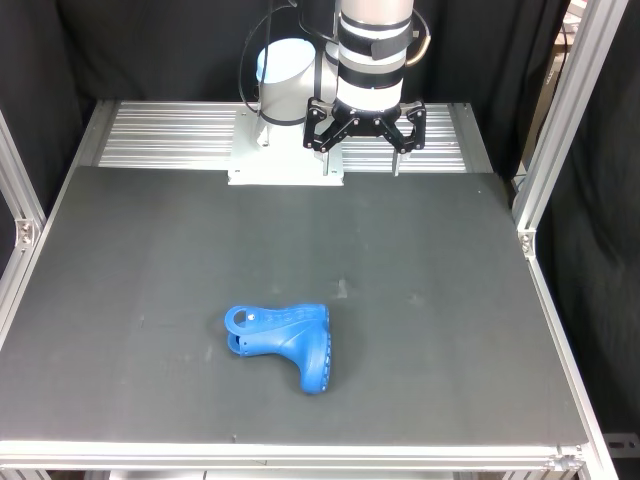}\hfill\rbimg{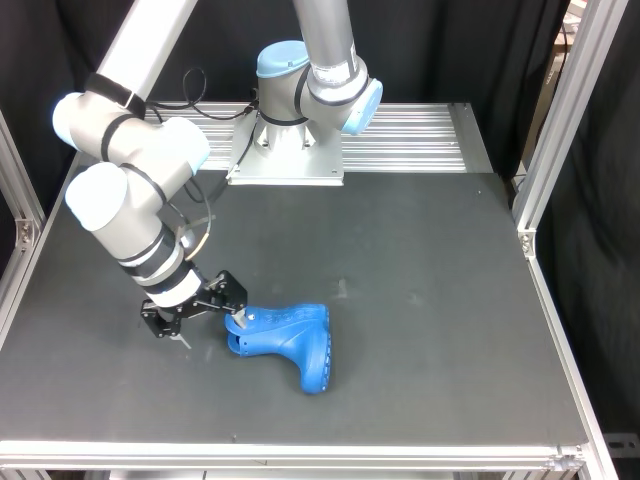}\hfill\rbimg{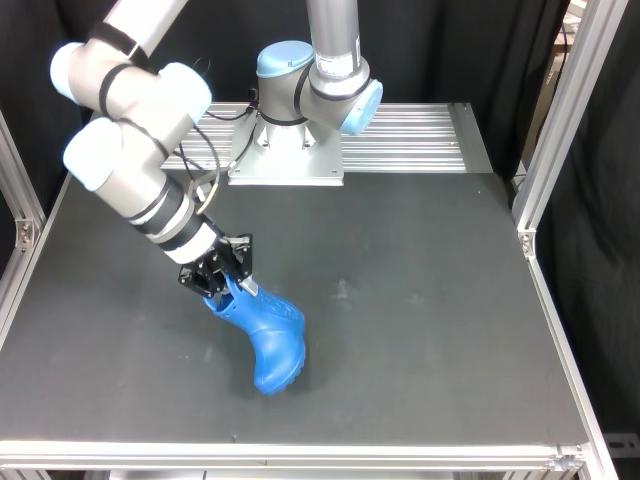}\hfill\rbimg{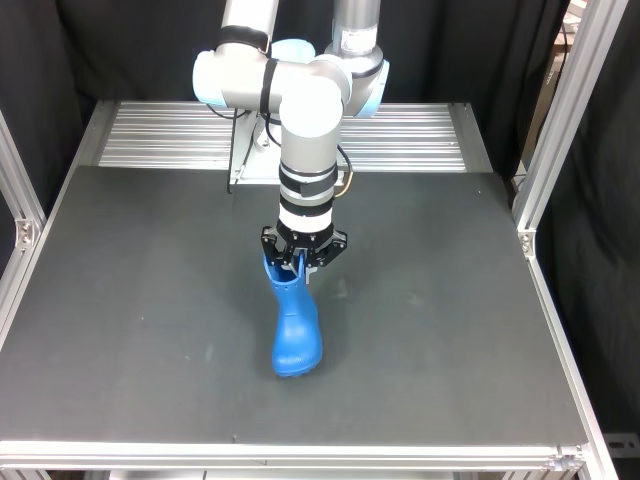}\hfill\rbimg{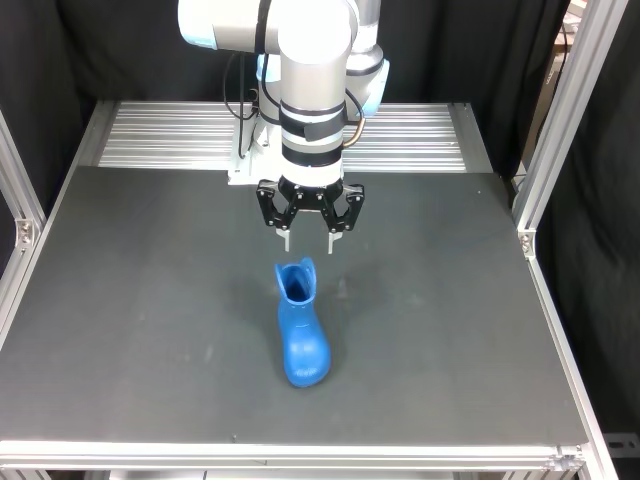}\par
\vspace{7pt}
\makebox[\linewidth][l]{\textit{``Pull a tissue out of the box.''}}\\[3pt]
\rbimg{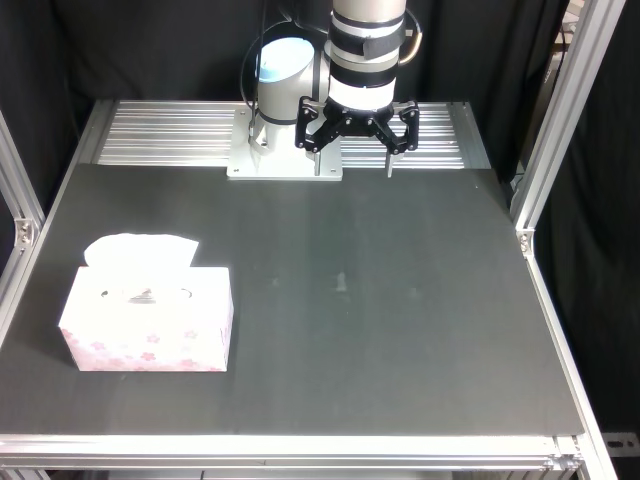}\hfill\rbimg{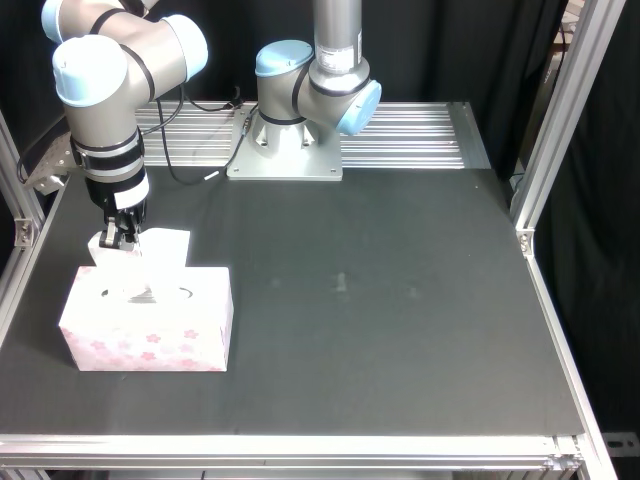}\hfill\rbimg{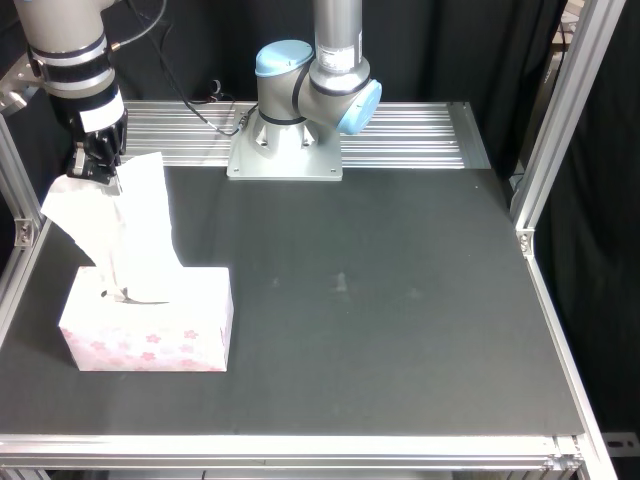}\hfill\rbimg{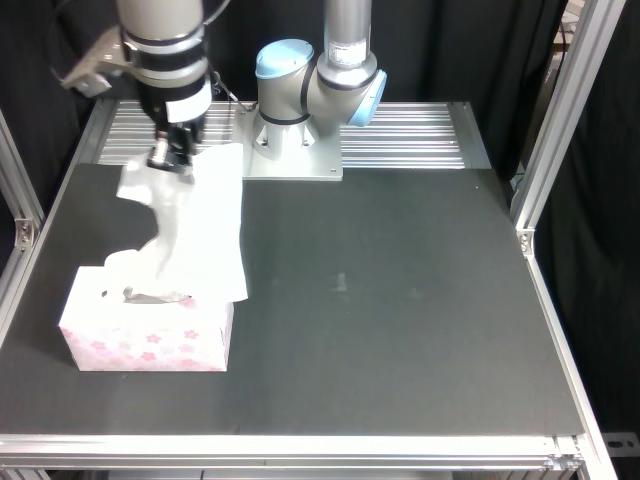}\hfill\rbimg{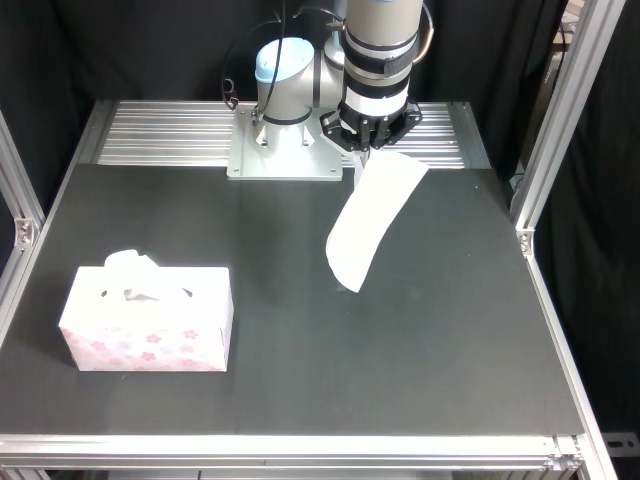}\par
\vspace{7pt}
\makebox[\linewidth][l]{\textit{``Close the laptop.''}}\\[3pt]
\rbimg{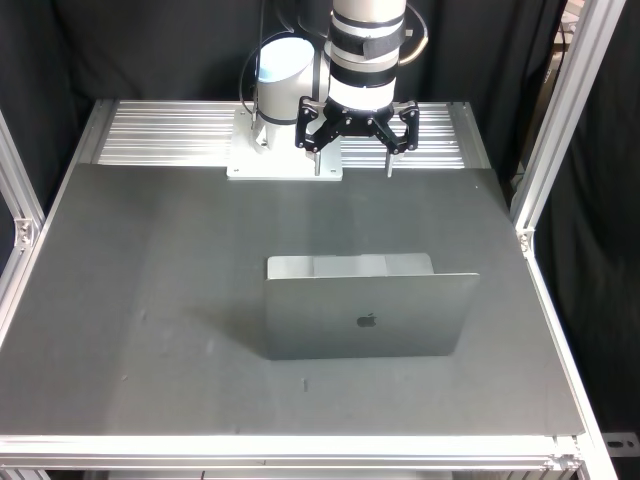}\hfill\rbimg{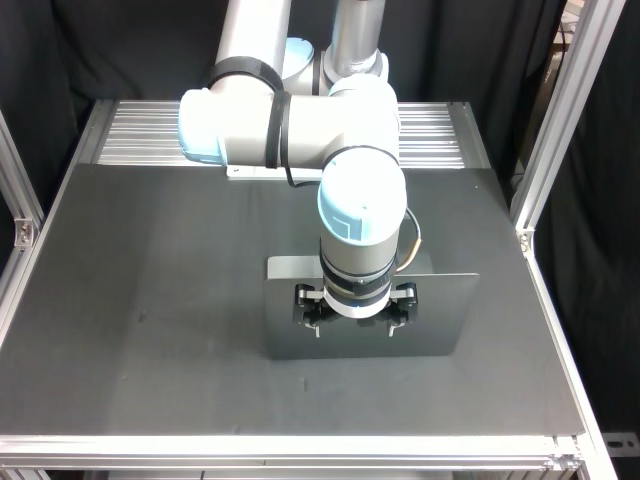}\hfill\rbimg{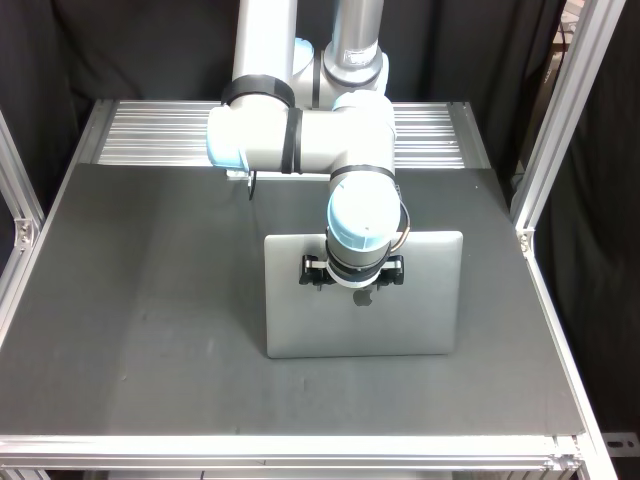}\hfill\rbimg{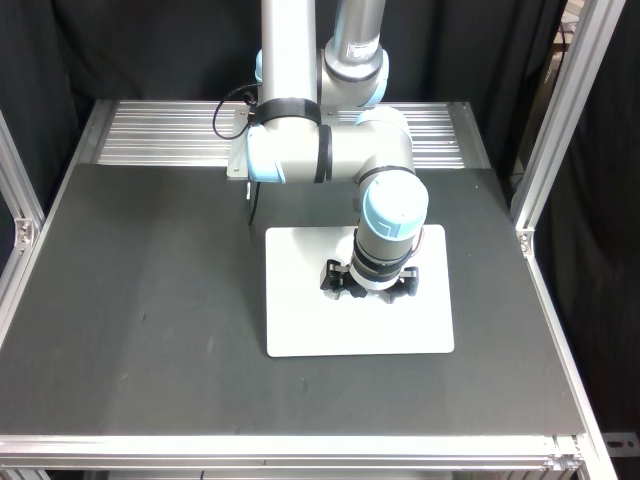}\hfill\rbimg{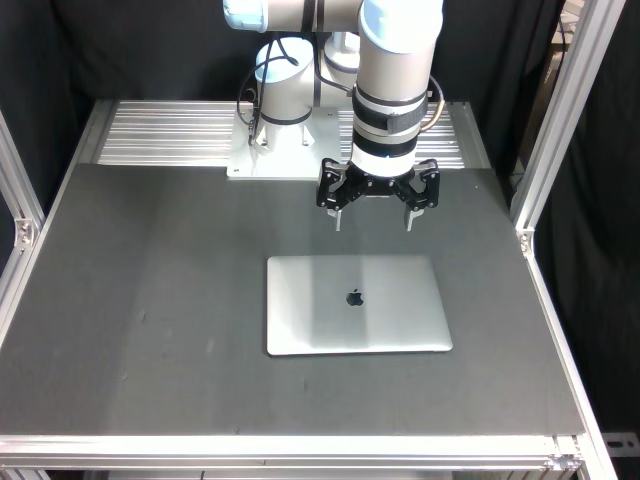}\par
\vspace{12pt}
\rbheading{(b) Robustness evaluation}
\makebox[\linewidth][l]{\footnotesize Each task is evaluated under all three conditions; representative scenes are shown below.}\par\vspace{7pt}
\begin{minipage}[t]{0.194\linewidth}
\centering\footnotesize
\textbf{No perturbation}\par\vspace{3pt}
{\color{rbink}\rule{\linewidth}{1.2pt}}\par\vspace{4pt}
\includegraphics[width=\linewidth]{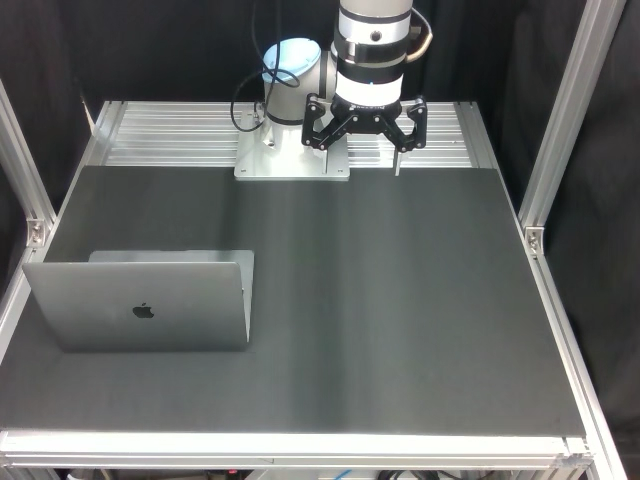}\par\vspace{3pt}
\parbox[t][20pt][t]{\linewidth}{\centering Standard scene}
\end{minipage}\hfill
\begin{minipage}[t]{0.3955\linewidth}
\centering\footnotesize
\textbf{Predefined perturbations}\par\vspace{3pt}
{\color{rbknown}\rule{\linewidth}{1.2pt}}\par\vspace{4pt}
\begin{minipage}[t]{0.4905\linewidth}
\centering
\includegraphics[width=\linewidth]{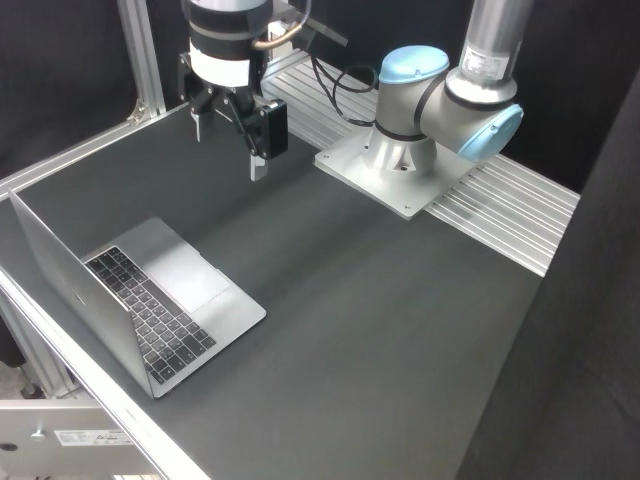}\par\vspace{3pt}
\parbox[t][20pt][t]{\linewidth}{\centering Viewpoint}
\end{minipage}\hfill
\begin{minipage}[t]{0.4905\linewidth}
\centering
\includegraphics[width=\linewidth]{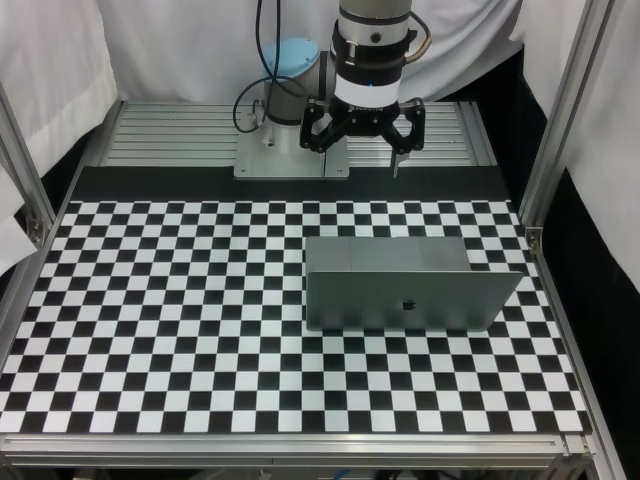}\par\vspace{3pt}
\parbox[t][20pt][t]{\linewidth}{\centering Background}
\end{minipage}\par\vspace{5pt}
\textcolor{rbknown}{Families used in encoder pretraining}
\end{minipage}\hfill
\begin{minipage}[t]{0.3955\linewidth}
\centering\footnotesize
\textbf{New perturbations}\par\vspace{3pt}
{\color{rbnew}\rule{\linewidth}{1.2pt}}\par\vspace{4pt}
\begin{minipage}[t]{0.4905\linewidth}
\centering
\includegraphics[width=\linewidth]{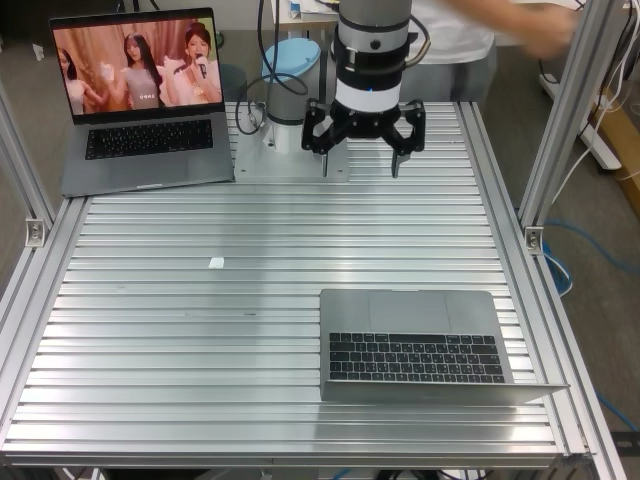}\par\vspace{3pt}
\parbox[t][20pt][t]{\linewidth}{\centering Dynamic\\background}
\end{minipage}\hfill
\begin{minipage}[t]{0.4905\linewidth}
\centering
\includegraphics[width=\linewidth]{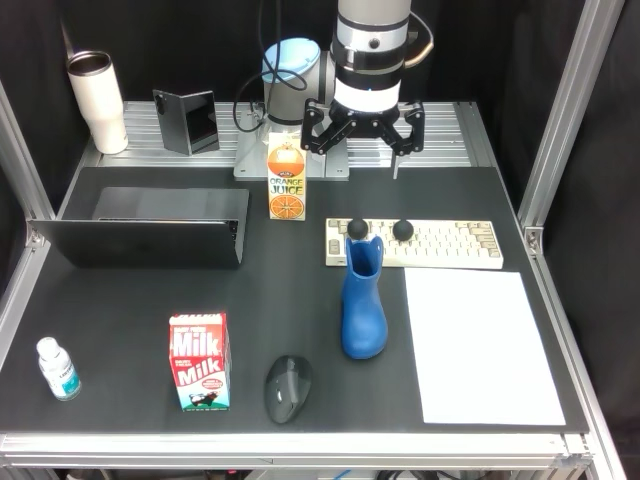}\par\vspace{3pt}
\parbox[t][20pt][t]{\linewidth}{\centering Foreground clutter}
\end{minipage}\par\vspace{5pt}
\textcolor{rbnew}{Families absent from encoder pretraining}
\end{minipage}\par
\endgroup
\caption{\textbf{Real-world visual robustness tasks and evaluation conditions.} (a) Five ordered frames from the training demonstration for each task: shoe upright placement, tissue pulling, and laptop closing. (b) Representative evaluation scenes: no perturbation; viewpoint and background changes from predefined transformation families; and dynamic backgrounds and foreground clutter from new families absent from Task-Domain Encoder pair construction. All three tasks are evaluated under all three conditions, with their instructions and manipulation goals unchanged. Quantitative results are reported in Table~\ref{tab:real_world_main}.}
\label{fig:real_world_robustness_examples}
\end{figure}

\par
\endgroup

\end{document}